\documentclass[11pt]{article}
\usepackage[utf8]{inputenc}
\usepackage[T1]{fontenc}
\usepackage{fullpage}
\usepackage{xspace}
\usepackage[dvipsnames, table]{xcolor}
\usepackage[toc,page,header]{appendix}
\usepackage{minitoc} 
\usepackage[maxbibnames=99, natbib=true, style=nature]{biblatex}
\usepackage{graphicx}
\usepackage{subcaption}
\usepackage[export]{adjustbox}
\usepackage{amsmath}
\usepackage{amssymb}
\usepackage[flushleft]{threeparttable} 
\usepackage{booktabs}
\usepackage{subfiles}
\usepackage{footmisc}
\usepackage{courier}
\usepackage{hyperref}
\hypersetup{
    colorlinks,
    citecolor=black,
    filecolor=black,
    linkcolor=black,
    urlcolor=black
}
\usepackage{subcaption}

\newcommand{\obssize}{$\sim16,000$\xspace}

\newcommand{\categorical}{\cellcolor{blue!25}}

\newcommand\nicetohave[1]{}

\newcommand{\cleanexpname}{Rerun\xspace}
\newcommand{\trueskillname}{TrueSkill\xspace}
\newcommand{\pflopsday}{\textrm{PFlops/s}$\cdot$\textrm{days}\xspace}
\newcommand{\totalcompute}{$770\pm50$ \pflopsday\xspace}
\newcommand{\dota}{Dota 2\xspace}
\newcommand{\openaifive}{OpenAI Five\xspace}
\newcommand{\openaiarena}{OpenAI Five Arena\xspace}
\newcommand{\openaifinals}{OpenAI Five Finals\xspace}
\newcommand{\genericcaptionsentence}[1]{\trueskillname over the course of training (see \autoref{appendix:trueskill}) and speedup measured by the rate to attain different \trueskillname thresholds (computed using \autoref{eqn:speedup}) granted by #1\xspace}

\title{\dota with Large Scale Deep Reinforcement Learning}
\author{
 \parbox{\linewidth}{
 \centering
 \textbf{OpenAI},
\footnote{
Authors listed alphabetically. Please cite as
OpenAI et al., and use the following bibtex for citation: \url{https://openai.com/bibtex/openai2019dota.bib}
}
\\ 
 Christopher~Berner, Greg~Brockman, Brooke~Chan, Vicki~Cheung, Przemys\l{}aw~``Psyho"~D\k{e}biak, Christy~Dennison, David~Farhi, Quirin~Fischer, Shariq~Hashme, Chris~Hesse, Rafal J\'{o}zefowicz, Scott~Gray, Catherine~Olsson, Jakub~Pachocki, Michael~Petrov, Henrique~Pond\'e de Oliveira Pinto, Jonathan~Raiman, Tim~Salimans, Jeremy~Schlatter, Jonas~Schneider, Szymon~Sidor, Ilya~Sutskever, Jie~Tang, Filip~Wolski, Susan~Zhang
 }
}
\begin{document}

\doparttoc 
\faketableofcontents 

\maketitle

\begin{abstract}
On April 13th, 2019, \openaifive became the first AI system to defeat the world champions at an esports game.
The game of \dota presents novel challenges for AI systems such as long time horizons, imperfect information, and complex, continuous state-action spaces, all challenges which will become increasingly central to more capable AI systems.
\openaifive leveraged existing reinforcement learning techniques, scaled to learn from batches of approximately 2 million frames every 2 seconds. 
We developed a distributed training system and tools for continual training which allowed us to train \openaifive for 10 months.
By defeating the \dota world champion (Team OG), \openaifive demonstrates that self-play reinforcement learning can achieve superhuman performance on a difficult task.
\end{abstract}

\section{Introduction}

The long-term goal of artificial intelligence is to solve advanced real-world challenges. 
Games have served as stepping stones along this path for decades, from Backgammon (1992) to Chess (1997) to Atari (2013)\cite{tesauro1994td, campbell2002deepblue, mnih2013playing}.
In 2016, AlphaGo defeated the world champion at Go using deep reinforcement learning and Monte Carlo tree search\cite{silver2016mastering}. 
In recent years, reinforcement learning (RL) models have tackled tasks as varied as robotic manipulation\cite{dactyl}, text summarization \cite{paulus2017deep}, and video games such as Starcraft\cite{alphastarblog} and Minecraft\cite{guss2019minerl}.

Relative to previous AI milestones like Chess or Go, complex video games start to capture the complexity and continuous nature of the real world. 
\dota is a multiplayer real-time strategy game produced by Valve Corporation in 2013, which averaged between 500,000 and 1,000,000 concurrent players between 2013 and 2019. 
The game is actively played by full time professionals; the prize pool for the 2019 international championship exceeded \$35 million (the largest of any esports game in the world)\cite{wiki:dota2, wiki:ti2018}.
The game presents challenges for reinforcement learning due to
long time horizons, partial observability, and high dimensionality of observation and action spaces.
\dota's rules are also complex --- the game has been actively developed for over a decade, with game logic implemented in hundreds of thousands of lines of code.

The key ingredient in solving this complex environment was to scale existing reinforcement learning systems to unprecedented levels, utilizing thousands of GPUs over multiple months. 
We built a distributed training system to do this which we used to train a \dota-playing agent called \openaifive. 
In April 2019, \openaifive defeated the \dota world champions (Team OG\footnote{\url{https://www.facebook.com/OGDota2/}}), the first time an AI system has beaten an esport world champion\footnote{Full game replays and other supplemental can be downloaded from: \url{https://openai.com/blog/how-to-train-your-openai-five/}}. We also opened \openaifive to the \dota community for competitive play; \openaifive won 99.4\% of over 7000 games.

One challenge we faced in training was that the environment and code continually changed as our project progressed. 
In order to train without restarting from the beginning after each change, we developed a collection of tools to resume training with minimal loss in performance which we call {\it surgery}. 
Over the 10-month training process, we performed approximately one surgery per two weeks. 
These tools allowed us to make frequent improvements to our strongest agent within a shorter time than the typical practice of training from scratch would allow.
As AI systems tackle larger and harder problems, further investigation of settings with ever-changing environments and iterative development will be critical.

In \autoref{sec:dota}, we describe \dota in more detail along with the challenges it presents. In \autoref{sec:training-system} we discuss the technical components of the training system, leaving most of the details to appendices cited therein. 
In \autoref{sec:results}, we summarize our long-running experiment and the path that lead to defeating the world champions. 
We also describe lessons we've learned about reinforcement learning which may generalize to other complex tasks.

\section{\dota}\label{sec:dota}

\dota is played on a square map with two teams defending bases in opposite corners.
Each team's base contains a structure called an ancient; the game ends when one of these ancients is destroyed by the opposing team.
Teams have five players, each controlling a hero unit with unique abilities.
During the game, both teams have a constant stream of small ``creep'' units, uncontrolled by the players, which walk towards the enemy base attacking any opponent units or buildings.
Players gather resources such as gold from creeps, which they use to increase their hero's power by purchasing items and improving abilities.\footnote{Further information the rules and gameplay of \dota is readily accessible online; a good introductory resource is \url{https://purgegamers.true.io/g/dota-2-guide/}}

To play \dota, an AI system must address various challenges:

\begin{itemize}
\item \textbf{Long time horizons.}
\dota games run at 30 frames per second for approximately 45 minutes.
\openaifive selects an action every fourth frame, yielding approximately 20,000 steps per episode. By comparison, chess usually lasts 80 moves, Go 150 moves\cite{Allis1994branching}. 

\item \textbf{Partially-observed state.}
Each team in the game can only see the portion of the game state near their units and buildings; the rest of the map is hidden. 
Strong play requires making inferences based on incomplete data, and modeling the opponent's behavior. 

\item\textbf{High-dimensional action and observation spaces.}
\dota is played on a large map containing ten heroes, dozens of buildings, dozens of non-player units, and a long tail of game features such as runes, trees, and wards. 
\openaifive observes \obssize total values (mostly floats and categorical values with hundreds of possibilities) each time step. 
We discretize the action space; on an average timestep our model chooses among 8,000 to 80,000 actions (depending on hero).
For comparison Chess requires around one thousand values per observation (mostly 6-possibility categorical values) 
and Go around six thousand values (all binary)\cite{silver2018general}. 
Chess has a branching factor of around 35 valid actions, and Go around 250\cite{Allis1994branching}.
\end{itemize}

Our system played \dota with two limitations from the regular game:
\begin{itemize}
    \item Subset of 17 heroes --- in the normal game players select before the game one from a pool of 117 heroes to play; we support 17 of them.\footnote{See \autoref{sec:multihero} for experiments characterizing the effect of hero pool size.}
    \item No support for items which allow a player to temporarily control multiple units at the same time (Illusion Rune, Helm of the Dominator, Manta Style, and Necronomicon). We removed these to avoid the added technical complexity of enabling the agent to control multiple units.
\end{itemize}

\section{Training System}\label{sec:training-system}
\subsection{Playing Dota using AI}

Humans interact with the \dota game using a keyboard, mouse, and computer monitor. 
They make decisions in real time, reason about long-term consequences of their actions, and more.
We adopt the following framework to translate the vague problem of ``play this complex game at a superhuman level" into a detailed objective suitable for optimization.

Although the \dota engine runs at 30 frames per second, \openaifive only acts on every 4th frame which we call a {\it timestep}.
Each timestep, \openaifive receives an {\it observation} 
from the game engine encoding all the information a human player would see such as units' health, position, etc (see \autoref{appendix:observationspace} for an in-depth discussion of the observation).
\openaifive then returns a discrete {\it action} to the game engine, encoding a desired movement, attack, etc. 

Certain game mechanics were controlled by hand-scripted logic rather than the policy: the order in which heroes purchase items and abilities, control of the unique courier unit, and which items heroes keep in reserve. 
While we believe the agent could ultimately perform better if these actions were not scripted, we achieved superhuman performance before doing so. 
Full details of our action space and scripted actions are described in \autoref{appendix:actionspace}. 

Some properties of the environment were randomized during training, including the heroes in the game and which items the heroes purchased. Sufficiently diverse training games are necessary to ensure robustness to the wide variety of strategies and situations that arise in games against human opponents. 
See \autoref{sec:random} for details of the domain randomizations.

\begin{figure}[ht]
\includegraphics[width=\textwidth]{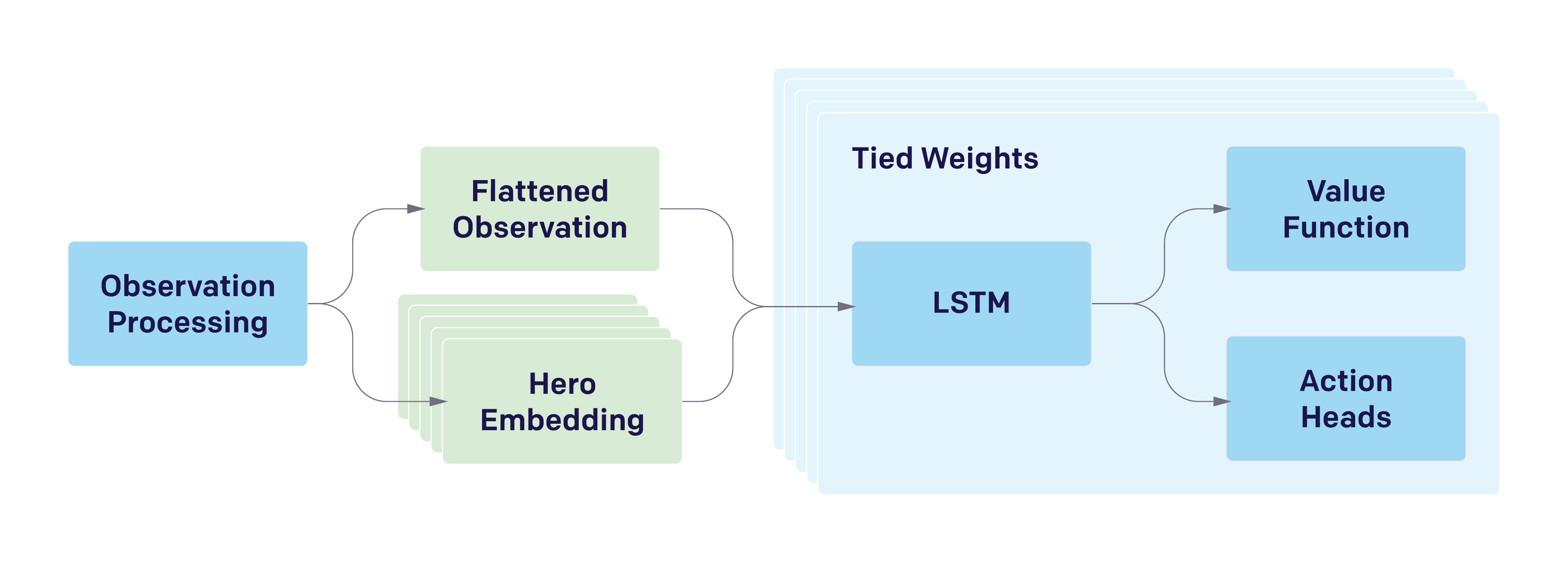}
\caption{\textbf{Simplified \openaifive Model Architecture:} 
The complex multi-array observation space is processed into a single vector, which is then passed through a 4096-unit LSTM.
The LSTM state is projected to obtain the policy outputs (actions and value function). 
Each of the five heroes on the team is controlled by a replica of this network with nearly identical inputs, each with its own hidden state.
The networks take different actions due to a part of the observation processing's output indicating which of the five heroes is being controlled.
The LSTM composes 84\% of the model's total parameter count.
See \autoref{fig:observationlogic} and \autoref{fig:actionspace} in \autoref{appendix:network-architecture} for a detailed breakdown of our model architecture.}
\label{fig:architecture}
\end{figure}

We define a {\it policy} ($\pi$) as a function from the history of observations to a probability distribution over actions, which we parameterize as a recurrent neural network with approximately 159 million parameters ($\theta$).
The neural network consists primarily of a single-layer 4096-unit LSTM \cite{gers1999learning} (see \autoref{fig:architecture}).
Given a policy, we play games by repeatedly passing the current observation as input and sampling an action from the output distribution at each timestep. 

Separate replicas of the same policy function (with identical parameters $\theta$) are used to control each of the five heroes on the team.
Because visible information and {\it fog of war} (area that is visible to players due to proximity of friendly units) are shared across a team in \dota, the observations are nearly\footnote{We do include a very small number of derived features which depend on the hero being controlled, for example the ``distance to me'' feature of each unit in the game.} identical for each hero.

Instead of using the pixels on the screen, we approximate the information available to a human player in a set of data arrays (see \autoref{appendix:observationspace} for full details of the observations space). 
This approximation is imperfect; there are small pieces of information which humans can gain access to which we have not encoded in the observations. 
On the flip side, while we were careful to ensure that all the information available to the model is also available to a human, the model does get to see \emph{all} the information available simultaneously every time step, whereas a human needs to actively click to see various parts of the map and status modifiers. 
\openaifive uses this semantic observation space for two reasons: First, because our goal is to study strategic planning and high-level decision-making rather than focus on visual processing. 
Second, it is infeasible for us to render each frame to pixels in all training games; this would multiply the computation resources required for the project many-fold.
Although these discrepancies exist, we do not believe they introduce significant bias when benchmarking against human players.
To allow the five networks to choose different actions, the LSTM receives an extra input from the observation processing, indicating which of the five heroes is being controlled, detailed in \autoref{fig:observationlogic}.

Because of the expansive nature of the problem and the size and expense of each experiment, it was not practical to investigate all the details of the policy and training system.
Many details, even some large ones, were set for historical reasons or on the basis of preliminary investigations without full ablations.

\subsection{Optimizing the Policy}
Our goal is to find a policy which maximizes the probability of winning the game against professional human experts. 
In practice, we maximize a {\it reward function} which includes additional signals such as characters dying, collecting resources, etc. 
We also apply several techniques to exploit the zero-sum multiplayer structure of the problem when computing the reward function --- for example, we symmetrize rewards by subtracting the reward earned by the opposing team. 
We discuss the details of the reward function in \autoref{appendix:rewards}.
We constructed the reward function once at the start of the project based on team members' familiarity with the game. 
Although we made minor tweaks when game versions changed, we found that our initial choice of what to reward worked fairly well. 
The presence of these additional signals was important for successful training (as discussed in \autoref{appendix:rewards}).

\begin{figure}
\includegraphics[width=\textwidth]{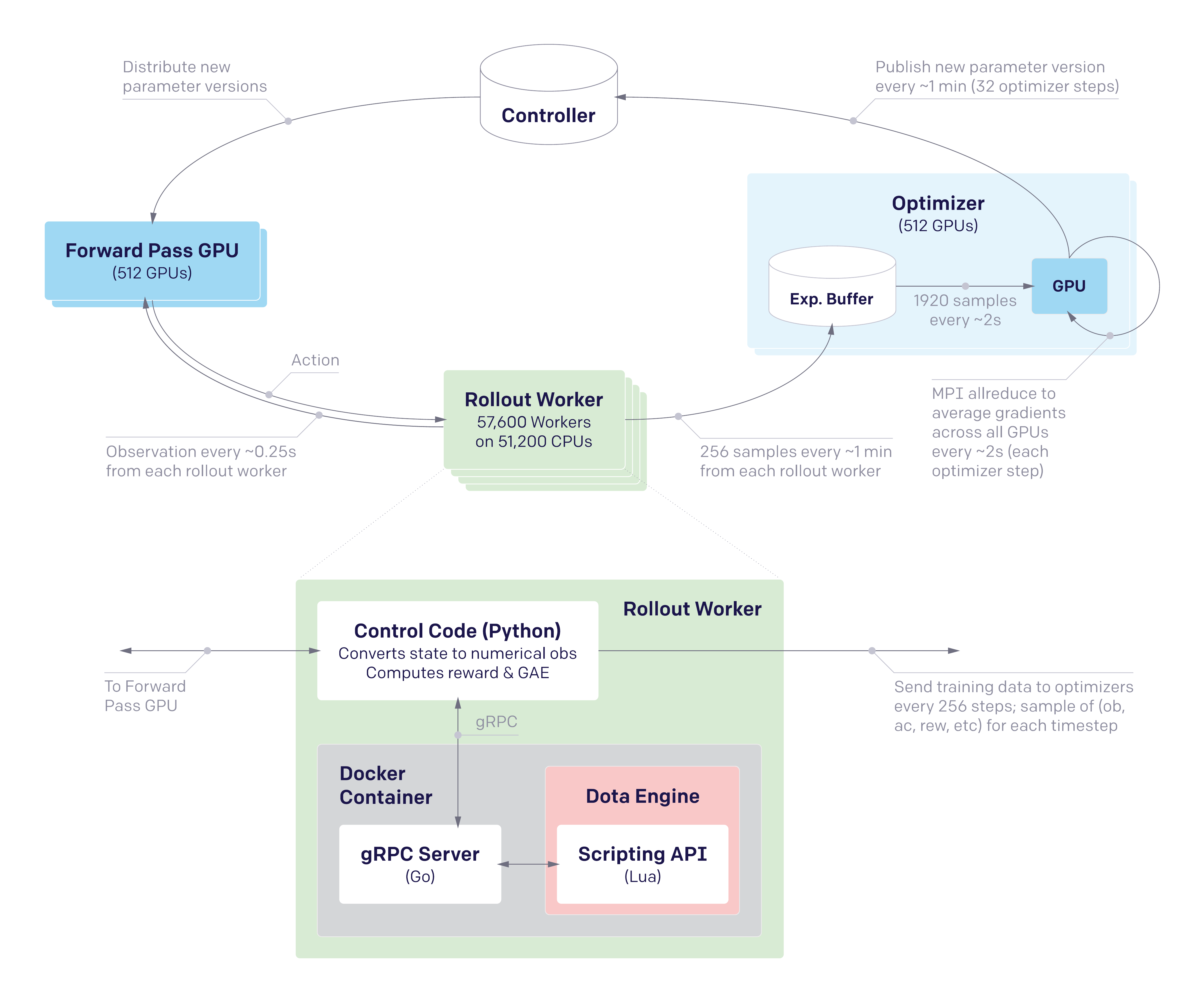}
\caption{\textbf{System Overview}: Our training system consists of 4 primary types of machines. Rollouts run the \dota game on CPUs. They communicate in a tight loop with Forward Pass GPUs, which sample actions from the policy given the current observation. Rollouts send their data to Optimizer GPUs, which perform gradient updates. The Optimizers publish the parameter versions to storage in the Controller, and the Forward Pass GPUs occasionally pull the latest parameter version. Machine numbers are for the \cleanexpname experiment described in \autoref{sec:rerun}; \openaifive's numbers fluctuated between this scale and approximately 3x larger.}
\label{fig:training-architecture}
\end{figure}

The policy is trained using Proximal Policy Optimization (PPO)\cite{schulman2017proximal}, a variant of advantage actor critic\cite{konda2000actor,mnih2016asynchronous}.\footnote{Early on in the project, we considered other algorithms including other policy gradient methods, q-learning, and evolutionary strategies. PPO was the first to show initial learning progress.}
The optimization algorithm uses Generalized Advantage Estimation \cite{schulman2016gae} (GAE), a standard advantage-based variance reduction technique \cite{konda2000actor} to stabilize and accelerate training. 
We train a network with a central, shared LSTM block, that feeds into separate fully connected layers producing policy and value function outputs.

The training system is represented in \autoref{fig:training-architecture}. 
We train our policy using collected self-play experience from playing \dota, similar to \cite{horgan2018apex}. 
A central pool of {\it optimizer GPUs} receives game data and stores it asynchronously in local buffers called {\it experience buffers}. 
Each optimizer GPU computes gradients using minibatches sampled randomly from its experience buffer. 
Gradients are averaged across the pool using NCCL2\cite{nccl} allreduce before being synchronously applied to the parameters.
In this way the effective batch size is the batch size on each GPU (120 samples, each with 16 timesteps) multiplied by the number of GPUs (up to 1536 at the peak), for a total batch size of 2,949,120 time steps (each with five hero policy replicas).

We apply the Adam optimizer \cite{kingma2014adam} using truncated backpropagation through time\cite{Williams90anefficient} over samples of 16 timesteps.
Gradients are additionally clipped per parameter to be within between $\pm 5 \sqrt{v}$ where $v$ is the running estimate of the second moment of the (unclipped) gradient.
Every 32 gradient steps, the optimizers publish a new {\it version} of the parameters to a central Redis\footnote{\url{http://redis.io}} storage called the {\it controller}.
The controller also stores all metadata about the state of the system, for stopping and restarting training runs.

``Rollout'' worker machines run self-play games.
They run these games at approximately 1/2 real time, because we found that we could run slightly more than twice as many games in parallel at this speed, increasing total throughput.
We describe our integration with the \dota engine in \autoref{appendix:gym-ocean}.
They play the latest policy against itself for 80\% of games, and play against older policies for 20\% of games (for details of opponent sampling, see \autoref{sec:selfplay}).
The rollout machines run the game engine but not the policy; they communicate with a separate pool of GPU machines which run forward passes in larger batches of approximately 60. 
These machines frequently poll the controller to gather the newest parameters.

Rollout machines send data asynchronously from games that are in progress, instead of waiting for an entire game to finish before publishing data for optimization\footnote{Rollout machines produce 7.5 steps per second; they send data every 256 steps, or 34 seconds of game play. 
Because our rollout games run at approximately half-speed, this means they push data approximately once per minute.}; see \autoref{fig:timescales} in \autoref{appendix:hyperparams} for more discussion of how rollout data is aggregated.
See \autoref{subfig:staleness} for the benefits of keeping the rollout-optimization loop tight.
Because we use GAE with $\lambda=0.95$, the GAE rewards need to be smoothed over a number of timesteps $\gg1/\lambda=20$; using 256 timesteps causes relatively little loss.

The entire system runs on our 
custom distributed training platform called Rapid\cite{dactyl}, running on Google Cloud Platform.
We use ops from the blocksparse library for fast GPU training\cite{gray2017blocksparse}.
For a full list of the hyperparameters used in training, see \autoref{appendix:hyperparams}.

\subsection{Continual Transfer via Surgery}\label{sec:surgery}
As the project progressed, our code and environment gradually changed for three different reasons:
\begin{enumerate}
    \item\label{item:surgey-change:architecture} As we experimented and learned, we implemented changes to the training process (reward structure, observations, etc) or even to the architecture of the policy neural network.     
    \item\label{item:surgey-change:new-stuff} Over time we expanded the set of game mechanics supported by the agent's action and observation spaces. These were not introduced gradually in an effort to build a perfect curriculum. Rather they were added incrementally as a consequence of following the standard engineering practice of building a system by starting simple and adding complexity piece by piece over time.    
    \item\label{item:surgey-change:version} From time to time, Valve publishes a new \dota version including changes to the core game mechanics and the properties of heroes, items, maps, etc; to compare to human players our agent must play on the latest game version.
\end{enumerate}
These changes can modify the shapes and sizes of the model's layers, the semantic meaning of categorical observation values, etc.

When these changes occur, most aspects of the old model are likely relevant in the new environment. 
But cherry-picking parts of the parameter vector to carry over is challenging and limits reproducibility. For these reasons training from scratch is the safe and common response to such changes.

However, training \openaifive was a multi-month process with high capital expenditure, motivating the need for methods that can persist models across domain and feature changes. 
It would have been prohibitive (in time and money) to train a fresh model to a high level of skill after each such change (approximately every two weeks).
For example, we changed to \dota version 7.21d, eight days before our match against the world champions (OG); this would not have been possible if we had not continued from the previous agent. 

Our approach, which we term ``surgery'', can be viewed as a collection of tools to perform offline operations to the old model $\pi_\theta$ to obtain a new model $\hat{\pi}_{\hat{\theta}}$ compatible with the new environment, which performs at the same level of skill even if the parameter vectors $\hat\theta$ and $\theta$ have different sizes and semantics. 
We then begin training in the new environment using $\hat\pi_{\hat\theta}$.
In the simplest case where the environment, observation, and action spaces did not change, our standard reduces to insisting that the new policy implements the same function from observed states to action probabilities as the old:
\begin{equation}\label{eqn:surgery}
    \forall o ~ \hat\pi_{\hat\theta}(o) = \pi_\theta(o)
\end{equation}
This case is a special case of {\it Net2Net}-style function preserving transformations \cite{chen15net2net}. 
We have developed tools to implement \autoref{eqn:surgery} exactly when possible (adding observations, expanding layers, and other situations), and approximately when the type of modification to the environment, observation space, or action space precludes satisfying it exactly. 
See \autoref{appendix:surgery} for further discussion of surgery.

In the end, we performed over twenty surgeries (along with many unsuccessful surgery attempts) over the ten-month lifetime of \openaifive (see \autoref{table:list-surgeries} in \autoref{appendix:surgery} for a full list). Surgery enabled continuous training without loss in performance (see \autoref{fig:rerun-graphs}). In \autoref{sec:rerun} we discuss our experimental verification of this method.

\section{Experiments and Evaluation}\label{sec:results}
\openaifive is a single training run that ran from June 30th, 2018 to April 22nd, 2019. 
After ten months of training using \totalcompute of compute, it defeated the \dota world champions in a best-of-three match and 99.4\% of human players during a multi-day online showcase.

In order to utilize this level of compute effectively we had to scale up along three axes. 
First, we used batch sizes of 1 to 3 million timesteps (grouped in unrolled LSTM windows of length 16). 
Second, we used a model with over 150 million parameters. 
Finally, OpenAI Five trained for 180 days (spread over 10 months of real time due to restarts and reverts).
Compared AlphaGo\cite{silver2016mastering}, we use 50 to 150 times larger batch size, 20 times larger model, and 25 times longer training time. 
Simultaneous works in recent months\cite{alphastarblog, unpublishedrobotics} have matched or slightly exceeded our scale.


\subsection{Human Evaluation}

Over the course of training, \openaifive played games against numerous amateur players, professional players, and professional teams in order to gauge progress. 
For a complete list of the professional teams \openaifive played against over time, see \autoref{appendix:humans}.

On April 13th, \openaifive played a high-profile game against OG, the reigning \dota world champions, winning a best-of-three (2-0) and demonstrating that our system can learn to play at the highest levels of skill.
For detailed analysis of our agent's performance during this game and its overall understanding of the environment, see \autoref{sec:results:understanding}.

Machine Learning systems often behave poorly when confronted with unexpected situations\cite{Dalvi:2004:AC:1014052.1014066}.
While winning a single high-stakes showmatch against the world champion indicates a very high level of skill, it does not prove a broad understanding of the variety of challenges the human community can present. 
To explore whether \openaifive could be consistently exploited by creative or out-of-distribution play, we ran {\it \openaiarena}, in which we opened \openaifive to the public for competitive online games from April 18-21, 2019. 
In total, Five played 3,193 teams in 7,257 total games, winning 99.4\%
\footnote{Human players often abandoned losing games rather than playing them to the end, even abandoning games right after an unfavorable hero selection draft before the main game begins. 
\openaifive does not abandon games, so we count abandoned games as wins for \openaifive. These abandoned games (3140 of the 7215 wins) likely includes a small number of games that were abandoned for technical or personal reasons.}.
Twenty-nine teams managed to defeat \openaifive for a total of $42$ games lost.

In \dota, the key measure of human dexterity is \emph{reaction time}\footnote{Contrast with RTS games like Starcraft, where the key measure is actions per minute due to the large number of units that need to be supplied with actions.}.
\openaifive can react to a game event in 217ms on average. 
This quantity does not vary depending on game state.
It is difficult to find reliable data on \dota professionals' reaction times, but typical human visual reaction time is approximately 250ms\cite{jain2015reaction}.
See \autoref{appendix:reaction-time} for more details.

\begin{figure}
    \centering
        \includegraphics[height=0.5\textwidth]{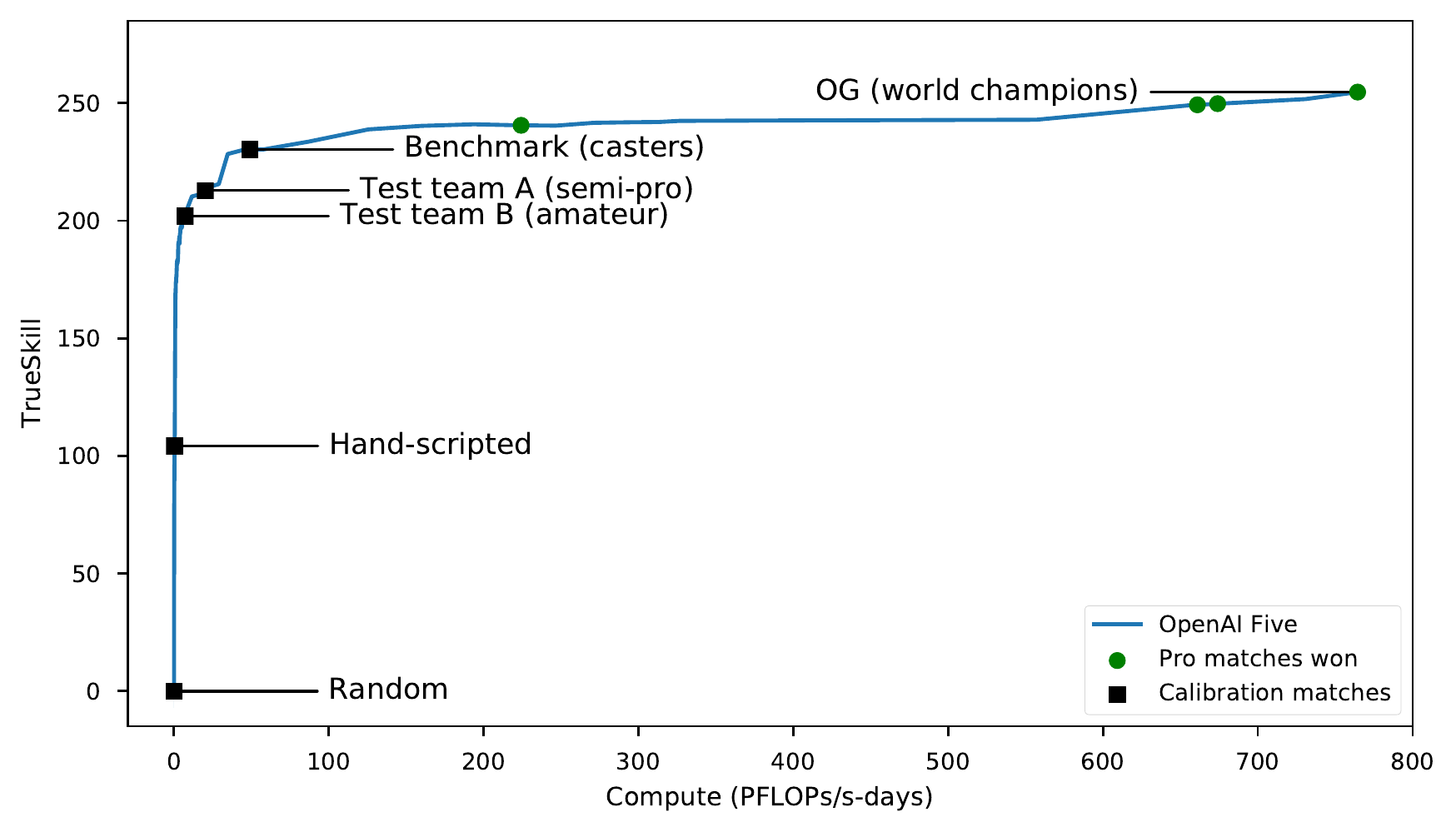}
    \caption{ \trueskillname over the course of training for \openaifive. To provide informal context for how \trueskillname corresponds to human skill, we mark the level at which OpenAI Five begins to defeat various opponents, from random to world champions.
    Note that this is biased against earlier models; this \trueskillname evaluation is performed using the final policy and environment (\dota version 7.21d, all non-illusion items, etc), even though earlier models were trained in the earlier environment.
    We believe this contributes to the inflection point around 600 PFLOPs/s-days --- around that point we gave the policy control of a new action (buyback) and performed a major \dota version upgrade (7.20).
    We speculate that the rapid increase to \trueskillname 200 early in training is due to the exponential nature of the scale --- a constant \trueskillname difference of approximately 8.3 corresponds to an 80\% winrate, and it is easier to learn how to consistently defeat bad agents.
    }
    \label{fig:final-ts}
\end{figure}

While human evaluation is the ultimate goal, we also need to evaluate our agents continually during training in an automated way. 
We achieve this by comparing them to a pool of fixed reference agents with known skill using the \trueskillname rating system \cite{herbrich2007trueskill}.
In our \trueskillname environment, a rating of 0 corresponds to a random agent, and a difference of approximately 8.3 \trueskillname between two agents roughly corresponds to an 80\% winrate of one versus the other (see \autoref{appendix:trueskill} for details of our \trueskillname setup). \openaifive's \trueskillname rating over time can be seen in \autoref{fig:final-ts}.

\openaifive's ``playstyle" is difficult to analyze rigorously (and is likely influenced by our shaped reward function) but we can discuss in broad terms the flavor of comments human players made to describe how our agent approached the game.
Over the course of training, \openaifive developed a distinct style of play with noticeable similarities and differences to human playstyles.
Early in training, \openaifive prioritized large group fights in the game as opposed to accumulating resources for later, which led to games where they were significantly behind if the enemy team avoided fights early.
This playstyle was risky and would result in quick wins in under 20 minutes if \openaifive got an early advantage, but had no way to recover from falling behind, leading to long and drawn out losses often over 45 minutes.

As the agents improved, the playstyle evolved to align closer with human play while still maintaining many of the characteristics learned early on. 
\openaifive began to concentrate resources in the hands of its strongest heroes, which is common in human play. Five relied heavily on large group battles, effectively applying pressure when holding a significant advantage, but also avoided fights and focused on gathering resources if behind.

The final agent played similar to humans in many broad areas, but had a few interesting differences.
Human players tend to assign heroes to different areas of the map and only reassign occasionally, but \openaifive moved heroes back and forth across the map much more frequently.
    Human players are often cautious when their hero has low health; \openaifive seemed to have a very finely-tuned understanding of when an aggressive attack with a low-health hero was worth a risk.
Finally \openaifive tended to more readily consume resources, as well as abilities with long {\it cooldowns} (time it takes to reload), while humans tend to hold on to those in case a better opportunity arises later. 

\subsection{Validating Surgery with \cleanexpname}\label{sec:rerun}
In order to validate the time and resources saved by our surgery method (see \autoref{sec:surgery}), we trained a second agent between May 18, 2019 and June 12, 2019, using only the final environment, model architecture, etc. 
This training run, called ``\cleanexpname'', did not go through a tortuous route of changing game rules, modifications to the neural network parameters, online experiments with hyperparameters, etc.

\begin{figure}
    \centering
    \includegraphics[scale=0.5]{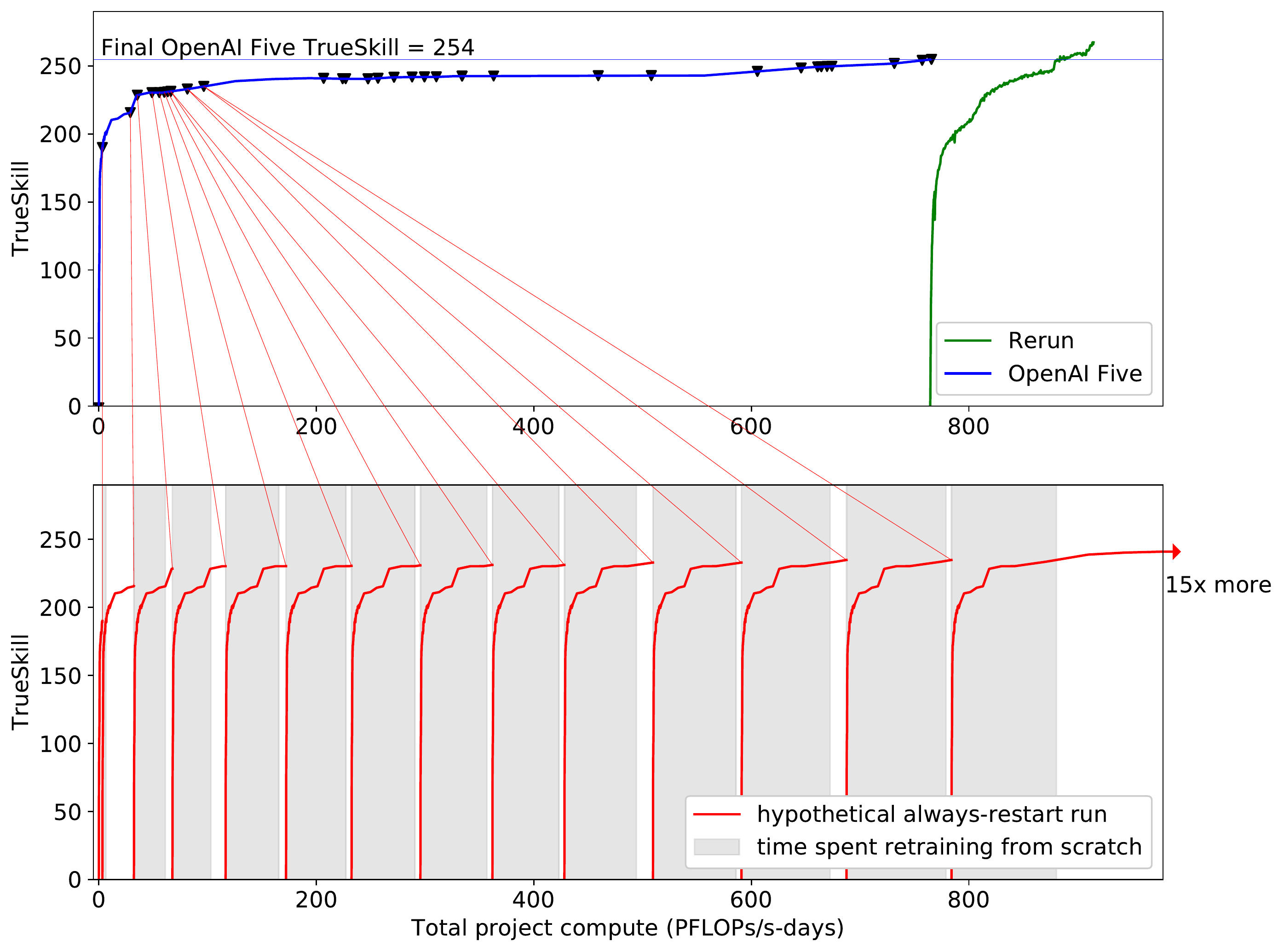}
    \caption{\textbf{Training in an environment under development:}
    In the top panel we see the full history of our project - we used surgery methods to continue training \openaifive at each environment or policy change without loss in performance; then we restarted once at the end to run \cleanexpname.
    On the bottom we see the hypothetical alternative, if we had restarted after each change and waited for the model to reach the same level of
    skill (assuming pessimistically that the curve would be identical to \openaifive).
    The ideal option would be to run \cleanexpname-like training from the very start, but this is impossible --- the \openaifive curve represents lessons learned that led to the final codebase, environment, etc., without which it would not be possible to train \cleanexpname.
    }
    \label{fig:rerun-graphs}
\end{figure}

\cleanexpname took 2 months and $150\pm5$ \pflopsday\ of compute (see \autoref{fig:rerun-graphs}).
This timeframe is significantly longer than the frequency of our surgery changes (which happened every 1-2 weeks). 
As a naive comparison, if we had trained from scratch after each of our twenty major surgeries, the project would have taken 40 months instead of 10 (in practice we likely would have made fewer changes).
Another benefit of surgery was that we had a very high-skill agent available for evaluation at all times, significantly tightening the iteration loop for experimental changes.
In \openaifive's regime --- exploring a novel task and building a novel environment --- perpetual training is a significant benefit.

Of course, in situations where the environment is pre-built and well-understood from the start, we see little need for surgery. 
\cleanexpname took approximately 20\% of the resources of \openaifive; if we had access to the final training environment ahead of time there would be no reason to start training e.g. on a different version of the game.

\cleanexpname continued to improve beyond \openaifive's skill, and reached over 98\% winrate against the final version of \openaifive.
We wanted to validate that our final code and hyperparameters would reproduce \openaifive performance, so we ceased training at that point. We believe \cleanexpname would have continued improving, both because of its upward trend and because we had yet to fully anneal hyperparameters like learning rate and horizon to their final \openaifive settings.

This process of surgery successfully allowed us to change the environment every week. 
However, the model ultimately plateaued at a weaker skill level than the from-scratch model was able to achieve. 
Learning how to continue long-running training without affecting final performance is a promising area for future work.

Ultimately, while surgery as currently conceived is far from perfect, with proper tooling it becomes a useful method for incorporating certain changes into long-running experiments without paying the cost of a restart for each.

\subsection{Batch Size}
In this section, we evaluate the benefits of increasing the batch size using small scale experiments.
Increasing the batch size in our case means two things: first, using twice as many optimizer GPUs to optimize over the larger batch, and second, using twice as many rollout machines and forward pass GPUs to produce twice as many samples to feed the increased optimizer pool.

One compelling benchmark to compare against when increasing the batch size
is {\it linear} speedup: using 2x as much compute gets to the same skill level in 1/2 the time. 
If this scaling property holds, it is possible to use the same {\it total} amount of GPU-days (and thus dollars) to reach a given result\cite{mccandlish2018empirical}.
In practice we see less than this ideal speedup, but the speedup from increasing batch size is still noticeable and allows us to reach the result in less wall time.

To understand how batch size affects training speed, we calculate the ``speedup'' of an experiment to reach various \trueskillname thresholds, defined as:
\begin{equation}\label{eqn:speedup}
    \textrm{speedup}(T) = \frac
    {\textrm{Versions for baseline to first reach \trueskillname $T$}}
    {\textrm{Versions for experiment to first reach \trueskillname $T$}}
\end{equation}

The results of varying batch size 
in the early part of training can be seen in \autoref{fig:lotsa-results}.
Full details of the experimental setup can be found in \autoref{appendix:substudies}. 
We find that increasing the batch size speeds up training through the regime we tested, up to batches of millions of observations.

Using the scale of \cleanexpname, we were able to reach superhuman performance in two months. 
In \autoref{subfig:batch-size}, we see that \cleanexpname's batch size (983k time steps) had a speedup factor of around 2.5x over the baseline batch size (123k). If we had instead used the smaller batch size, then, we might expect to wait 5 months for the same result. 
We speculate that it would likely be longer, as the speedup factor of 2.5 applies at \trueskillname 175 early in training, but it appears to increase with higher \trueskillname. 

Per results in \cite{mccandlish2018empirical}, we hoped to find (in the early part of training) linear
speedup from increasing batch size; i.e. that it would be 2x faster to train an agent to certain thresholds if we use 2x the compute and data.
Our results suggest that speedup is less than linear.
However, we speculate that this may change later in training when the problem becomes more difficult. 
Also, given the relevant compute costs, in this ablation study we did not tune hyperparameters such as learning rate separately for each batch size.

\subsection{Data Quality}

\begin{figure}
    \centering
    \begin{subfigure}[t]{0.31\linewidth}
        \includegraphics[height=\textwidth]{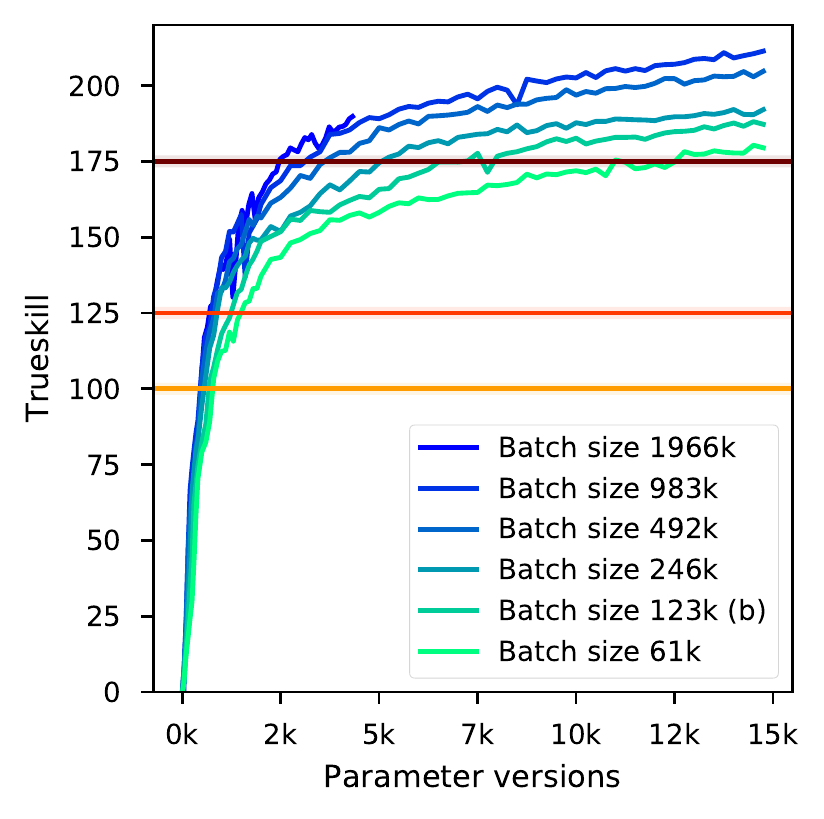}
        \includegraphics[height=\textwidth]{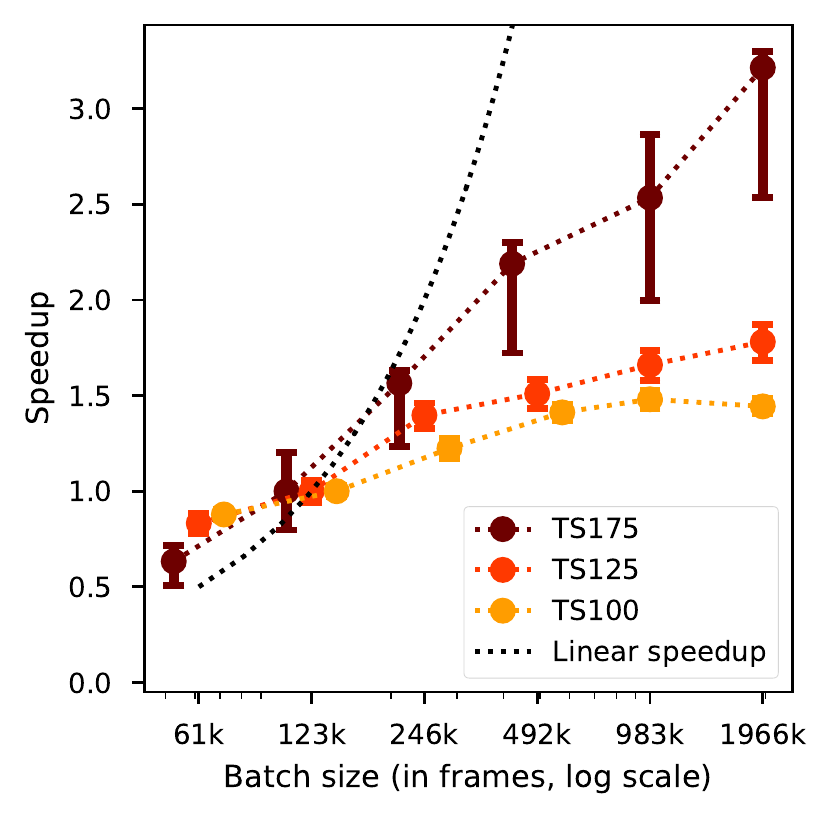}
        \caption{\label{subfig:batch-size}
        \textbf{Batch size:} Larger batch size speeds up training.
        In the early part of training studied here, the speedup is sublinear in the computation and samples required.
        See \autoref{sec:methods:batchsize} for experiment details.
        }
    \end{subfigure}
    \hspace{0.01\linewidth}
    \begin{subfigure}[t]{0.31\linewidth}
        \includegraphics[height=\textwidth]{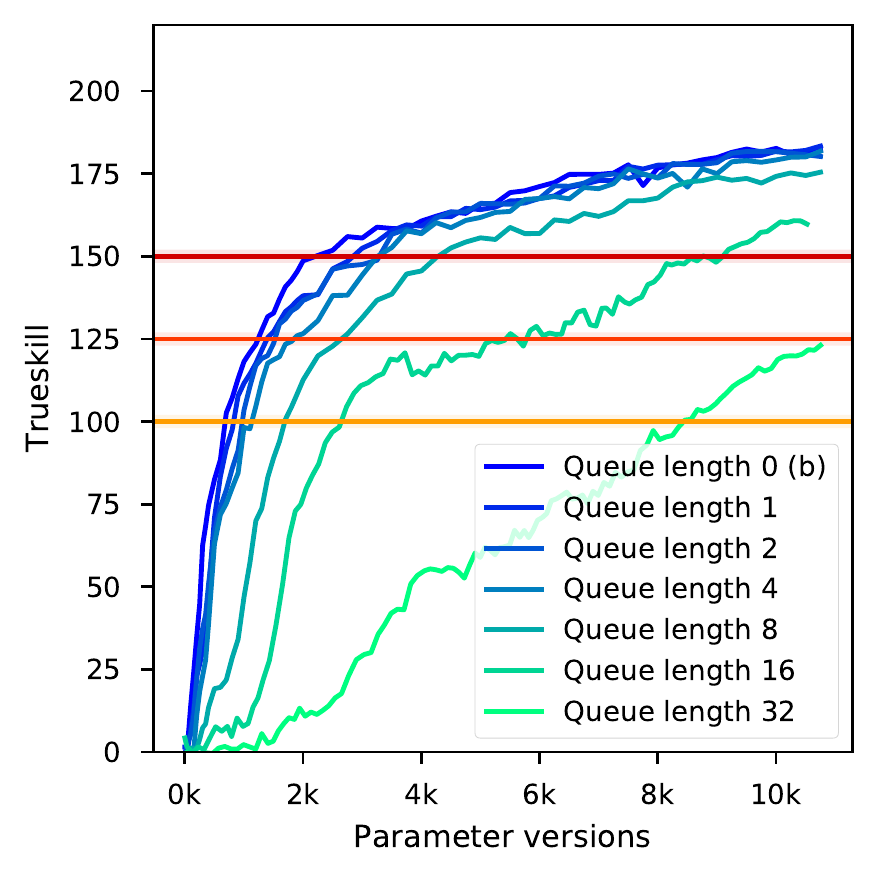}
        \includegraphics[height=\textwidth]{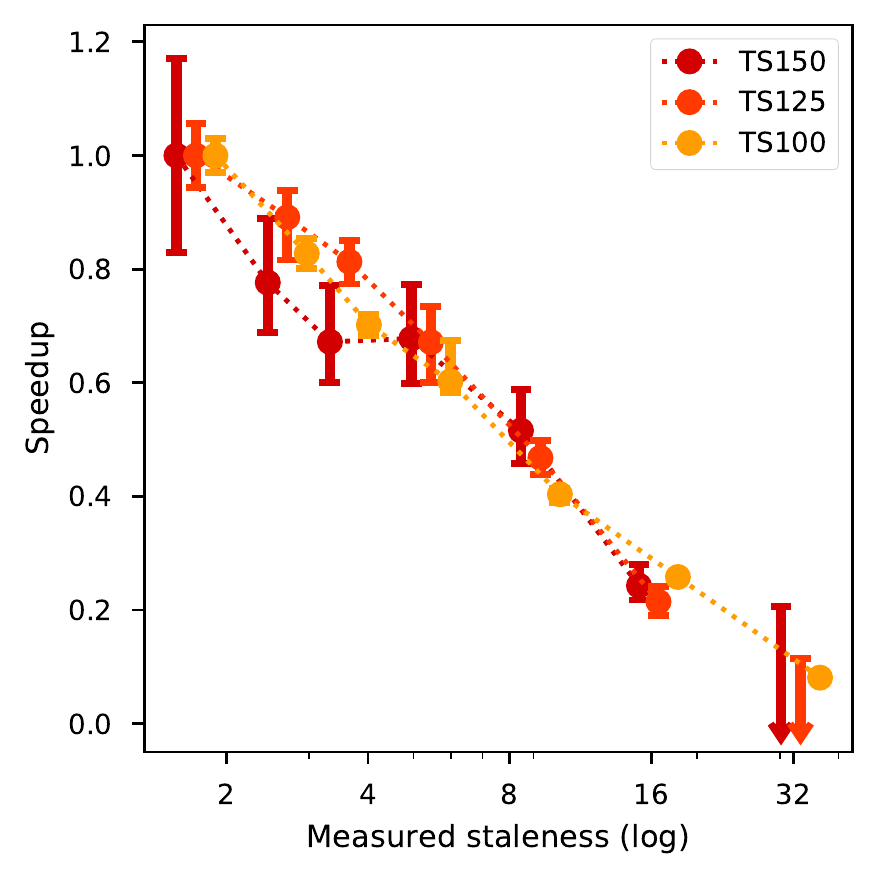}
        \caption{\label{subfig:staleness}
        \textbf{Data Staleness:} Training on stale rollout data causes significant losses in training speed.
        Queue length estimates the amount of artificial staleness introduced; see \autoref{sec:methods:staleness} for experiment details.
        }
    \end{subfigure}
    \hspace{0.01\linewidth}
    \begin{subfigure}[t]{0.31\linewidth}
        \includegraphics[height=\textwidth]{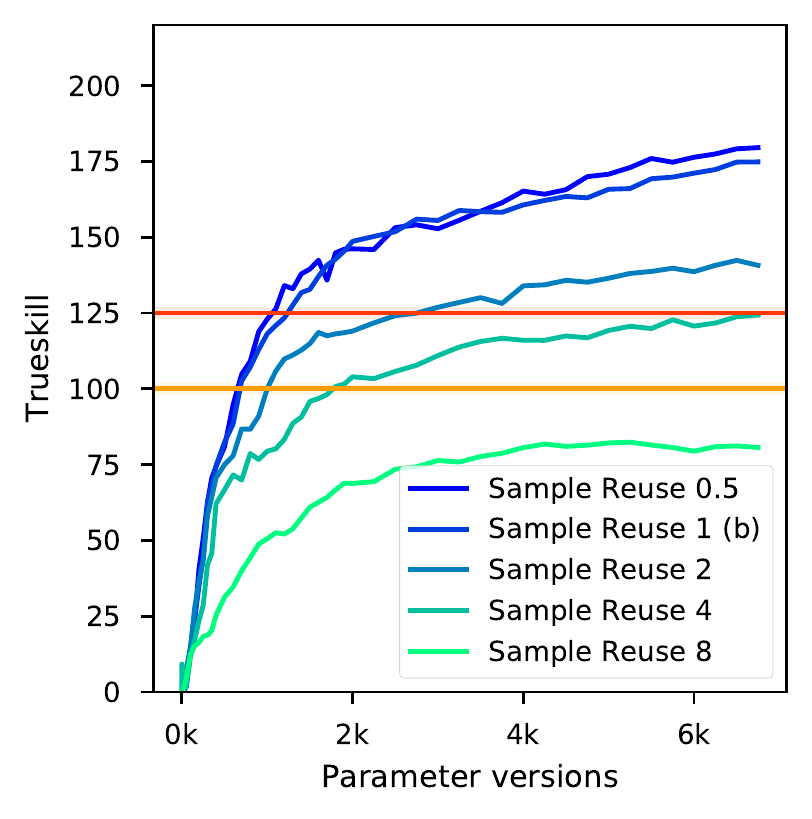}
        \includegraphics[height=\textwidth]{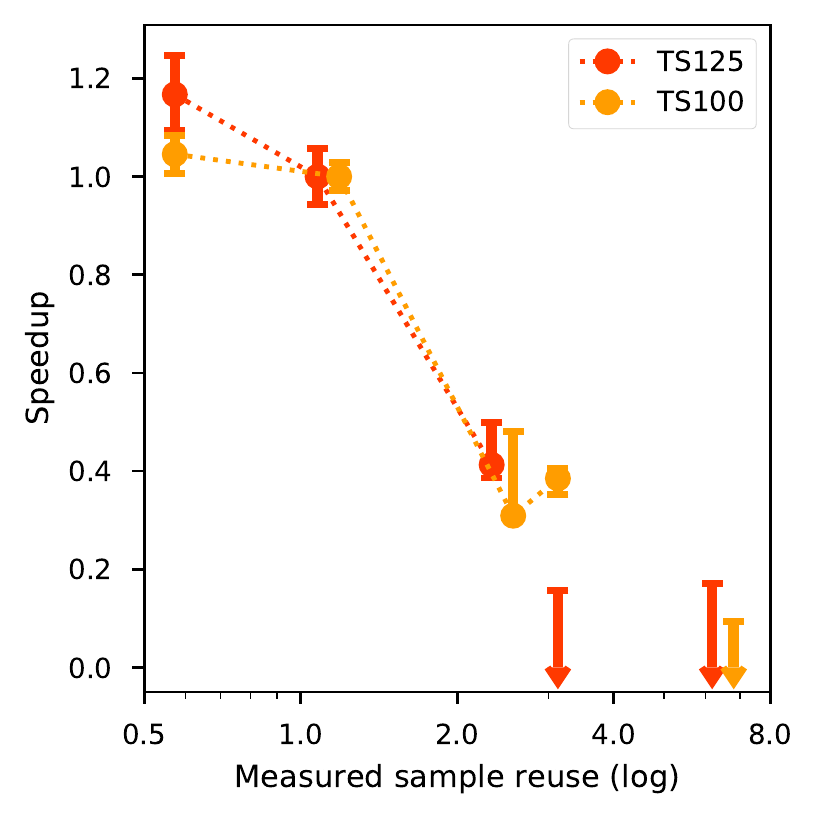}
        \caption{
        \label{subfig:sample-reuse}
        \textbf{Sample Reuse:} Reusing each sample of training data causes significant slowdowns. See \autoref{sec:methods:samplereuse} for experiment details.
        }
    \end{subfigure}
    \caption{\textbf{Batch Size and data quality in early training:} For each parameter, we ran multiple training runs varying only that parameter. These runs cover early training (approximately one week) at small scale (8x smaller than \cleanexpname). On the left we plot \trueskillname over time for each run. 
    On the right, we plot the ``speedup'' to reach fixed \trueskillname thresholds of 100, 125, 150, and 175 as a function of the parameter under study compared to the baseline (marked with `b'); see \autoref{eqn:speedup}. 
    Higher speedup means that training was faster and more efficient. 
    These four thresholds are chosen arbitrarily; a few are omitted when the uncertainties are too large (for example in \autoref{subfig:sample-reuse} fewer than half the experiments reach 175, so that speedup curve would not be informative).
    }
    \label{fig:lotsa-results}
\end{figure}

One unusual feature of our task is the length of the games; each rollout can take up to two hours to complete.
For this reason it is infeasible for us to optimize entirely on fully {\it on-policy} trajectories; 
if we waited to apply gradient updates for an entire rollout game to be played using the latest parameters, we could make only one update every two hours.
Instead, our rollout workers and optimizers operate asynchronously: rollout workers download the latest parameters, play a small portion of the game, and upload data to the experience buffer, while optimizers continually sample from whatever data is present in the experience buffer to optimize (\autoref{fig:training-architecture}).

Early on in the project, we had rollout workers collect full episodes before sending it to the optimizers and downloading new parameters. 
This means that once the data finally enters the optimizers, it can be several hours old, corresponding to thousands of gradient steps. 
Gradients computed from these old parameters were often useless or destructive.
In the final system rollout workers send data to optimizers after only 256 timesteps, but even so this can be a problem. 

We found it useful to define a metric for this called {\it staleness}. If a sample was generated by parameter version $N$ and we are now optimizing version $M$, then we define the {\it staleness} of that data to be $M-N$. 
In \autoref{subfig:staleness}, we see that increasing staleness by $\sim8$ versions causes significant slowdowns. 
Note that this level of staleness corresponds to a few minutes in a multi-month experiment.
Our final system design targeted a staleness between 0 and 1 by sending game data every 30 seconds of gameplay and updating to fresh parameters approximately once a minute, making the loop faster than the time it takes the optimizers to process a single batch (32 PPO gradient steps).
Because of the high impact of staleness, in future work it may be worth investigating whether optimization methods more robust to off-policy data could provide significant improvement in our asynchronous data collection regime.

Because optimizers sample from an experience buffer, the same piece of data can be re-used many times. 
If data is reused too often, it can lead to overfitting on the reused data\cite{horgan2018apex}.
To diagnose this, we defined a metric called the {\it sample reuse} of the experiment as the instantaneous ratio between the rate of optimizers consuming data and rollouts producing data. 
If optimizers are consuming samples twice as fast as rollouts are producing them, then on average each sample is being used twice and we say that the sample reuse is 2.
In \autoref{subfig:sample-reuse}, we see that reusing the same data even 2-3 times can cause a factor of two slowdown, and reusing it 8 times may prevent the learning of a competent policy altogether.
Our final system targets sample reuse $\sim1$ in all our experiments.

These experiments on the early part of training indicate that high quality data matters even more than compute consumed; small degradations in data quality have severe effects on learning.
Full details of the experiment setup can be found in \autoref{appendix:substudies}. 

\subsection{Long term credit assignment}
\begin{figure}
    \centering
    \begin{subfigure}[t]{0.5\linewidth}
        \centering  
        \includegraphics[width=0.9\textwidth]{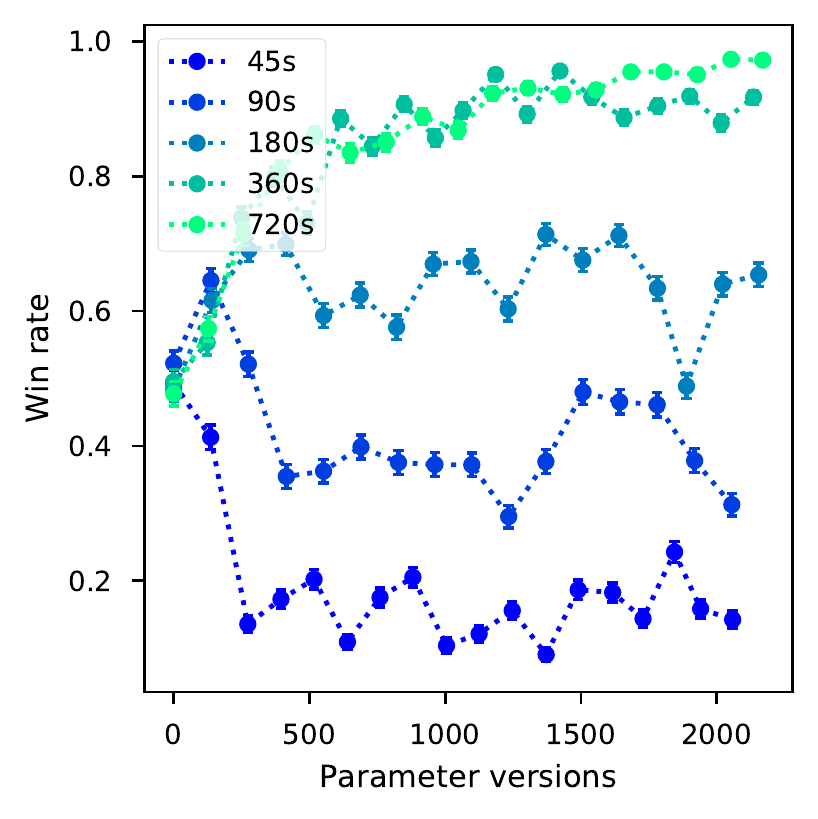}
    \end{subfigure}
    \caption{\textbf{Effect of horizon on agent performance.} We resume training from a trained agent using different horizons (we expect long-horizon planning to be present in highly-skilled agents, but not from-scratch agents). The base agent was trained with a horizon of 180 seconds ($\gamma=0.9993$), and we include as a baseline continued training at horizon 180s. Increasing horizon increases win rate over the trained agent at the point training was resumed, with diminishing returns at high horizons.}
    \label{fig:horizon}
\end{figure}
\dota has extremely long time dependencies. Where many reinforcement learning environment episodes last hundreds of steps (\cite{silver2016mastering, cobbe2018coinrun, jaderberg2018human, moravvcik2017deepstack}), games of \dota can last for tens of thousands of time steps. 
Agents must execute plans that play out over many minutes, corresponding to thousands of timesteps.
This makes our experiment a unique platform to test the ability of these algorithms to understand long-term credit assignment.

In \autoref{fig:horizon}, we study the time horizon over which our agent discounts rewards, defined as 
\begin{equation}\label{eqn:horizon}
    H = \frac{T}{1-\gamma}
\end{equation}
Here $\gamma$ is the discount factor \cite{schulman2016gae} and $T$ is the real game time corresponding to each step (0.133 seconds).
This measures the game time over which future rewards are integrated, and we use it as a proxy for the long-term credit assignment which the agent can perform. 

In \autoref{fig:horizon}, we see that resuming training a skilled agent using a longer horizon makes it perform better, up to the longest horizons we explored (6-12 minutes). 
This implies that our optimization was capable of accurately assigning credit over long time scales, and capable of learning policies and actions which maximize rewards 6-12 minutes into the future.
As the environments we attempt to solve grow in complexity, long-term planning and thinking will become more and more important for intelligent behavior. 

\section{Related Work}
\nicetohave{discuss other families of algorithms, early q-learning attempts}
The \openaifive system builds upon several bodies of work combining deep reinforcement learning, large-scale optimization of deep learning models, and using self-play to explore environments and strategies.

Competitive games have long served as a testbed for learning. Early systems mastered Backgammon~\cite{tesauro1994td}, Checkers~\cite{SCHAEFFER1992273checkers}, and Chess~\cite{campbell2002deepblue}.
Self-play was shown to be a powerful algorithm for learning skills within high-dimensional continuous environments~\cite{bansal2017emergent} and a method for automatically generating curricula~\cite{sukhbaatar2017intrinsic}. 
Our use of self-play is similar in spirit to fictitious play~\cite{brown:fp1951}, which has been successfully applied to poker~\cite{DBLP:journals/corr/HeinrichS16} - in this work we learn a distribution over opponents and use the latest policy rather than an average policy.

Using a combination of imitation learning human games and self-play, Silver et al. demonstrated a master-level Go player~\cite{silver2016mastering}. 
Building upon this work, AlphaGoZero, AlphaZero, and ExIt discard imitation learning in favor of using Monte-Carlo Tree Search during training to obtain higher quality trajectories~\cite{silver2017mastering,anthony2017thinking,silver2018general} and apply this to Go, Chess, Shogi, and Hex. 
Most recently, human-level play has been demonstrated in 3D first-person multi-player environments \cite{jaderberg2018human}, professional-level play in the real-time strategy game StarCraft~2 using AlphaStar~\cite{alphastarblog}, and superhuman performance in Poker \cite{brown2019superhuman}.

AlphaStar is particularly relevant to this paper.
In that effort, which ran concurrently to our own, researchers trained agents to play Starcraft 2, another complex game with real-time performance requirements, imperfect information, and long time horizons.
The model for AlphaStar used a similar hand-designed architecture to embed observations and an autoregressive action decoder, with an LSTM core to handle partial observability. 
Both systems used actor critic reinforcement learning methods as part of the overall objective.
\openaifive has certain sub-systems hard-coded (such as item buying), whereas AlphaStar handled similar decisions (e.g. building order) by conditioning (during training) on statistics derived from human replays.
\openaifive trained using self play, while AlphaStar used a league consisting of multiple agents, where agents were trained to beat certain subsets of other agents.
Finally, AlphaStar's value network observed full information about the game state (including observations hidden from the policy); this method improved their training and exploring its application to \dota is a promising direction for future work. 

Deep reinforcement learning has been successfully applied to learning control policies from high dimensional input. 
In 2013, Mnih et al.\cite{mnih2013playing} show that it is possible to combine a deep convolutional neural network with a Q-learning algorithm\cite{watkins1992q} and a novel experience replay approach to learn policies that can reach superhuman performance on the Atari ALE games. 
Following this work, a variety of efforts have pushed performance on the remaining Atari games\cite{mnih2016asynchronous}, reduced the sample complexity, and introduced new challenges by focusing on intrinsic rewards \cite{kulkarni2016hierarchical,burda2018exploration,ecoffet2018montezuma}.

As more computational resources have become available, a body of work has developed addressing the use of distributed systems in training.
Larger batch sizes were found to accelerate training of image models\cite{Goyal2017Imagenet, You2017Imagenet, You2018minutenet}.
Proximal Policy Optimization\cite{schulman2017proximal} and
A3C \cite{mnih2016a3c} improve the ability to asynchronously collect rollout data. 
Recent work has demonstrated the benefit of distributed learning on a wide array of problems including single-player video games\cite{espeholt2018impala} and robotics\cite{dactyl}.

The motivation for our surgery method is similar to prior work on {\it Net2Net} style function preserving transformations \cite{chen15net2net} which attempt to add model capacity without compromising performance, whereas our surgery technique was used in cases where the inputs, outputs, and recurrent layer size changed.
Past methods have grown neural networks by incrementally training and freezing parts of the network \cite{fahlman90cascade}, \cite{wang17growing}, \cite{DBLP:journals/corr/abs-1806-01780}.
\citet{li18learningwithoutforgetting} and \citet{rusu16progressive} use similar methods to use a trained model to quickly learn novel tasks.
Distillation \cite{hinton15distilling} and imitation learning \cite{ross11dagger, levine13gps} offer an alternate approach to surgery for making model changes in response to a shifting environment. In concurrent work, \citet{unpublishedrobotics} has reported success using behavioral cloning for similar purposes.

\section{Conclusion}
When successfully scaled up, modern reinforcement learning techniques can achieve superhuman performance in competitive esports games.
The key ingredients are to expand the scale of compute used, by increasing the batch size and total training time.
In order to extend the training time of a single run to ten months, we developed surgery techniques for continuing training across changes to the model and environment.
While we focused on \dota, we hypothesize these results will apply more generally and these methods can solve any zero-sum two-team continuous environment which can be simulated in parallel across hundreds of thousands of instances.
In the future, environments and tasks will continue to grow in complexity. 
Scaling will become even more important (for current methods) as the tasks become more challenging.

\newpage
\section*{Acknowledgements}
Myriad individuals contributed to this work from within OpenAI, within the \dota community, and elsewhere. 
We extend our utmost gratitude to everyone who helped us along the way! 
We would like to especially recognize the following contributions:
\begin{itemize}
\item Technical discussions with numerous people within OpenAI including Bowen Baker, Paul Christiano, Danny Hernandez, Sam McCandlish, Alec Radford
\item Review of early drafts by Bowen Baker, Danny Hernandez, Jacob Hilton, Quoc Le, Luke Metz, Matthias Plappert, Alec Radford, Oriol Vinyals 
\item Event Support from Larissa Schiavo, Diane Yoon, Loren Kwan
\item Communication, writing, and outreach support from Ben Barry, Justin Wang, Shan Carter, Ashley Pilipiszyn, Jack Clark
\item OpenAI infrastructure support from Eric Sigler 
\item Google Cloud Support (Solomon Boulos, JS Riehl, Florent de Goriainoff, Somnath Roy, Winston Lee, Andrew Sallaway, Danny Hammo, Jignesh Naik)
\item Microsoft Azure Support (Jack Kabat, Jason Vallery, Niel Mackenzie, David Kalmin, Dina Frandsen)
\item \dota Support from Valve (special thanks to Chris Carollo)
\item \dota guides and builds from Tortedelini (Michael Cohen) and buyback saving strategy from Adam Michalik
\item \dota expertise and community advice from Blitz (William Lee)
\item \dota Casters: Blitz (William Lee), Capitalist (Austin Walsh), Purge (Kevin Godec), ODPixel (Owen Davies), Sheever (Jorien van der Heijden), Kyle Freedman
\item \dota World Champions (OG): ana (Anathan Pham), Topson (Topias Taavitsainen), Ceb (S\'{e}bastien Debs), JerAx (Jesse Vainikka), N0tail (Johan Sundstein)
\item \dota Professional Teams: Team Secret, Team Lithium, Alliance, SG E-sports
\item Benchmark Players: Moonmeander (David Tan), Merlini (Ben Wu), Capitalist (Austin Walsh), Fogged (Ioannis Loucas), Blitz (William Lee)
\item Playtesting: 
Alec Radford, Bowen Baker, Alex Botev, 
Pedja Marinkovic, 
Devin McGarry, Ryan Perron, Garrett Fisher, Jordan Beeli, Aaron Wasnich,  David Park, Connor Mason, James Timothy Herron, Austin Hamilton, Kieran Wasylyshyn,
Jakob Roedel, William Rice, Joel Olazagasti, 
Samuel Anderson
\item We thank the entire \dota community for their support and enthusiasm. We especially profusely thank all 39,356 \dota players from 225 countries who participated in \openaiarena and all the players who played against the 1v1 agent during the LAN event at The International 2017!
\end{itemize}

\section*{Author Contributions}
This manuscript is the result of the work of the entire OpenAI Dota team. For each major area, we list the primary contributors in alphabetical order.
\begin{itemize}
    \item Greg~Brockman, Brooke~Chan, Przemys\l{}aw~``Psyho"~D\k{e}biak, Christy~Dennison, David~Farhi, Scott~Gray, Jakub~Pachocki, Michael~Petrov, Henrique~Pond\'e de Oliveira Pinto, Jonathan~Raiman, Szymon~Sidor, Jie~Tang, Filip~Wolski, and Susan~Zhang developed and trained OpenAI Five, including developing surgery, expanding tools for large-scale distributed RL, expanding the capabilities to the 5v5 game, and running benchmarks against humans including the OG match and OpenAI Arena.
    \item Christopher~Berner, Greg~Brockman, Vicki~Cheung, Przemys\l{}aw~``Psyho"~D\k{e}biak, Quirin~Fischer, Shariq~Hashme, Chris~Hesse, Rafal J\'{o}zefowicz, Catherine~Olsson, Jakub~Pachocki, Tim~Salimans, Jeremy~Schlatter, Jonas~Schneider, Szymon~Sidor, Ilya~Sutskever, and Jie~Tang developed the 1v1 training system, including the Dota 2 gym interface, building the first Dota agent, and initial exploration of batch size scaling.
    \item Brooke~Chan, David~Farhi, Michael~Petrov, Henrique~Pond\'e de Oliveira Pinto, Jonathan~Raiman, Jie~Tang, and Filip~Wolski wrote this manuscript, including running \cleanexpname and all of the ablation studies.
    \item Jakub~Pachocki and Szymon~Sidor set research direction throughout the project, including developing the first version of Rapid to demonstrate initial benefits of large scale computing in RL.
    \item Greg Brockman and Rafal J\'{o}zefowicz kickstarted the team. 
\end{itemize}

\newpage
\printbibliography

\appendix
\newpage
\part{Appendix} 
\parttoc 
\newpage

\section{Compute Usage}\label{sec:compute}
We estimate the optimization compute usage as follows: We break the experiment in segments between each major surgery or batch size change. For each of those, we calculate the number of gradient descent steps taken ($\textrm{number of iterations} \times 32$). We estimate the compute per step per GPU using TensorFlow's \texttt{tf.profiler.total\_float\_ops}, then multiply together:
\begin{align}
\textrm{total compute} = \sum_{\textrm{segment}} & 32 \times (\textrm{iteration}_\textrm{end} - \textrm{iteration}_\textrm{start}) \times \\
& \textrm{(\# gpus)} \times \textrm{(compute per step per gpu)}
\end{align}

Our uncertainty on this estimate comes primarily from ambiguities about what computation ``counts.''
For example the tensorflow metrics include all ops in the graph including metric logging, nan-checking, etc. 
It also includes the prediction of auxiliary heads such as win probability, which are not necessary for gameplay or training.
It does not count non-GPU compute on the optimizer machines such as exporting parameter versions to the rollouts.
We estimate these and other ambiguities to be around 5\%. 
In addition, for \openaifive (although not for \cleanexpname) we use a simplified history of the experiment, rather than keeping track of every change and every time something crashed and needed to be restarted; we estimate this does not add more than 5\% error. 
We combine these rough error estimates into a (very crude) net ambiguity estimate of 5-10\%.

This computation concludes that \openaifive used \totalcompute of total optimization compute on GPUs at the time of playing the world champions (April 13, 2019), and 820$\pm$50 total optimization compute when it was finally turned off on April 22nd, 2019. 
\cleanexpname, on the other hand, used $150\pm5$ \pflopsday 
 between May 18th and July 12th, 2019.

We adopted the methodology from \cite{openaicomputeblog} to facilitate comparisons. This has several important caveats. 
First, the above computation only considers compute used for optimization.
In fact this is a relatively small portion of the total compute budget for the training run.
In addition to the GPU machines doing optimization (roughly 30\% of the cost by dollars spent) 
there are approximately the same number of GPUs running forward passes for the rollout workers (30\%), 
as well as the actual rollouts CPUs running the selfplay games (30\%) 
and the overhead of controllers, \trueskillname evaluators, CPUs on the GPU machines, etc (10\%). 

Second, with any research project one needs to run many small studies, ablations, false starts, etc. 
One also inevitably wastes some computing resources due to imperfect utilization.
Traditionally the AI community has not counted these towards the compute used by the project, as it is much easier to count only the resources used by the final training run. 
However, with our advent of surgery, the line becomes much fuzzier.
After 5 months of training on an older environment, we \emph{could} have chosen to start from scratch in the new environment, or performed surgery to keep the old model.
Either way, the same total amount of compute gets used; but the above calculation ignores all the compute used up until the last time we chose to restart.
For these reasons the compute number for \openaifive should be taken with a large grain of salt, but this caveat does not apply to \cleanexpname, which was trained without surgery.

\section{Surgery}\label{appendix:surgery}
\begin{table}
\centering
\begin{tabular}{cccp{9cm}}
\label{allsurgeries} 
Date & Iteration & \# params & Change\\
\hline
6/30/2018 & 1 & 43,436,520 & Experiment started \\ 
8/17/2018 & 81,821 & 43,559,322 & \dota version 7.19 adds new items, abilities, etc.\\ 
8/18/2018 & 84,432 & 43,805,274 & Change environment to single courier;\newline remove ``cheating'' observations \\ 
8/26/2018 & 91,471 & 156,737,674 & Double LSTM size \\ 
9/27/2018 & 123,821 & 156,809,485 & Support for more heroes \\ 
10/3/2018 & 130,921 & 156,809,501 & Obs: Roshan spawn timing \\ 
10/12/2018 & 140,402 & 156,811,805 & Item: Bottle\\ 
10/19/2018 & 144,121 & 156,286,925 & Obs: Stock counts;\newline Obs: Remove some obsolete obs \\ 
10/24/2018 & 150,111 & 156,286,867 & Obs: Neutral creep \& rune spawn timers \\ 
11/7/2018 & 161,482 & 156,221,309 & Obs: Item swap cooldown;\newline Obs: Remove some obsolete obs\\ 
11/28/2018 & 185,749 & 156,221,669 & Item: Divine rapier;\newline Obs: Improve observation of stale enemy heroes \\ 
12/10/2018 & 193,701 & 157,378,165 & Obs: Modifiers on nonhero units. \\ 
12/14/2018 & 196,800 & 157,650,795 & Action: Consumables on allies;\newline Obs: Line of sight information;\newline Obs: next item this hero will purchase;\newline Action: buyback \\ 
12/20/2018 & 203,241 & 157,679,655 & \dota version 7.20 adds new items, new item slot, changes map, etc;\newline Obs: number of empty inventory slots \\ 
1/23/2019 & 211,191 & 158,495,991 & Obs: Improve observations of area of effects;\newline Obs: improve observation of modifiers' duration;\newline Obs: Improve observations about item Power Treads. \\ 
4/5/2019 & 220,076 & 158,502,815 & \dota version 7.21 adds new items, abilities, etc. \\ 
\end{tabular}
\caption{All successful surgeries and major environment changes performed during the training of \openaifive. This table does not include surgeries which were ultimately reverted due to training failures, nor minor environment changes (such as improvements to partial reward weights or scripted logic). ``Obs'' indicates than a new observation was added as an input to the model or an existing one was changed. ``Action'' indicates that a new game action was made available, along with appropriate observations about the state of that action. ``Item'' indicates that a new item was introduced, including observation of the item and the action to use the item. The \dota version updates (7.19, 7.20 and 7.21) include many new items, actions, and observations.}
\label{table:list-surgeries}
\end{table}

As discussed in \ref{sec:surgery}, we designed ``surgery'' tools for continuing to train a single set of paramters across changes to the environment, model architecture, observation space, and action space. 
The goal in each case is to resume training after the change without the agent losing any skill from the change. 
\autoref{table:list-surgeries} lists the major surgeries we performed in the lifetime of the \openaifive experiment.

For changes which add parameters, one of the key questions to ask is how to initialize the new parameters. If we initialize the parameters randomly and continue optimization, then noise will flow into other parts of the model, causing the model to play badly and causing large gradients which destroy the learned behaviors. 

In the rest of this appendix we provide details of the tools we used to continue training across each type of change. 
In general we had a high-skill model $\pi_\theta$ trained to act in one environment, and due to a change to the problem design we need to begin training a newly-shaped model $\hat{\pi}_{\hat\theta}$ in a new environment. 
Ultimately the goal is for the \trueskillname of agent $\hat{\pi}_{\hat\theta}$ to match that of $\pi_\theta$.

\paragraph{Changing the architecture}
In the most straightforward situation, the observation space, action space, and environment do not change.
In this case, per \autoref{eqn:surgery}, we can insist that the new policy $\hat\pi_{\hat\theta}$ implement exactly the same mathematical function from observations to actions as the old policy.

A simple example here would be adding more units to an internal fully-connected layer of the model. Suppose that before the change, some part of the interior of the model contained an input vector $x$ (dimension $d_x$), which is transformed to an activation vector $y=W_1x+B_1$ (dimension $d_y$), which is then consumed by another fully-connected layer $z=W_2y+B_2$ (dimension $d_z$). We desire to increase the dimension of of $y$ from $d_y$ to $\hat{d}_y$. This causes the shapes of three parameter arrays to change: $W_1$ (from $[d_x, d_y]$ to $[d_x, \hat{d}_y]$), $B_1$ (from $[d_y]$ to $[\hat{d}_y]$), and $W_2$ (from $[d_y, d_z]$ to $[\hat{d}_y, d_z]$).

In this case we initialize the new variables in the first layer as:
\begin{equation}\label{eqn:surgery-example}
\hat{W}_1 = \left[
\begin{array}{c}
W_1 \\ R()
\end{array}
\right]
\hspace{2cm}
\hat{B}_1 = \left[
\begin{array}{c}
B_1 \\ R()
\end{array}
\right]
\hspace{2cm}
\hat{W}_2 = \left[
\begin{array}{cc}
W_2 & 0
\end{array}
\right]
\end{equation}

Where $R()$ indicates a random initialization. 
The initializations of $\hat{W}_1$ and $\hat{B}_1$ ensure that the first $d_y$ dimensions of activations $\hat{y}$ will be the same data as the old activations $y$, and the remained will be randomized. 
The randomization ensures that symmetry is broken among the new dimensions.
The initialization of $\hat{W}_2$, on the other hand, ensures that the next layer will ignore the new random activations, and the next layer's activations will be the same as in the old model; $\hat{z}=z$.
The weights which are initialized to zero will move away from zero due to the gradients, if the corresponding new dimensions in $y$ are useful to the downstream function.

Initializing neural network weights to zero is a dangerous business, because it can introduce undesired symmetries between the indices of the output vector. 
However we found that in most cases of interest, this was easy to avoid by only zero-ing the minimal set of weights. 
In the example above, the symmetry is broken by the randomization of $\hat{W}_1$ and $\hat{B}_1$.

A more advanced version of this surgery was required when we wanted to increase the model capacity dramatically, by increasing the hidden dimension of our LSTM from 2048 units to 4096 units.
Because the LSTM state is recurrent, there was no way to achieve the separation present in \autoref{eqn:surgery-example}; if we randomize the new weights they will impact performance, but if we set them to zero then the new hidden dimensions will be symmetric and gradient updates will never differentiate them. 
In practice we set the new weights to random small values --- rather than randomize new weight values on the same order of magnitude as the existing weights, we randomized new weights significantly smaller. 
The scale of randomization was set empirically by choosing the highest scale which did not noticeably decrease the agent's \trueskillname.

\paragraph{Changing the Observation Space}
Most of our surgeries caused the observation space changes, for example when we added 3 new float observations encoding the time until neutral creeps, bounties, and runes would spawn.
In these cases it is impossible to insist that the new policy implement the same function from observation space to action space, as the input domain has changed.
However, in some sense the input domain has {\it not} changed; the game state is still the same.
In reality our system is not only a function $\pi:o\to a$; before the policy sees the observation arrays, an ``encoder'' function $E$ has turned a game state $s$ into an input array $o$:
\begin{equation}
    \left(
    \textrm{Game State Protobuf } s
    \right)
    \xrightarrow{E}
    \left(
    \textrm{Observation Arrays } o
    \right)
    \xrightarrow{\pi}
    \left(
    \textrm{Action } a
    \right)
\end{equation}

By adding new observations we are enhancing the encoder function $E$, making it take the same game state and simply output richer arrays for the model to consume.
Thus in this case while we cannot ensure that $\hat\pi_{\hat\theta} = \pi_\theta$, we can ensure the functions are identical if we go one step back:
\begin{equation}
    \forall s \hspace{1em}
    \hat\pi_{\hat\theta}(\hat{E}(s))
    = \pi_\theta(E(s))
\end{equation}

When the change is simply additive, this can then be enforced as in the previous section. Suppose the new observations extend a vector $x$ from dimension $d_x$ to dimension $\hat{d}_x$, and the input vector $x$ is consumed by a weight matrix $W$ via $y=Wx$ (and $y$ is then processed by the rest of the model downstream). Then we initialize the new weights $\hat W$ as:

\begin{equation}
\hat{W} = \left[
\begin{array}{cc}
W & 0
\end{array}
\right]
\end{equation}

As before, this ensures that the rest of the model is unchanged, as the output is unchanged ($\hat y=y$). 
The weights which are initialized to zero will move away from zero due to the gradients, if the corresponding observations are found to be useful.

\paragraph{Changing the Environment or Action Space}
The second broad class of changes are those which change the environment itself, either by making new actions available to the policy (e.g. when we replaced scripted logic for the Buyback action with model-controlled logic) or by simply changing the \dota rules (for example when we moved to \dota version 7.21, or when we added new items). For some of these changes, such as upgrading \dota version, we found simply making the change on the rollout workers to be relatively stable; the old policy played well enough in the new environment that it was able to smoothly adapt.

Even so, whenever possible, we attempted to ``anneal'' in these new features, starting with 0\% of rollout games played with the new environment or actions, and slowly ramping up to 100\%.
This prevents a common problem where a change in one part of the agent's behavior could force unnecessary relearning large portions of the strategy. 
For example, when we attempted to give the model control of the Buyback action without annealing, the model-based control of the action was (at first) worse than the scripted version had been, causing the agent to adapt its overall strategies to games where allies and enemies alike often misuse this action.
This would cause the agent to significantly drop in overall skill; while it would likely eventually recover, it may require ``repeating" the investment of a large amount of compute.
By annealing the new action in gradually, we ensure that the model never loses overall skill due to a sudden change of one part of the environment; when we observe the model losing \trueskillname during the annealing process, we revert and attempt the anneal at a slower rate.
This annealing process makes sense even if the environment is becoming fundamentally ``harder" because our agent's skill is measured through winrates against other models; the opponent also has to play in the new environment.

\paragraph{Removing Model Parts}
Requiring exact policy equivalence after the surgery outlaws many types of surgery. 
For example, most surgeries which remove parameters are not possible in this framework. 
For this reason our model continued to observe some ``deprecated'' observations, which were simply always set to constants.
Further work such as \cite{unpublishedrobotics} has already begun to explore alternate methods of surgery which avoid this constraint.

\paragraph{Smooth Training Restart}
The gradient moments stored by the Adam optimizer present a nuisance when restarting training with new parameter shape. 
To ensure that the moments have enough time to properly adjust, we use a learning rate of 0 for the first several hours of training after surgery. 
This also ensures that the distribution of rollout games has entered steady state by the time we begin training in earnest.

One additional nuisance when changing the shape of the model is the entire history of parameters which are stored (in the past opponent manager, see \autoref{sec:selfplay}), and used as opponents in rollouts. 
Because the rollout GPUs will be running the newest code, all of these past versions must be updated in the same way as the current version to ensure compatibility. 
If the surgery operation fails to exactly preserve the policy function, these frozen past agents will forever play worse, reducing the quality of the opponent pool.
Therefore it is crucial to ensure agent behavior is unchanged after surgery. 


\paragraph{Benefits of Surgery}
These surgeries primarily permitted us to have a tighter iteration loop for these features. When we added a new game feature which we expect to only matter at high skill, it would simply be impossible to test and iterate on it by training from scratch. Using surgery from the current \openaifive, we could have a more feasible process, which allowed us to safely include many minor features and improvements that otherwise would have been impossible to verify, such as adding long-tail items (Bottle, Rapier), minor improvements to the observation space (stock counts, modifiers on nonheroes), and others.

\section{Hyperparameters}\label{appendix:hyperparams}
The optimization algorithm has several important hyperparameters that have different settings throughout the training process. 
Over the course of training of \openaifive, these hyperparameters were modified by looking for improvement plateaus.
Because of compute limitations which prevented us from testing hyperparameter changes in separate experiments, \openaifive's long-running training process included numerous experimental hyperparameter changes. 
Some of these worked well and were kept, others were reverted as our understanding developed over the course of the 10-month period. 
As it is impossible for us to scan over any of these hyperparameters when our experiment is so large, we make no claim that the hyperparams used are optimal. 

When we ran \cleanexpname we simplified the hyperparameter schedule based on the lessons we had learned. 
In the end we made changes to only four key hyperparameters:
\begin{itemize}
    \item Learning Rate
    \item Entropy penalty coefficient (see \autoref{sec:methods:exploration})
    \item Team Spirit (see \autoref{appendix:rewards})
    \item GAE time horizon (see \autoref{eqn:horizon})
\end{itemize}

\begin{figure}
\centering
\begin{subfigure}{0.59\linewidth}
\begin{tabular}{l|lllll}
Iteration & 0 & 15k & 23k & 43k & 54k \\
Time (days) & 0 & 13 & 20 & 33 & 42 \\
\trueskillname & 0 & 210 & 232 & 245 & 258 \\
\midrule[1pt]
Team Spirit & 0.3 & \multicolumn{4}{|l}{0.8} \\
\hline
GAE Horizon & \multicolumn{2}{l|}{180 secs} & \multicolumn{3}{l}{360 secs}\\
\hline
Entropy coefficient & \multicolumn{3}{l|}{0.01} & \multicolumn{2}{l}{0.001} \\
\hline
Learning Rate & \multicolumn{4}{l|}{5e-5} & 5e-6
\end{tabular}
\end{subfigure}\hfill
\begin{subfigure}{0.39\linewidth}
\includegraphics[width=\textwidth]{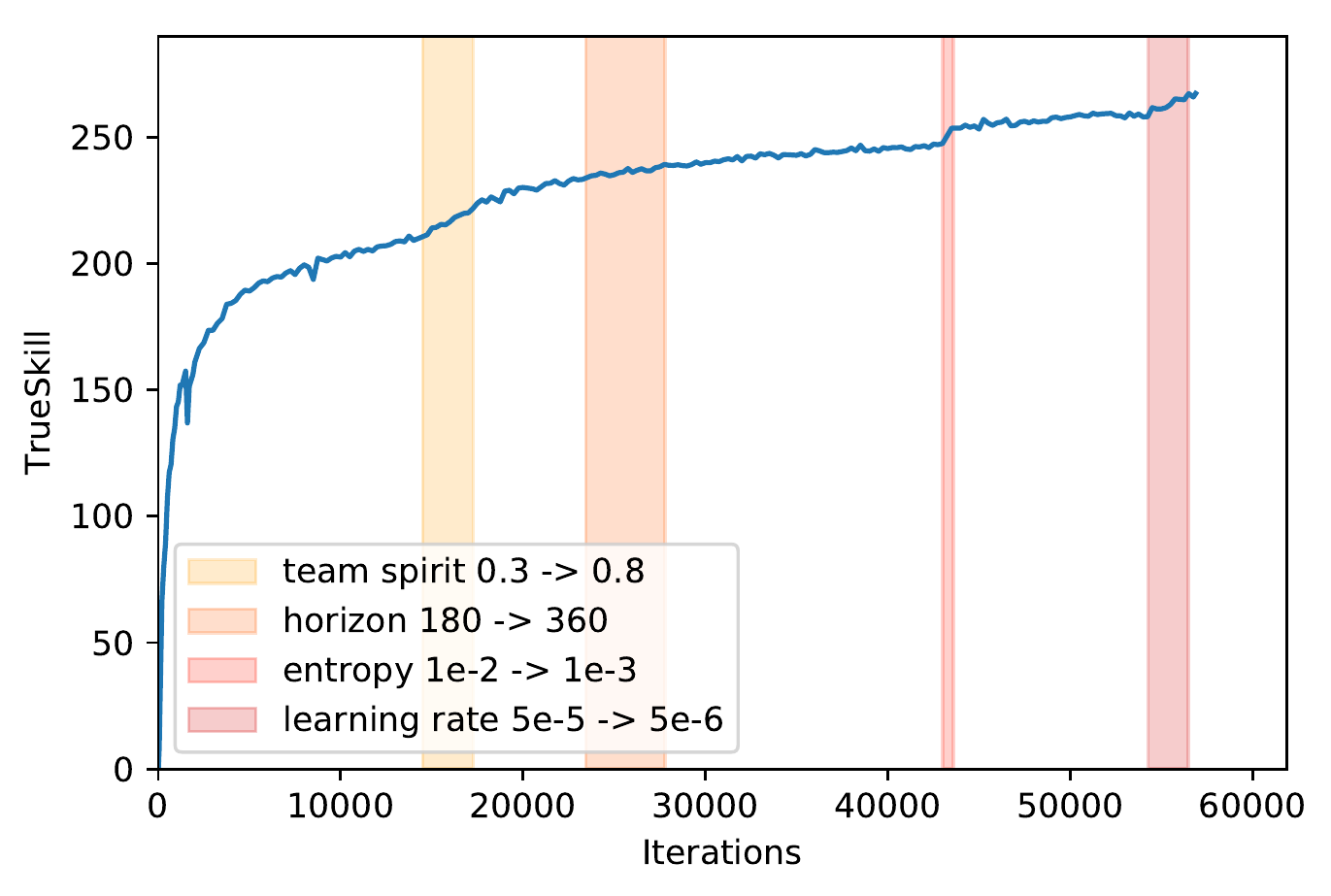}
\end{subfigure}
\caption{Hyperparameter changes during \cleanexpname. Changes are displayed in table-form on the left, and called out in the trueskill vs iterations graph of the training run on the right. Each hyperparameter change was applied gradually over the course of 1-2 days, corresponding to several thousand iterations (the reported time in the table is the start of the change). Our pre-planned schedule included further changes to bring the experiment into line with \openaifive's final hyperparameters (Horizon to 840 sec, team spirit to 1.0, and learning rate to 1e-6), but \cleanexpname reached \openaifive's skill level before we reached those hyperparameters.
}
\label{table:hyperparams-schedule}
\label{fig:hyperparams-rerun}
\end{figure}

This schedule is far from optimized as it was used in only our second iteration of this large experiment. In future work it could likely be significantly improved.

There are many other hyperparams that were not changed during the final \cleanexpname experiment. 
Their values are listed in \autoref{table:hyperparams}. 
Some of these were changed in the original \openaifive out of necessity (e.g. batch size changed many times as more or less compute resources became available, or SampleReuse changed as the relative speeds of different machine pools fluctuated), and others were changed experimentally in the original \openaifive run but were ultimately not important as evidenced by \cleanexpname working without those changes (e.g. increasing the time horizon from 360 seconds to 840 seconds).

\begin{table}
\begin{threeparttable}
\begin{tabular}{l|cccc}
Param & \cleanexpname & \openaifive & Baseline\\
\hline
Frameskip\tnote{e} & 4 & 4 & 4 \\
LSTM Unroll length\tnote{e} & 16 & 16 & 16 \\
Samples Per Segment\tnote{e} & 16 & 16 & 16 \\
Number of optimizer GPUs & 512 & 480$\leftrightarrow$1,536& 64 \\
Batch Size/optimizer GPU (samples) & 120 & 120$\leftrightarrow$128 & 120 \\
Total Batch Size (samples)\tnote{a} & 61,440 & 61,440$\leftrightarrow$196,608 & 7,680 \\
Total Batch Size (timesteps)\tnote{a} & 983,040 & 983,040$\leftrightarrow$3,145,728 & 122,880\\
Number of rollout GPUs & 512 & 500$\leftrightarrow$1,440 & 64 \\
Number of rollout CPUs & 51,200 & 80,000$\leftrightarrow$172,800 & 6,400 \\
Steps per Iteration & 32 & 32 & 32 \\
LSTM Size & 4096 & 2048 $\rightarrow$ 4096 & 4096 \\
Sample Reuse & 1.0 $\leftrightarrow$ 1.1 & 0.8$\leftrightarrow$2.7 & 1.0$\leftrightarrow$1.1 \\
Team Spirit & 0.3 $\rightarrow$ 0.8 & 0.3 $\rightarrow$ 1.0 & 0.3 \\ 
GAE Horizon & 180 secs $\rightarrow$ 360 secs & 60 secs $\rightarrow$ 840 secs & 180 secs\\
GAE $\lambda$ & 0.95 & 0.95 & 0.95 \\
PPO clipping & 0.2 & 0.2 & 0.2 \\
Value loss weight\tnote{c} & 1.0 & $0.25\leftrightarrow 1.0$ & 1.0 \\
Entropy coefficient & 0.01 $\rightarrow$ 0.001 & 0.01 $\rightarrow$ 0.001 & 0.01 \\
Learning rate & 5e-5 $\rightarrow$ 5e-6 & 5e-5 $\leftrightarrow$ 1e-6 & 5e-5\\
Adam $\beta_1$ & 0.9 & 0.9 & 0.9 \\
Adam $\beta_2$ & 0.999 & 0.999 & 0.999 \\
Past opponents\tnote{b} & 20\% & 20\% & 20\% \\
Past Opponents Learning Rate\tnote{d} & 0.01 & 0.01 & 0.01\\
\end{tabular}
\begin{tablenotes}
    \item[a] Batch size can be measured in samples (each an unrolled LSTM of 16 frames) or in individual timesteps.
    \item[b] Fraction of games played against past opponents (as opposed to self-play).
    \item[c] We normalize rewards using a running estimate of the standard deviation, and the value loss weight is applied post-normalization.
    \item[d] See \autoref{sec:selfplay}.
    \item[e] See \autoref{fig:timescales} for definitions of the various timescale subdivisions of a rollout episode.
\end{tablenotes}
\end{threeparttable}
\caption{\textbf{Hyperparameters: } The \openaifive and \cleanexpname columns indicate what was done for those individual experiments. For those which were modified during training, $x\rightarrow y$ indicates a smooth monotonic transition (usually a linear change over one to three days), and $x\leftrightarrow y$ indicates a less controlled variation due to either ongoing experimentation or distributed systems fluctuations. The ``Baseline'' indicates the default values for all the experiments in \autoref{appendix:substudies} (each individual experiment used these hyperparameters other than any it explicitly studied, for example in \autoref{sec:methods:batchsize} the batch size was changed from the baseline in each training run, but all the other hyperparameters were from this table).}
\label{table:hyperparams}
\end{table}

\begin{figure}
    \centering
    \includegraphics[width=0.5\textwidth]{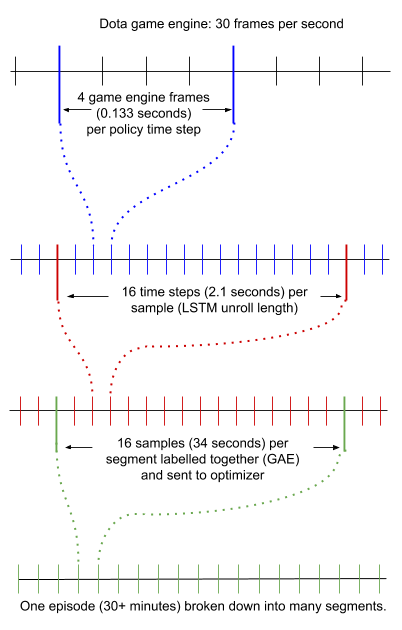}
    \caption{\textbf{Timescales and Staleness: }The breakdown of a rollout game.
    Rather than collect an entire game before sending it to the optimizers, rollout machines send data in shorter segments.
    The segment is further subdivided into samples of 16 policy actions which are optimized together using truncated BPTT.
    Each policy action bundles together four game engine frames.
    }
    \label{fig:timescales}
\end{figure}

\section{Evaluating agents’ understanding}\label{sec:results:understanding}

It is often difficult to infer the intentions of an RL agent.
Some actions are obviously useful --- hitting an enemy that is low on health, or freezing them as they're trying to escape --- but many other decisions can be less obvious.
This is tightly coupled with questions on intentionality: does our agent plan on attacking the tower, or doe it opportunistically deal the most damage possible in next few seconds?

To assess this, we attempt to predict future state of various features of the game from agent's LSTM state:
\begin{itemize}
\item \textbf{Win probability}: Binary label of either 0 or 1 at the end of the game.
\item \textbf{Net worth rank}: Which rank among the team (1-5) in terms of total resources collected will this hero be at the end of the game? This prediction is used by scripted item-buying logic to decide which agents buy items shared by the team such as wards. 
In human play (which the scripted logic is based on) this task is traditionally performed by heroes who will have the lowest net worth at the end of the game.
\item \textbf{Team objectives / enemy buildings}: whether this hero will help the team destroy a given enemy building in the near future.
\end{itemize}

We added small networks of fully-connected layers that transform LSTM output into predictions of these values.
For historical reasons, win probability passes gradients to the main LSTM and rest of the agent with a very small weight; the other auxiliary
predictions use Tensorflow's \texttt{stop\_gradient} method to train on their own.

One difficulty in training these predictors is that we train our agent on 30-second segments of the game (see \autoref{fig:timescales}), and any given 30-second snippet may not contain the ground truth (e.g. for win probability and networth position, we only have ground truth on the very last segment of the game).
We address this by training these heads in a similar fashion to how we train value functions.
If a segment contains the ground truth label, we use the ground truth label for all time steps in that segment; if not, we use the model's prediction at the end of the segment as the label.
For win probability, for example, more precisely the label $y$ for a segment from time $t_1$ to $t_2$ is given by:
\begin{equation}
    y = \left\{
        \begin{array}{ll}
            1 & \textrm{last segment of the game, we win} \\
            0 & \textrm{last segment of the game, we lose} \\
            \hat{y}(t_2) & \textrm{else} \\
        \end{array}
    \right.
\end{equation}
Where $\hat{y}(t_2)$ is the model's predicted win probability at the end of the segment.
Although this requires information to travel backward through the game, we find it trains these heads to a degree of calibration and accuracy.

For the team objectives, we are additionally interested in whether the event will happen soon. 
For these we apply an additional discount factor with horizon of 2 minutes.
This means that the enemy building predictions are not calibrated probabilities, but rather probabilities discounted by the expected time to the event.

\subsection{Understanding \openaifinals}\label{sec:results:understanding:finals}

We used these supervised predictions to look closer at the game 1 from \openaifinals.

\begin{figure}
    \centering
    \begin{subfigure}[t]{0.7\textwidth}
        \centering
        \includegraphics[width=\textwidth]{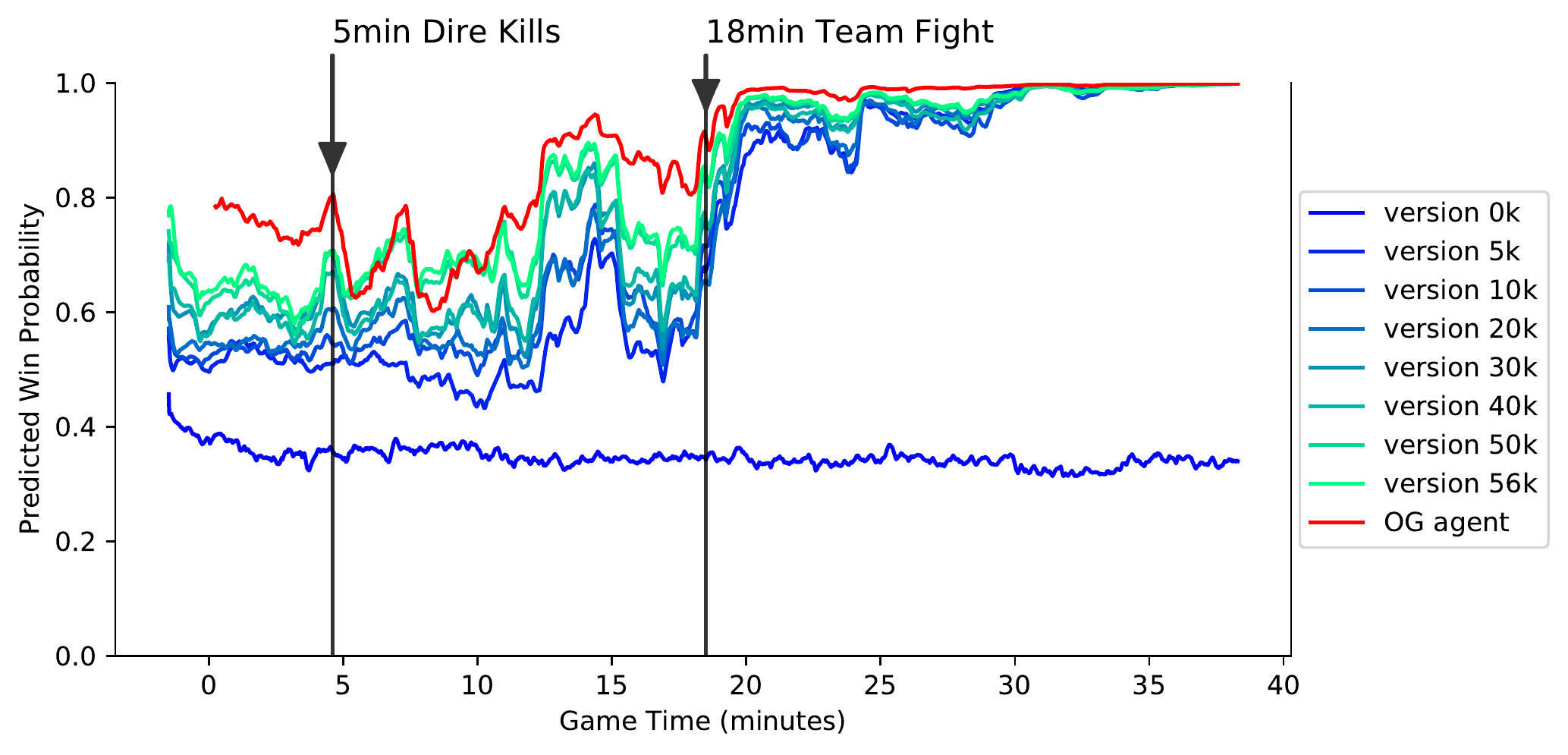}
    \end{subfigure}
    \caption{\textbf{Win Probability prediction of game 1 of \openaifinals}
    In red we show the (\openaifive) agent's win probability prediction over the course of the game (which can be viewed by downloading the replay from \url{https://openai.com/blog/how-to-train-your-openai-five/}).
    Marked are two significant events that significantly affected win probability prediction.
    At roughly 5 minutes in the human team killed several of \openaifive's heroes, making it doubt its lead.
    At roughly 18 minutes in, \openaifive team killed three human heroes in a row, regrouped all their heroes at the mid lane, and marched on declaring 95\% probability of victory.
    Versions 0-56k are progressive versions of \cleanexpname agent predicting win probabilities by replaying the same game; as we can see, prediction converges to that of the bot
    that actually played the game (original \openaifive), despite training over self-play games from separate training runs.
    }
    \label{fig:winrates-OG-1}
\end{figure}

In \autoref{fig:winrates-OG-1} we explore the progression of win probability predictions over the course of training \cleanexpname, illustrating the evolution of understanding. 
Version 5,000 of the agent (early in the training process and low performance) already has a sense of what situations in the game may lead to eventual win. 
The prediction continues to get better and better as training proceeds.
This matches human performance at this task, where even spectators with relatively little gameplay experience can estimate who is ahead based on simple heuristics, but with more gameplay practice human experts can estimate the winner more and more accurately.

On the winrate graph two dramatic game events are marked, at roughly the 5 and 18 minute point.
One of them illustrates OpenAI Five's win probability drop, due to an unexpected loss of 3 heroes in close succession.
The other shows how the game turns from good to great as a key enemy hero is killed.

\begin{figure}
    \centering
    \includegraphics[width=\textwidth]{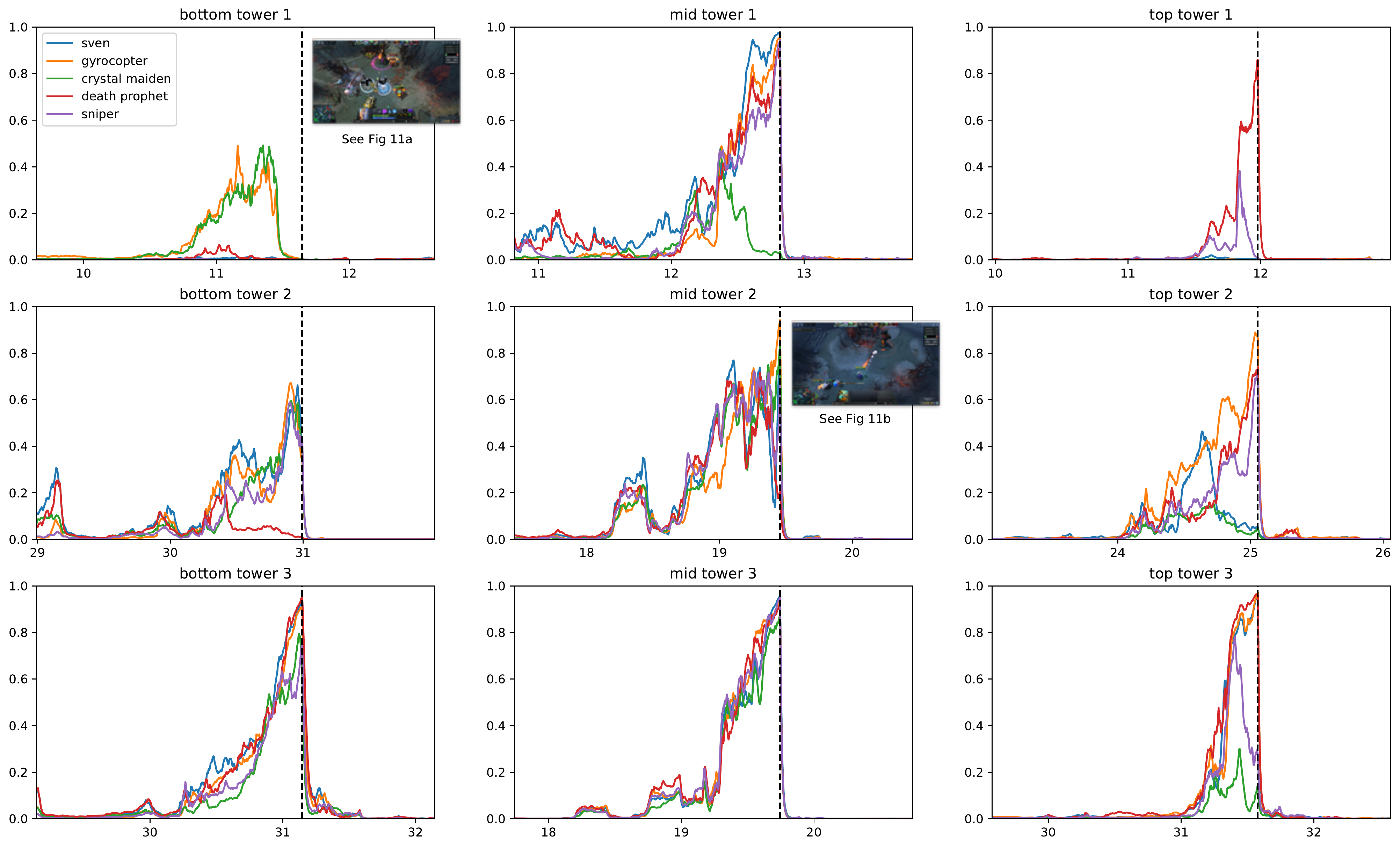}
    \caption{Continuous prediction of destroying enemy buildings by \openaifive in Finals game 1.
    Predictions by different heroes differ as they specifically predict whether they will participate in bringing given building down.
    Predictions should not be read as calibrated probabilities, because they are trained with a discount factor.
    See \autoref{fig:objectives-screenshots-OG-1:tower1bot} and \autoref{fig:objectives-screenshots-OG-1:tower2mid} for descriptions of the events corresponding to two of these buildings.}
    \label{fig:objectives-OG-1}
\end{figure}

\begin{figure}
    \centering
    \begin{subfigure}[t]{0.85\textwidth}
        \centering
        \includegraphics[width=\textwidth]{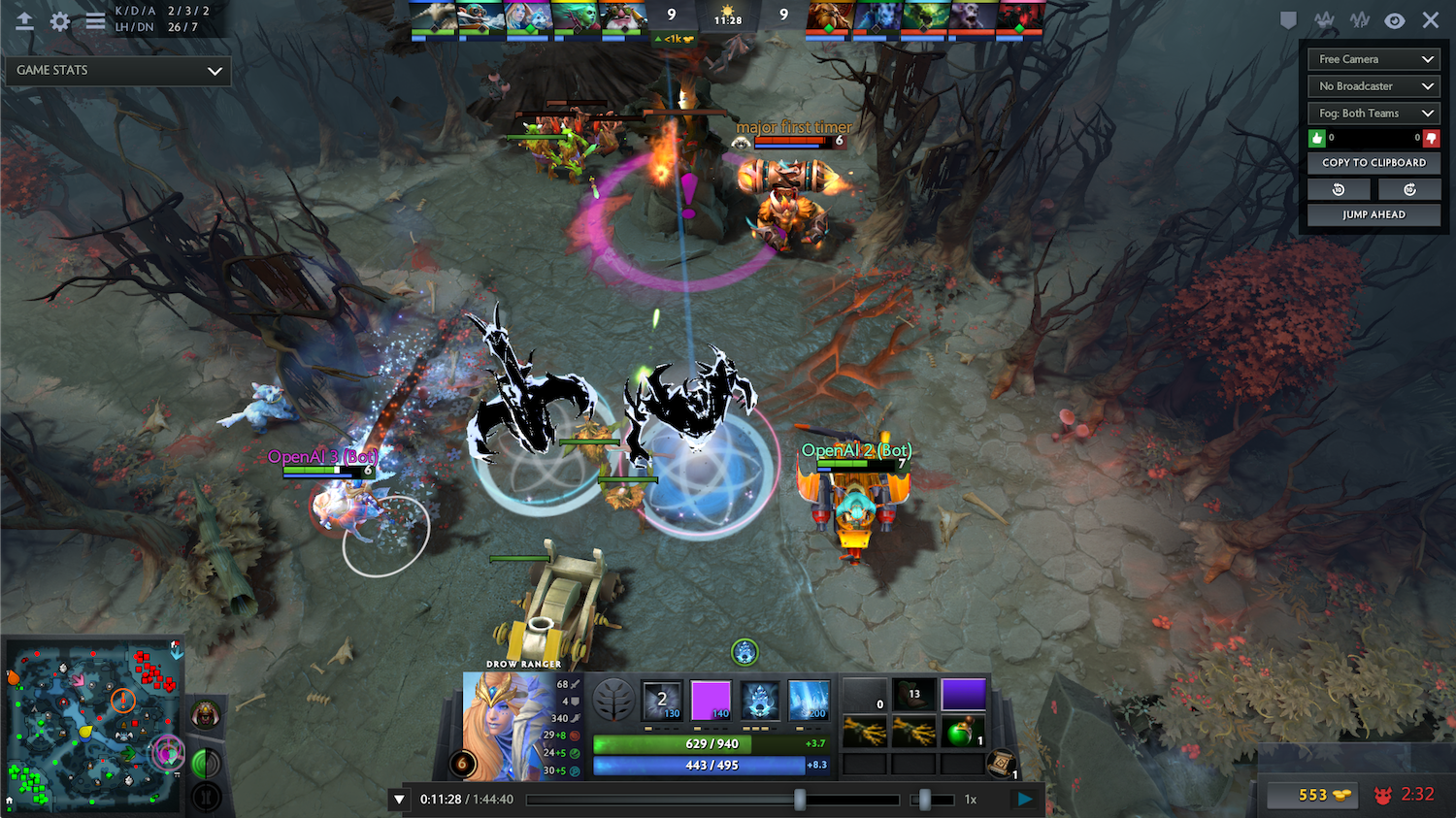}
        \caption{}
        \label{fig:objectives-screenshots-OG-1:tower1bot}
    \end{subfigure}\\
    \begin{subfigure}[t]{0.85\textwidth}
        \centering
        \includegraphics[width=\textwidth]{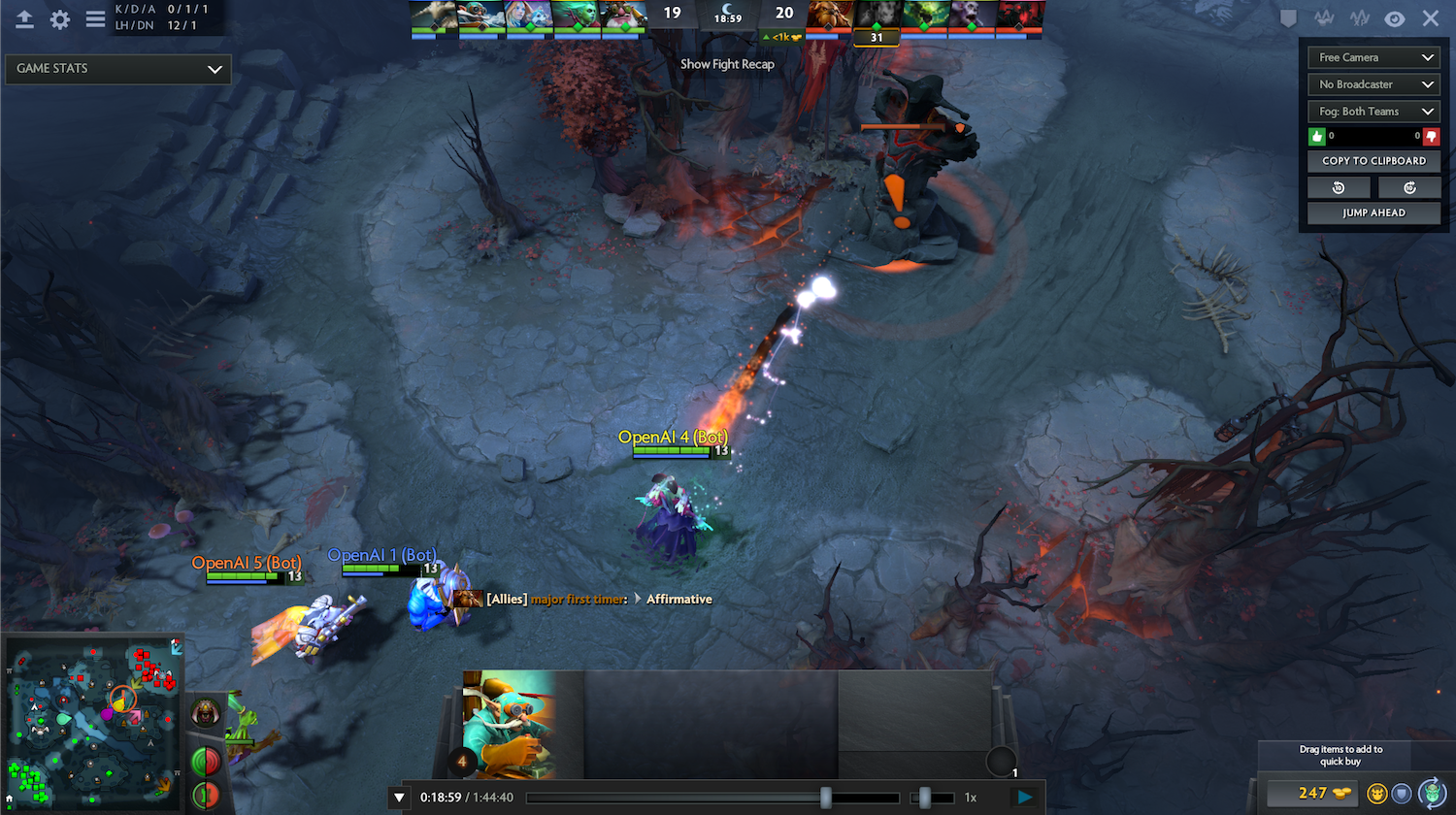}
        \caption{}
        \label{fig:objectives-screenshots-OG-1:tower2mid}
    \end{subfigure}
    \caption{Screenshots right before two of the dire tower falls in the \openaifinals game 1.
    In \ref{fig:objectives-screenshots-OG-1:tower1bot}, Gyrocopter and Crystal Maiden attack the bottom tower 1 (upper left in
    \autoref{fig:objectives-OG-1}) and plan perhaps to kill it (their predictions go up).
    But they are chased away by the incoming dire (human) heroes, and their plan changes (the prediction that they will participate in the tower kill falls back to zero).
    Radiant creeps kill the tower half a minute later.
    In \ref{fig:objectives-screenshots-OG-1:tower2mid}, all radiant heroes attack mid tower 2 (center in \autoref{fig:objectives-OG-1}).
    However just before it falls, few dire heroes show up trying to save it, and most radiant heroes end up chasing them a fair distance away from the building. 
    The prediction for those heroes to participate in the tower kill drops accordingly.
    }
    \label{fig:objectives-screenshots-OG-1}
    
\end{figure}

We also looked at heroes participation in destroying objectives.
In \autoref{fig:objectives-OG-1} we can see different heroes' predictions for each of the objectives in game 1 of \openaifinals.
In several cases all heroes predict they will participate in the attack (and they do).
In few cases one or two heroes are left out, and indeed by watching the game replay we see that those heroes are busy in the different part of the map during that time.
In \autoref{fig:objectives-screenshots-OG-1} we illustrate these predictions with more details for two of the events.

\subsection{Hero selection}\label{sec:results:understanding:selection}

In the normal game of \dota, two teams at the beginning of the game go through the process of selecting heroes.
This is a very important step for future strategy, as heroes have different skill sets and special abilities.
\openaifive, however, is trained purely on learning to play the best game of \dota possible given randomly selected heroes.

\begin{figure}
    \centering
    \includegraphics[width=\textwidth, trim=0 295 0 0,clip]{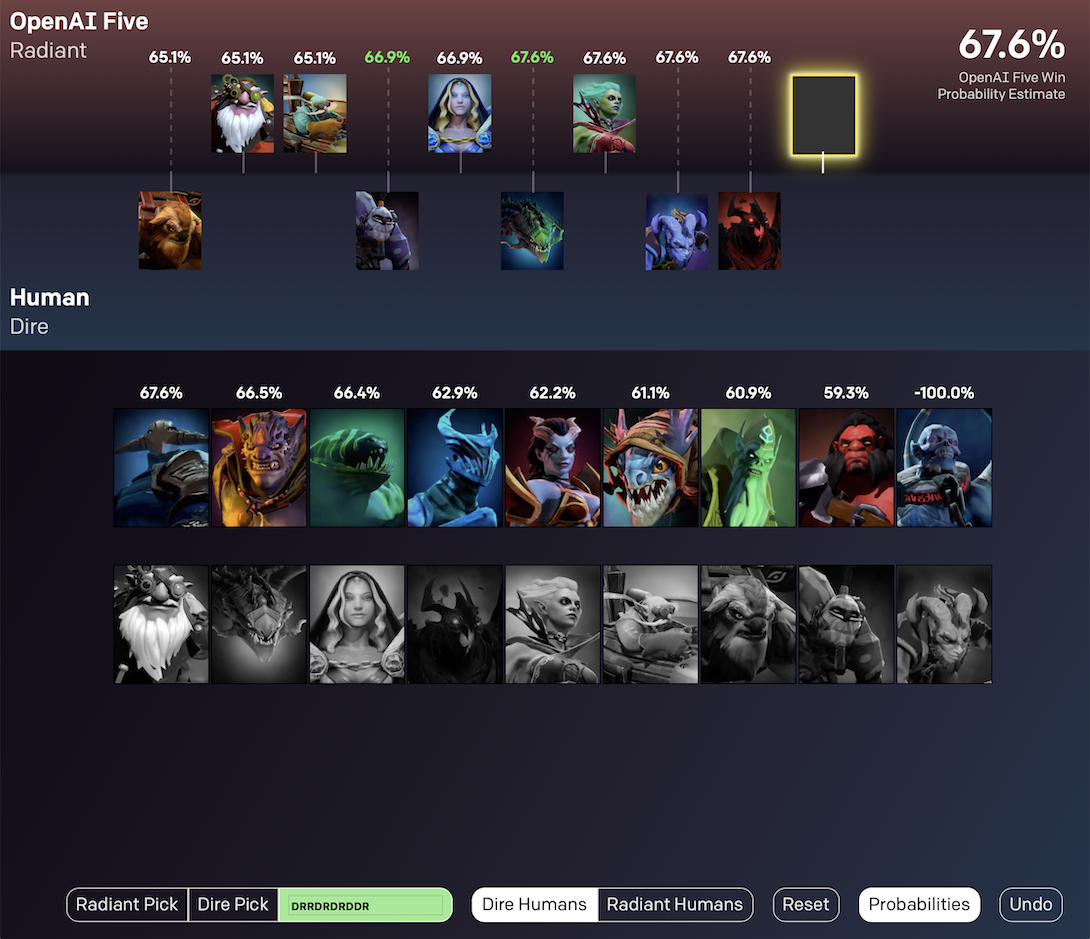}
    \caption{When drafting heroes, our drafting program would pick the one that maximizes worst-case scenario of opponent hero selection (minimax algorithm).
    In this example (from OpenAI Finals game 1), OpenAI Five deems the humans' first pick suboptimal, immediately updating its expected win probability of $52.8\%$ to $65.1\%$.
    The drafter then makes two choices (which it believes to be optimal of course).
    The humans' second and third choices further decreases their chances of victory (according to the agent), indicated by the green win probability.
    However, for the human team's last two choices, OpenAI Five agrees they were optimal, as can be seen by the win probability remaining constant (even though choice 4, Riki, is a character very differently played by humans and by \openaifive).
}
    \label{fig:hero-selection}
\end{figure}

Although we could likely train a separate drafting agent to play the draft phase, we do not need to; instead we can use the win probability predictor.
Because the main varying observation that agents see at the start of the game is which heroes are on each team, the win probability at the start of the game estimates the strength of a given matchup. 
Because there are only 4,900,896 combinations of two 5-hero teams from the pool of 17 heroes, we can precompute agent's predicted win probability from the first few frames of every lineup.
Given these precomputed win probabilities, we apply a dynamic programming algorithm to draft the best hero available on each turn.
Results of this approach in a web-based drafting program that we have built can be seen on \autoref{fig:hero-selection}.

In addition to building a hero selection tool, we also learned about our agent's preferences from this exercise.
In many ways \openaifive's preferences match human player's preferences such as placing a high value (within this pool) on the hero Sniper.
In other ways it does not agree with typical human knowledge,
for example 
it places low value on Earthshaker.
Our agent had trouble dealing with geometry of this hero's ``Fissure'' skill, making this hero worse than others in training rollouts.

Another interesting tidbit is that at the very start of the draft, before any heroes are picked, \openaifive believes that the Radiant team has a 54\% win chance (if picking first in the draft) or 53\% (if picking second).
Our agent's higher estimate for the Radiant side over the Dire agrees with conventional wisdom within the \dota community.
Of course, this likely depends on the set of heroes available.

\section{Observation Space}\label{appendix:observationspace}
\begin{figure}
    \centering
    \includegraphics[width=0.75\textwidth]{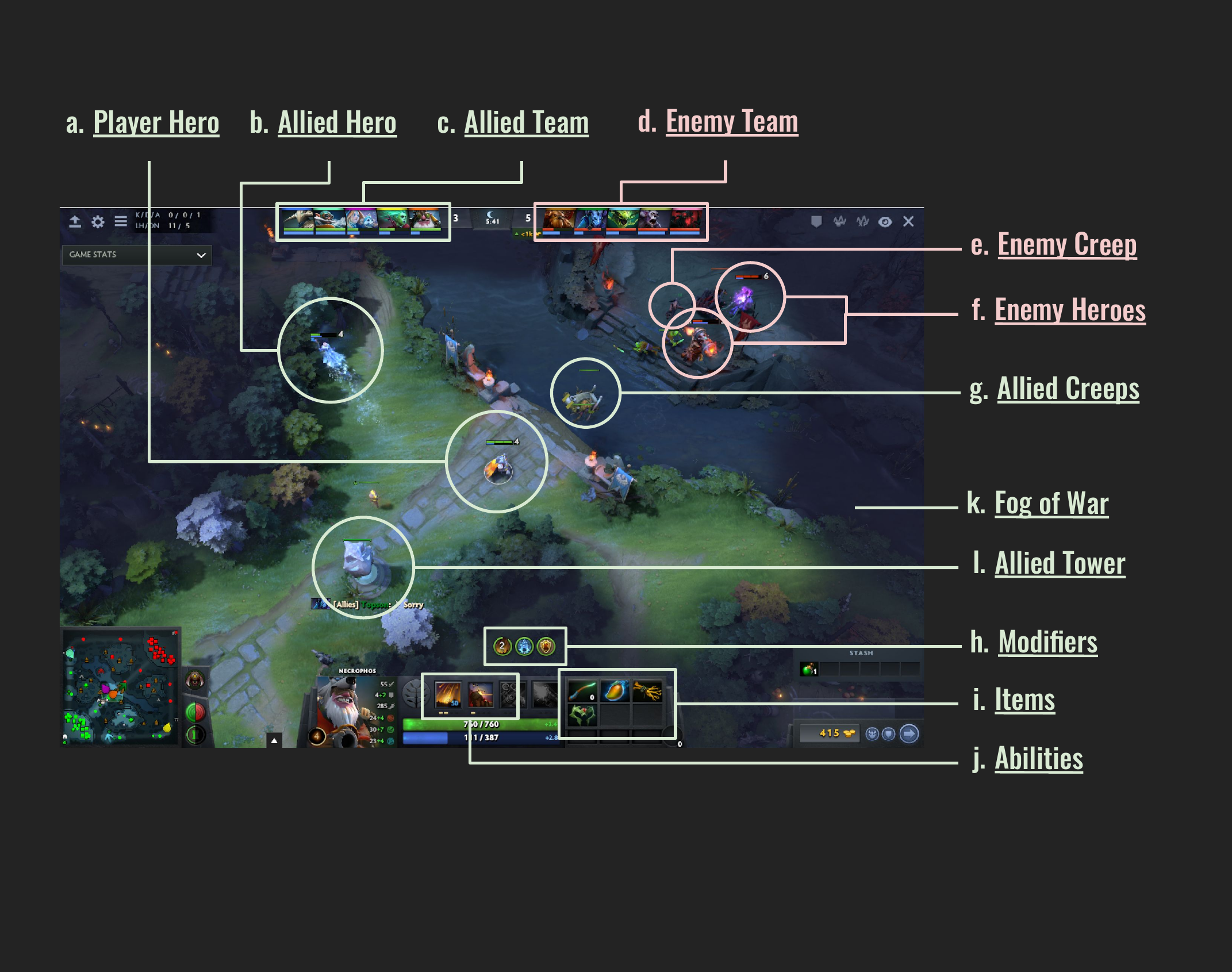}
    \caption{\dota's human ``Observation Space''}
    \label{fig:dotahud}
\end{figure}
\begin{figure}
    \centering
    \includegraphics[width=\textwidth]{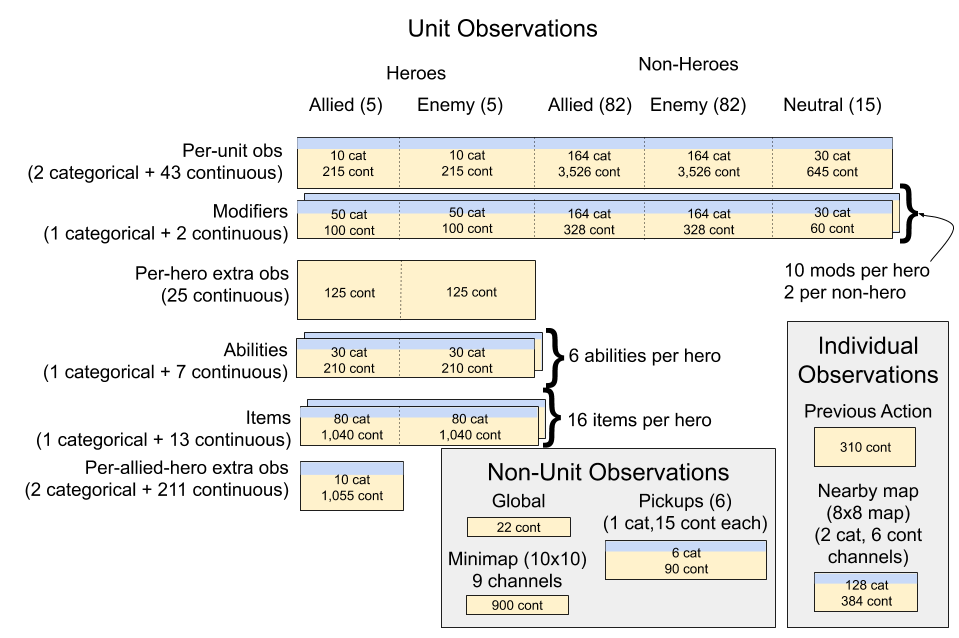}
    \caption{\textbf{Observation Space Overview:} 
    The arrays that \openaifive observes at each timestep.
    Most of \openaifive's observations  are unit-centered; for 189 different units on the map, we observe a set of basic properties. These units are grouped along the top of the figure.
    We observe some data about all units, some extra data about the primary units (the heroes), and even more data about the heroes on our team.
    A few observations are not tied to any unit. 
    Finally, two observations having to do with hero control (terrain near me, and my previous action) are only observed about the individual hero that this LSTM replica operates.
    In this diagram blue bands represent categorical data and yellow bands represent continuous or boolean data; most entities (units, modifiers, abilities, items, and pickups), have some of each.
    Each piece of the figure summarizes the total dimensionality of that portion of the input.
    All together, an \openaifive hero observes 1,200 categorical values 
    and 14,534 continuous/boolean 
    values.\nicetohave{Add little images to the minimap and other places where it will help.}}
    \label{fig:obs-space}
\end{figure}

\begin{table}
\begin{scriptsize}
\begin{threeparttable}
\begin{subtable}[t]{0.3\textwidth}
\begin{tabular}[t]{|p{4cm}|c|}
  \hline
  \textbf{Global data}   & \textbf{22}\\\hline
  time since game started & 1 \\\hline
  is it day or night? &  \\
  time to next day/night change & 2 \\\hline
  time to next spawn: creep, neutral, bounty, runes & 4 \\\hline
  time since seen enemy courier &  \\
  is that $>$ 40 seconds?\tnote{a} & 2 \\\hline
  min\&max time to Rosh spawn & 2 \\\hline
  Roshan's current max hp & 1 \\\hline
  is Roshan definitely alive? & 1 \\\hline
  is Roshan definitely dead? & 1 \\\hline
  Next Roshan drops cheese? & 1 \\\hline
  Next Roshan drops refresher? & 1 \\\hline
  Roshan health randomization\tnote{b} & 1 \\\hline
  Glyph cooldown (both teams) & 2 \\\hline
  Stock counts\tnote{c} & 4 \\\hline
\end{tabular}
\begin{tabular}{|p{4cm}|c|}
  \hline
  \textbf{Per-unit (189 units)}   &  \textbf{43}\\\hline
  position (x, y, z) & 3 \\\hline
  facing angle (cos, sin) & 2 \\\hline
  currently attacking?\tnote{e} & \\
  time since last attack\tnote{d} & 2 \\\hline
  max health & \\
  last 16 timesteps' hit points & 17 \\\hline
  attack damage, attack speed & 2 \\\hline
  physical resistance & 1 \\\hline
  invulnerable due to glyph? & \\
  glyph timer & 2 \\\hline
  movement speed & 1 \\\hline
  on my team? neutral? & 2 \\\hline
  animation cycle time & 1 \\\hline
  eta of incoming ranged \& tower creep projectile (if any) & \\
  \# melee creeps atking this unit\tnote{d} & 3 \\\hline
  [Shrine only] shrine cooldown & 1 \\\hline
  vector to me (dx, dy, length)\tnote{e} & 3 \\\hline
  am I attacking this unit?\tnote{e} & \\
  is this unit attacking me?\tnote{d,e} & \\
  eta projectile from unit to me\tnote{e} & 3 \\\hline
  \categorical unit type & \categorical 1 \\\hline
  \categorical current animation & \categorical 1 \\
  \hline
\end{tabular}
\end{subtable}
\hfill
\begin{subtable}[t]{0.3\textwidth}
\begin{tabular}[t]{|p{4cm}|c|}
  \hline
  \textbf{Per-hero add'l (10 heroes)}   &  \textbf{25}\\\hline
  is currently alive? & 1 \\\hline
  number of deaths & 1 \\\hline
  hero currently in sight? & \\
  time since this hero last seen & 2 \\\hline
  hero currently teleporting? & \\
  if so, target coordinates (x, y) & \\
  time they've been channeling & 4 \\\hline
  respawn time & 1 \\\hline
  current gold (allies only) & 1 \\\hline
  level & 1 \\\hline
  mana: max, current, \& regen & 3 \\\hline
  health regen rate & 1 \\\hline
  magic resistance & 1 \\\hline
  strength, agility, intelligence & 3 \\\hline
  currently invisible? & 1 \\\hline
  is using ability? & 1\\\hline
  \# allied/enemy creeps/heroes in line btwn me and this hero\tnote{e} & 4 \\\hline
  \hline
  \textbf{Per-allied-hero additional (5 allied heroes)}   & \textbf{211} \\\hline
  Scripted purchasing settings\tnote{b} & 7 \\\hline
  Buyback: has?, cost, cooldown & 3 \\\hline
  Empty inventory \& backpack slots & 2 \\\hline
  Lane Assignments\tnote{b} & 3 \\\hline
  Flattened nearby terrain: 14x14 grid of passable/impassable? & 196 \\\hline
  \categorical scripted build id & \categorical\\
  \categorical next item to purchase\tnote{b} & \categorical 2 \\\hline
  \hline
  \textbf{Nearby map (8x8)\tnote{e}}   & \textbf{6} \\\hline
  terrain: elevation, passable? & 2 \\\hline
  allied \& enemy creep density & 2 \\\hline
  area of effect spells in effect.\tnote{f} & 2 \\\hline
  \categorical area of effect spells in effect.\tnote{f} & \categorical 2 \\\hline
  \hline
  \textbf{Previous Sampled Action\tnote{e}}   & \textbf{310} \\\hline
  Offset? (Regular, Caster, Ward) & 3x2x9 \\\hline
  Unit Target's Embedding & 128 \\\hline
  Primary Action's Embedding & 128 \\\hline
\end{tabular}
\end{subtable}
\hfill
\begin{subtable}[t]{0.3\textwidth}
\begin{tabular}[t]{|p{4cm}|c|}
  \hline
  \textbf{Per-modifier (10 heroes x 10 modifiers \& 179 non-heroes x 2 modifiers)}   &  \textbf{2}\\\hline
  remaining duration & 1 \\\hline
  stack count & 1 \\\hline
  \categorical modifier name & \categorical 1 \\
  \hline
  \hline
  \textbf{Per-item (10 heroes x 16 items)}   & \textbf{13}\\\hline
  location one-hot (inventory/backpack/stash) & 3 \\\hline
  charges & 1 \\\hline
  is on cooldown? & \\
  cooldown time & 2 \\\hline
  is disabled by recent swap? & \\
  item swap cooldown & 2 \\\hline
  toggled state & 1 \\\hline
  special Power Treads one-hot & \\
  (str/agi/int/none) & 4\\\hline
  \categorical item name & \categorical 1 \\
  \hline
  \hline
  \textbf{Per-ability (10 heroes x 6 abilities)}   &  \textbf{7}\\\hline
  cooldown time & 1 \\\hline
  in use? & 1 \\\hline
  castable & 1 \\\hline
  Level 1/2/3/4 unlocked?\tnote{d} & 4\\\hline
  \categorical ability name & \categorical 1 \\
  \hline
  \hline
  \textbf{Per-pickup (6 pickups)}   &  15\\\hline
  status one-hot (present/not present/unknown) & 3 \\\hline
  location (x, y) & 2 \\\hline
  distance from all 10 heroes & 10 \\\hline
  \categorical pickup name & \categorical 1 \\
  \hline
  \hline
  \textbf{Minimap (10 tiles x 10 tiles)}   &  \textbf{9}\\\hline
  fraction of tile visible & 1 \\\hline
  \# allied \& enemy creeps & 2 \\\hline
  \# allied \& enemy wards & 2 \\\hline
  \# enemy heroes & 1 \\\hline
  cell (x, y, id) & 3 \\\hline
\end{tabular}
\end{subtable}
\end{threeparttable}
\begin{threeparttable}
\begin{tablenotes}
    \item[a] These observations are leftover from an early version of Five which played a restricted 1v1 version of the game. They are likely obsolete and not needed, but this was not tested.
    \item[b] These observations are about our per-game randomizations. See \autoref{sec:methods:exploration}.
    \item[c] For items: gem, smoke of deciept, observer ward, infused raindrop.
    \item[d] Observations are not visible per-se, but can be estimated. We use scripted logic to estimate them from visible observations.
    \item[e] These observations (only) are different for the five different heroes on the team.
    \item[f] This observation appears twice, and serves as an example of the difficulties of surgery. Although this is a categorical input, we began by treating it as a float input to save on engineering work (this observation is unlikely to be very important). Later the time came to upgrade it to a properly embedded categorical input, but our surgery tools do not support removing existing observations. Hence we added the new observation, but were forced to leave the deprecated observation as well.
\end{tablenotes}
\end{threeparttable}
\end{scriptsize}
\caption{
\textbf{Full Observation Space: }
All observations \openaifive receives at each time step. Blue rows are categorical data. Entries with a question mark are boolean observations (only take values 0 or 1 but treated as floats otherwise). The bulk of the observations are per-unit observations, observed for each of 189 units: heroes (5), creeps (30), buildings (21), wards (30), and courier (1) for each team, plus 15 neutrals. If the number of visible units in a category is less than the allotted number, the rest are padded with zeroes. If more, we observe only the units closest to allied heroes. Units in fog of war are not observed. When enemy heroes are in fog of war, we reuse the observation from the last time step when the unit was visible.
}
\label{table:full-obs}
\end{table}

At each time step one of our heroes observes \obssize inputs about the game state (mostly real numbers with some integer categorical data as well). 
See \autoref{fig:obs-space} for a schematic outline of our observation space and \autoref{table:full-obs} for a full listing of the observations. 

Instead of using the pixels on the screen, we approximate the information available to a human player in a set of data arrays. 
This approximation is imperfect; there are small pieces of information which humans can gain access to which we have not encoded in the observations. 
On the flip side, while we were careful to ensure that all the information available to the model is also available to a human, the model does get to see \emph{all} the information available simultaneously every time step, whereas a human needs to click into various menus and options to get that data. 
Although these discrepancies are a limitation, we do not believe they meaningfully detract from our ability to benchmark against human players.

Humans observe the game via a rendered screen, depicted in \autoref{fig:dotahud}. 
\openaifive uses a more semantic observation space than this for two reasons: First, because our goal is to study strategic planning and gameplay rather than focus on visual processing. 
Second, it is infeasible for us to render each frame to pixels in all training games; this would multiply the computation resources required for the project manyfold.

All float observations (including booleans which are treated as floats that happen to take values 0 or 1) are normalized before feeding into the neural network. For each observation, we keep a running mean and standard deviation of all data ever observed; at each timestep we subtract the mean and divide by the st dev, clipping the final result to be within (-5, 5).

\section{Action Space}\label{appendix:actionspace}

\dota is usually controlled using a mouse and keyboard. The majority of the actions involve a high-level command (attack, use a certain spell, or activate a certain item), along with a target (which might be an enemy unit for an attack, or a spot on the map for a movement). For that reason we represent the action our agent can choose at each timestep as a single {\it primary action} along with a number of {\it parameter actions}. 

The number of primary actions available varies from time step to time step, averaging 8.1 in the games against OG. The primary actions available at a given time include universal actions like noop, move, attack, and others; use or activate one of the hero's spells; use or activate one of the hero's items; situational actions such as Buyback (if dead), Shrine (if near a shrine), or Purchase (if near a shop); and more.
For many of the actions we wrote simple action filters, which determine whether the action is available; these check if there is a valid target nearby, if the ability/item is on cooldown, etc. At each timestep we restrict the set of available actions using these filters and present the final choices to the model.

\begin{figure}
\centering
\begin{subfigure}[t]{0.3\textwidth}
\includegraphics[width=\textwidth]{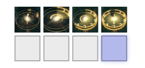}
\caption{\textbf{Delay:} An integer from 0 to 3 indicating which frame during the next frameskip to take the action on (see \autoref{appendix:reaction-time}). 
If 0, the action will be taken immediately when the game engine processes this time step; if 3, the action will be taken on the last game frame before the next policy observation. 
This parameter is never ignored.}
\label{fig:action-head-delay}
\end{subfigure}
\hfill
\begin{subfigure}[t]{0.3\textwidth}
\includegraphics[width=\textwidth]{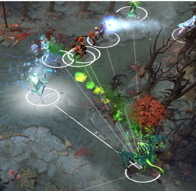}
\caption{\textbf{Unit Selection:} One of the 189 visible units in the observation. 
For actions and abilities which target units, either enemy units or friendly units.
For many actions, some of the possible unit targets will be invalid; attempting an action with an invalid target results in a noop.}
\label{fig:action-head-unit}
\end{subfigure}
\hfill
\begin{subfigure}[t]{0.3\textwidth}
\includegraphics[width=\textwidth]{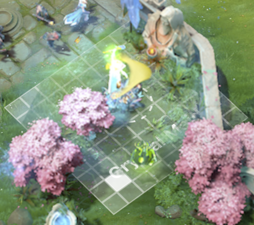}
\caption{\textbf{Offset:} A 2D $(X, Y)$ coordinate indicating a spatial offset, used for abilities which target a location on the map. 
The offset is interpreted relative to the caster or the unit selected by the Unit Selection parameter, depending on the ability. 
Both $X$ and $Y$ are discrete integer outputs ranging from -4 to +4 inclusive, producing a grid of 81 possible coordinate pairs.}
\label{fig:action-head-offset}
\end{subfigure}
\caption{\textbf{Action Parameters}}
\label{fig:action-params}
\end{figure}

In addition to a primary action, the model chooses action parameters. At each timestep the model outputs a value for each of them; depending on the primary action, some of them are read and others ignored (when optimizing, we mask out the ignored ones since their gradients would be pure noise). There are 3 parameter outputs, Delay (4 dim), unit selection (189 dim), and offset (81 dim), described in \autoref{fig:action-params}.

All together this produces a combined factorized action space size of up to $30\times 4 \times 189 \times 81=1,837,080$ dimensions (30 being the maximum number of primary actions we support). This number ignores the fact that the number of primary actions is usually much lower; some parameters are masked depending on the primary action; and some parameter combinations are invalid and those actions are treated as no-ops.

To get a better picture, we looked at actual data from the two games played against Team OG, and simply counted number of available actions at each step.
The average number of available actions varies significantly across heroes, as different heroes have different numbers spells and items with larger parameter counts.
Across the two games the average number of actions for a hero varied from 8,000 to 80,000.


Unit Selection and Offset are actually implemented within the model as several different, mutually exclusive parameters depending on the primary action. For Unit Selection, we found that using a single output head caused that head to learn very well to target tactical spells and abilities. One ability called ``teleport,'' however, is significantly different from all the others --- rather than being used in a tactical fight, it is used to strategically reposition units across the map. Because the action is much more rare, the learning signal for targeting this ability would be drowned out if we used a single model output head for both. For this reason the model outputs a normal Unit Selection parameter and a separate Teleport Selection parameter, and one or the other is used depending on the primary action. Similarly, the Offset parameter is split into ``Regular Offset,'' ``Caster Offset'' (for actions which only make sense offset from the caster), and ``Ward Placement Offset'' (for the rare action of placing observer wards).

We categorize all primary actions into 6 ``Action target types'' which determines which parameters the action uses, listed in \autoref{table:action-target-types}.

\begin{table}
\begin{tabular}{lll}
Action Target Type & Example & Parameters \\\hline
No Target & Power Treads & Delay
\\
Point Target & Move & Delay, Offset (Caster)
\\
Unit Target & Attack & Delay, Unit Selection (Regular)
\\
Unit Offset Target & Sniper's Shrapnel & Delay, Unit Selection (Regular), Offset (Regular)
\\
Teleport Target & Town Portal Scroll & Delay, Unit Selection (Teleport), Offset (Regular)
\\
Ward Target & Place Observer Ward & Delay, Offset (Ward)
\\
\end{tabular}
\caption{\textbf{Action Target Types}}
\label{table:action-target-types}
\end{table}

\subsection{Scripted Actions}\label{sec:game:rep:scriptedac}
Not all actions that a human takes in a game of \dota are controlled by our RL agent. Some of the actions are {\it scripted}, meaning that we have written a rudimentary rules-based system to handle these decisions. 
Most of these are for historical reasons --- at the start of the project we gave the model control over a small set of the actions, and we gradually expanded it over time. 
Each additional action that we remove from the scripted logic and hand to the model's control gives the RL system a higher potential skill cap, but comes with an cost measured in engineering effort to set it up and risks associated with learning and exploration.
Indeed even when adding these new actions gradually and systematically, we occasionally encountered instabilities; for example the agent might quickly learn never to take a new action (and thus fail to explore the small fraction of circumstances where that action helps), and thus moreover fail to learn (or unlearn) the dependent parts of the gameplay which require competent use of the new action. 

In the end there were still several systems that we had not yet removed from the scripted logic by the time the agent reached superhuman performance.
While we believe the agent could ultimately perform better if these actions were not scripted, we saw no reason to do remove the scripting because superhuman performance had already been achieved.
The full set of remaining scripted actions is:
\begin{enumerate}
    \item \textbf{Ability Builds:} Each hero has four spell abilities. Over the course of the game, a player can choose which of these to ``level up,'' making that particular skill more powerful. For these, in evaluation games we follow a fixed schedule (improve ability X at level 1, then Y at level 2, then Z at level 3, etc). In training, we randomize around this fixed script somewhat to ensure the model is robust to the opponent choosing a different schedule. 
    \item \textbf{Item Purchasing:} As a hero gains gold, they can purchase items. We divide items into \emph{consumables} --- items which are consumed for a one-time benefit such as healing --- and everything else. 
    For consumables, we use a simple logic which ensures that the agent always has a certain set of consumables; when the agent uses one up, we then purchase a new one. 
    After a certain time in the game, we stop purchasing consumables.
    For the non-consumables we use a system similar to the ability builds - we follow a fixed schedule (first build X, then Y, then Z, etc). 
    Again at training time we randomly perturb these builds to ensure robustness to opponents using different items.\footnote{This randomization is done randomly deleting items from the build order and randomly inserting new items sampled from the distribution of which items that hero usually buys in human games. This is the only place in our system which relies on data from human games.}
    \nicetohave{Put the actual item + ability builds in a table?}
    \item \textbf{Item Swap:} Each player can choose 6 of the items they hold to keep in their ``inventory'' where they are actively usable, leaving up to 3 inactive items in their ``backpack.'' Instead of letting the model control this, we use a heuristic which approximately keeps the most valuable items in the inventory.
    \item \textbf{Courier Control:} Each side has a single ``Courier'' unit which cannot fight but can carry items from the shop to the player which purchased them. We use a state-machine based logic to control this character.
\end{enumerate}

\section{Reward Weights}\label{appendix:rewards}
\begin{table}
\begin{threeparttable}
\begin{tabular}{c|c|c|p{8cm}}
    Name          & Reward & Heroes & Description \\\hline
    Win           & 5      & Team & \\
    Hero Death    & -1     & Solo & \\
    Courier Death & -2     & Team & \\
    XP Gained     & 0.002  & Solo & \\
    Gold Gained   & 0.006  & Solo & For each unit of gold gained. Reward is not lost when the gold is spent or lost. \\
    Gold Spent    & 0.0006 & Solo & Per unit of gold spent on items without using courier. \\
    Health Changed& 2      & Solo & Measured as a fraction of hero's max health.\tnote{$\ddagger$} \\
    Mana Changed  & 0.75   & Solo & Measured as a fraction of hero's max mana. \\
    Killed Hero   & -0.6   & Solo & For killing an enemy hero. The gold and experience reward is very high, so this reduces the total reward for killing enemies. \\
    Last Hit      & -0.16  & Solo & The gold and experience reward is very high, so this reduces the total reward for last hit to $\sim 0.4$. \\
    Deny          & 0.15   & Solo & \\
    Gained Aegis  & 5      & Team & \\
    Ancient HP Change & 5     & Team & Measured as a fraction of ancient's max health. \\
    Megas Unlocked   & 4      & Team & \\
    T1 Tower\tnote{*}      & 2.25   & Team & \\
    T2 Tower\tnote{*}      & 3   & Team & \\
    T3 Tower\tnote{*}    & 4.5 & Team & \\
    T4 Tower\tnote{*}    & 2.25 & Team & \\
    Shrine\tnote{*}      & 2.25 & Team & \\
    Barracks\tnote{*}    & 6 & Team & \\
    Lane Assign\tnote{$\dagger$}  & -0.15 & Solo & Per second in wrong lane. \\
\end{tabular}
\begin{tablenotes}
    \item[*] For buildings, two-thirds of the reward is earned linearly as the building loses health, and one-third is earned as a lump sum when it dies.
    \item[$\dagger$] See \autoref{sec:lane-assignments}.
    \item[$\ddagger$] Hero's health is quartically interpolated between 0 (dead) and 1 (full health); health at fraction $x$ of full health is worth $\left(x + 1 - (1-x)^4\right)/2$. This function was not tuned; it was set once and then untouched for the duration of the project.
\end{tablenotes}
\end{threeparttable}
\caption{Shaped Reward Weights}
\label{table:rewards}
\end{table}

Our agent's ultimate goal is to win the game.
In order to simplify the credit assignment problem (the task of figuring out which of the many actions the agent took during the game led to the final positive or negative reward), we use a more detailed reward function.
Our shaped reward is modeled loosely after potential-based shaping functions \cite{Ng99policyinvariance}, though the guarantees therein do not apply here. 
We give the agent reward (or penalty) for a set of actions which humans playing the game generally agree to be good (gaining resources, killing enemies, etc).

All the results that we reward can be found in \autoref{table:rewards}, with the amount of the reward. Some are given to every hero on the team (``Team'') and some just to the hero who took the action ``Solo.'' Note that this means that when team spirit is 1.0, the total amount of reward is five times higher for ``Team'' rewards than ``Solo'' rewards.

In addition to the set of actions rewarded and their weights, our reward function contains 3 other pieces: 
\begin{itemize}
    \item \textbf{Zero sum: } The game is zero sum (only one team can win), everything that benefits one team necessarily hurts the other team. We ensure that all our rewards are zero-sum, by subtracting from each hero's reward the average of the enemies' rewards.
    \item \textbf{Game time weighting: } Each player's ``power'' increases dramatically over the course of a game of \dota. 
    A character who struggled to kill a single weak creep early in the game can often kill many at once with a single stroke by the end of the game. This means that the end of the game simply produces more rewards in total (positive or negative). 
    If we do not account for this, the learning procedure focuses entirely on the later stages of the game and ignores the earlier stages because they have less total reward magnitude. 
    We use a simple renormalization to deal with this, multiplying all rewards other than the win/loss reward by a factor which decays exponentially over the course of the game. Each reward $\rho_i$ earned a time $T$ since the game began is scaled:
    \begin{equation}
        \rho_i \leftarrow \rho_i \times 0.6^{(T/\textrm{10 mins})}
    \end{equation}
    \item \textbf{Team Spirit: }
    Because we have multiple agents on one team, we have an additional dimension to the credit assignment problem, where the agents need learn which of the five agent's behavior cause some positive outcome.
    The partial rewards defined in \autoref{table:rewards} are an attempt to make the credit assignment easier, but they may backfire and in fact add more variance if an agent receives reward when a \emph{different} agent takes a good action.
    To attempt dealing with this, we have introduced {\it team spirit}.
    It measures how much agents on the team share in the spoils of their teammates.
    If each hero earns raw individual reward $\rho_i$, then we compute the hero's final reward $r_i$ as follows:
    \begin{equation}
        r_i = (1 - \tau) \rho_i + \tau \overline{\rho}
    \end{equation}
    with scalar $\overline{\rho}$ being equal to mean of $\rho$.
    If team spirit is 0, then it's every hero for themselves; each hero only receives reward for their own actions $r_i=\rho_i$.
    If team spirit is 1, then every reward is split equally among all five heroes; $r_i = \overline{\rho}$.
    For a team spirit $\tau$ in between, team spirit-adjusted rewards are linearly interpolated between the two.
    
    Ultimately we care about optimizing for team spirit $\tau=1$; we want the actions to be chosen to optimize the success of the entire team. However we find that lower team spirit reduces gradient variance in early training, ensuring that agents receive clearer reward for advancing their mechanical and tactical ability to participate in fights individually. 
    See \autoref{sec:methods:exploration} for an ablation of this method.
\end{itemize}

\begin{figure}
    \centering
    \includegraphics[width=0.5\textwidth]{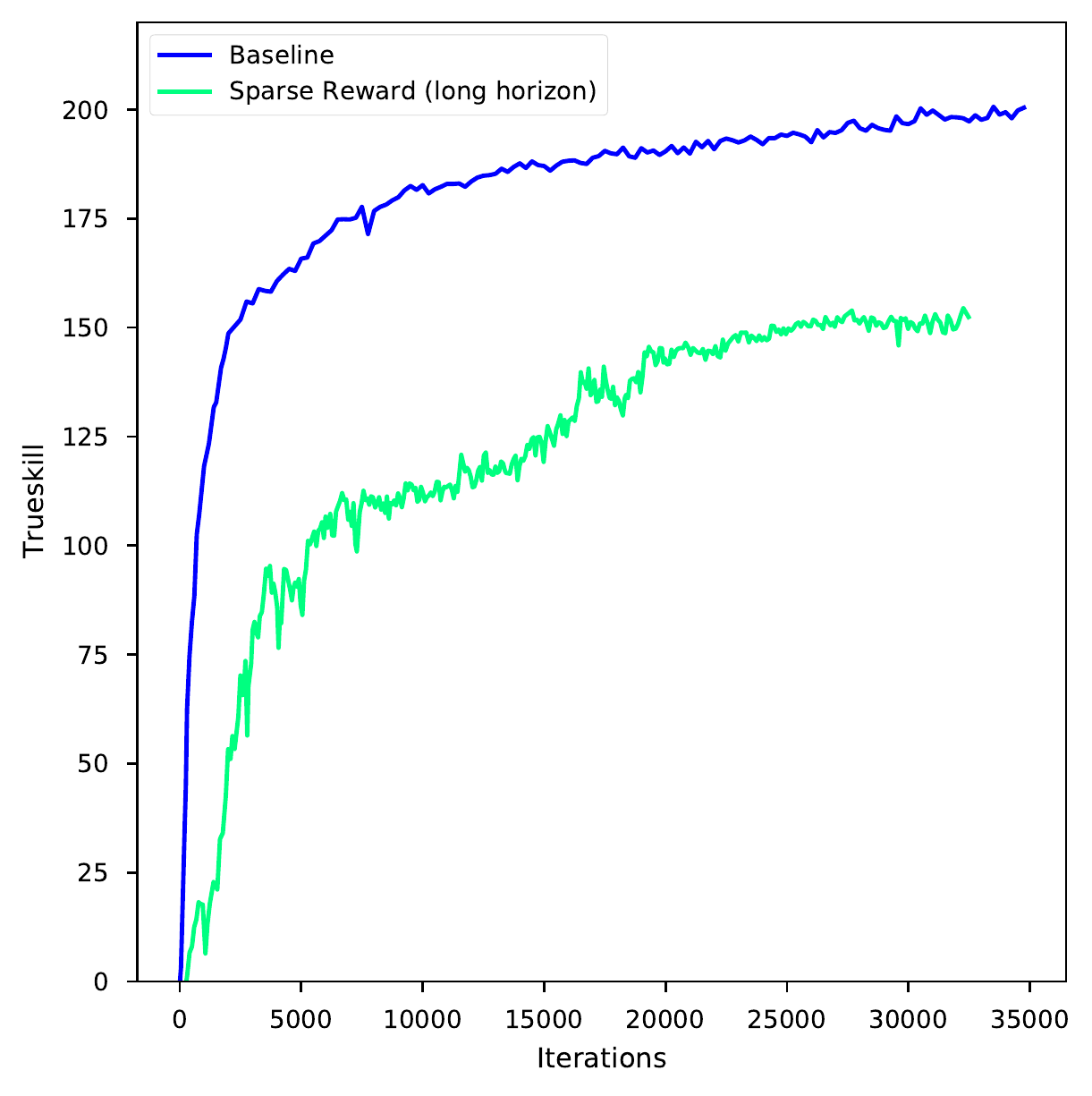}
    \caption{\textbf{Sparse rewards in \dota:} \trueskillname over the course of training for experiments run with 0-1 loss only. For the sparse reward run horizon was set to 1 hour ($\gamma=0.99996$) (versus 180 seconds for the baseline run). 
    The baseline otherwise uses identical settings and hyperparameters including our shaped reward.
    The sparse reward succeeds at reaching \trueskillname 155; for reference, a hand-coded scripted agent reaches \trueskillname 100.
    }
    \label{fig:sparse-rewards}
\end{figure}

We ran a small-scale ablation with partial reward weights disabled (see \autoref{fig:sparse-rewards}). Surprisingly, the model learned to play well enough to beat a hand-coded scripted agent consistently, though with a large penalty to sample efficiency relative to the shaped reward baseline.
From watching these games, it appears that this policy does not play as effectively at the beginning of the game, but has learned to coordinate fights nearer to the end of the game.
Investigating the tradeoffs and benefits of sparse rewards is an interesting direction for future work.

\section{Neural Network Architecture}\label{appendix:network-architecture}

\begin{figure}
\begin{subfigure}{\textwidth}
\includegraphics[width=\textwidth]{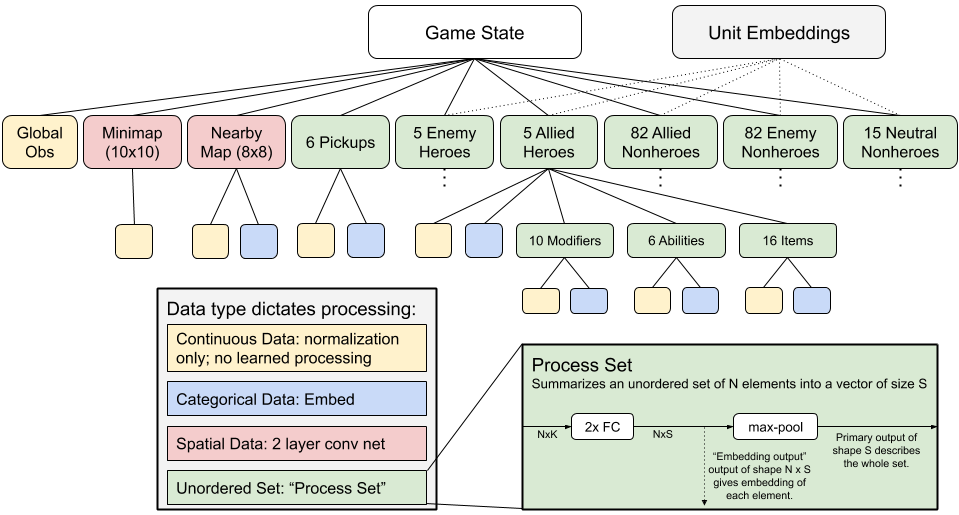}
\caption{\textbf{Flattening the observation space:} First we process the complicated observation space into a single vector. The observation space has a tree structure; the full game state has various attributes such as global continuous data and a set of allied heroes. Each allied hero in turn has a set of abilities, a set of modifiers, etc. We process each node in the tree according to its data type. For example for spatial data, we concatenate the data within each cell and then apply a 2 layer conv net. For unordered sets, a common feature of our observations, we use a ``Process Set'' module. Weights in the Process Set module for processing abilities/items/modifiers are shared across allied and enemy heroes; weights for processing modifiers are shared across allied/enemy/neutral nonheroes. In addition to the main Game State observation, we extract the the Unit Embeddings from the ``embedding output'' of the units' process sets, for use in the output (see \autoref{fig:actionspace}).}
\label{fig:flattening-obs}
\end{subfigure}
\begin{subfigure}{\textwidth}
\begin{center}\includegraphics[width=0.5\textwidth]{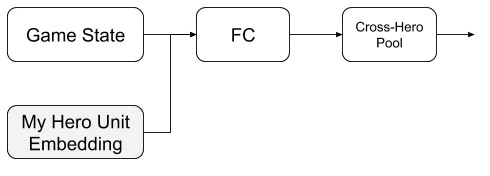}\end{center}
\caption{\textbf{Preparing for LSTM:} 
In order to tell each LSTM which of the team's heroes it controls, we append the controlled hero's Unit Embedding from the Unit Embeddings output of \autoref{fig:flattening-obs} to the Game State vector. 
Almost all of the inputs are the same for each of the five replica LSTMs (the only differences are the nearby map, previous action, and a very small fraction of the observations for each unit). 
In order to allow each replica to respond to the non-identical inputs of other replicas if needed, we add a ``cross-hero pool'' operation, in which we maxpool the first 25\% of the vector across the five replica networks.}
\label{fig:prelstm}
\end{subfigure}
\caption{\textbf{Observation processing in \openaifive}}
\label{fig:observationlogic}
\end{figure}

\begin{figure}
\includegraphics[width=\textwidth]{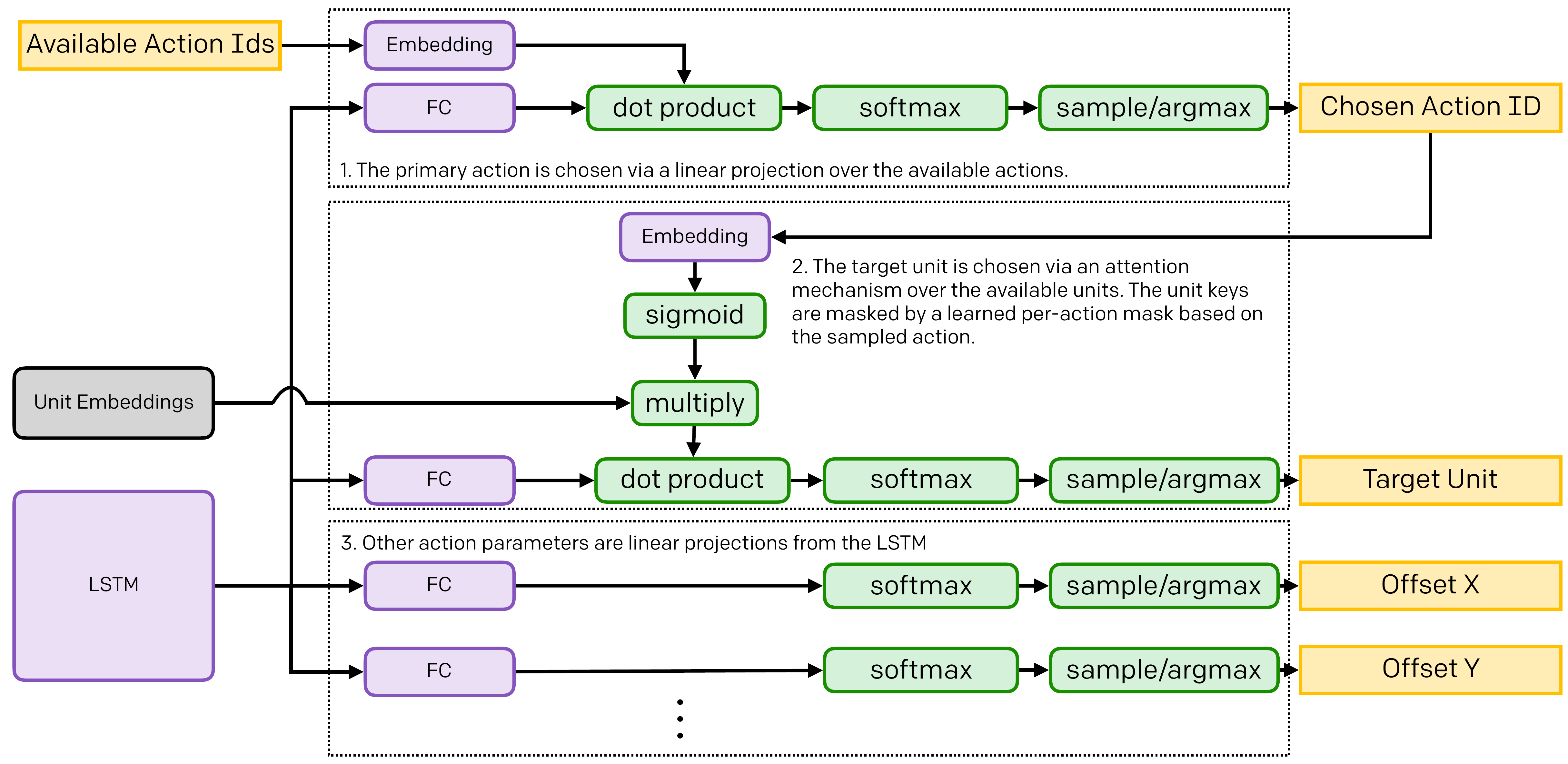}
\caption{The hidden state of the LSTM and unit embeddings are used to parameterize the actions.}
\label{fig:actionspace}
\end{figure}

A simplified diagram of the joint policy and value network is shown in the main text in \autoref{fig:architecture}.
The combined policy + value network uses 158,502,815 parameters (in the final version).

The policy network is designed to receive observations from our bot-API observation space, and interact with the game using a rich factorized action space. 
These structured observation and action spaces heavily inform the neural network architecture used. 
We use five replica neural networks, each responsible for the observations and actions of one of the heroes in the team.
At a high level, this network consists of three parts: first the observations are processed and pooled into a single vector summarizing the state (see \autoref{fig:observationlogic}), then that is processed by a single-layer large LSTM, then the outputs of that LSTM are projected to produce outputs using linear projections (see \autoref{fig:actionspace}).

To provide the full details, we should clarify that \autoref{fig:architecture} is a slight over-simplification in three ways: 
\begin{enumerate}
    \item In practice the Observation Processing portion of the model is also cloned 5 times for the five different heroes. 
    The weights are identical and the observations are nearly identical --- but there are a handful of derived features which are different for each replica (such as ``distance to me'' for each unit; see \autoref{table:full-obs} for the list of observations that vary). 
    Thus the five replicas produce nearly identical, but perhaps not entirely identical, LSTM inputs. 
    These non-identical features form a small portion of the observation space, and were not ablated; it is possible that they are not needed at all.
    \item The ``Flattened Observation'' and ``Hero Embedding'' are processed before being sent into the LSTM (see \autoref{fig:prelstm}) by a fully-connected layer and a ``cross-hero pool'' operation, to ensure that the non-identical observations can be used by other members of the team if needed.
    \item The ``Unit Embeddings'' from the observation processing are carried along beside the LSTM, and used by the action heads to choose a unit to target (see \autoref{fig:actionspace}).
\end{enumerate}

In addition to the action logits, the value function is computed as another linear projection of the LSTM state. 
Thus our value function and action policy share a network and share gradients. 

\newpage

\section{Human Games}\label{appendix:humans}

\begin{table}
\begin{tabular}{c|c|c|c|l}
Opponent & Result & Duration & Version & Restrictions\\
\hline
\multicolumn{5}{l}{June 6, 2018 - Internal Event}\\
\hline
Internal team & win & 15:15 (surr) & 7.13 & Mirror match, multiple couriers, no invis\\
Internal team & win & 20:51 & 7.13 & Mirror match, multiple couriers, no invis\\
Audience team & win & 31:33 & 7.13 & Mirror match, multiple couriers, no invis\\
Audience team & win & 23:33 (surr) & 7.13 & Mirror match, multiple couriers, no invis\\
\hline
\multicolumn{5}{l}{August 5, 2018 - Benchmark}\\
\hline
Caster team & win & 21:38 (surr) & 7.16 & Drafted, multiple couriers\\
Caster team & win & 24:56 (surr) & 7.16 & Drafted, multiple couriers\\
Caster team & lose & 35:47 & 7.16 & Audience draft, multiple couriers\\
\hline
\multicolumn{5}{l}{August 9, 2018 - Private eval}\\
\hline
Team Secret & win & 17:00 (surr) & 7.16 & Drafted, multiple couriers\\
Team Secret & lose & 48:46 & 7.16 & Drafted, multiple couriers\\
Team Secret & lose & 38:55 & 7.16 & Drafted, multiple couriers\\
\hline
\multicolumn{5}{l}{August 22-23, 2018 - The International}\\
\hline
Pain Gaming & lose & 52:29 & 7.19 & Pre-set lineup\\
Chinese Legends & lose & 45:44 & 7.19 & Pre-set lineup\\
\hline
\multicolumn{5}{l}{October 5, 2018 - Private eval}\\
\hline
Team Lithium & win & 48:57 & 7.19 & TI pre-set lineup\\
Team Lithium & win & 48:16 & 7.19 & TI pre-set lineup\\
Team Lithium & win & 31:33 & 7.19 & Drafted\\
\hline
\multicolumn{5}{l}{January 16, 2019 - Private eval}\\
\hline
SG Esports & win & 24:29 (surr) & 7.19 & TI pre-set lineup\\
SG Esports & win & 25:08 (surr) & 7.19 & Drafted\\
SG Esports & win & 27:36 (surr) & 7.20 & Mirror match\\
SG Esports & win & 25:30 (surr) & 7.20 & Mirror match\\
\hline
\multicolumn{5}{l}{February 1, 2019 - Private eval}\\
\hline
Alliance & win & 17:11 & 7.20d & Drafted\\
Alliance & win & 31:33 & 7.20d & Drafted\\
Alliance & win & 28:16 & 7.20d & Reverse drafted\\
\hline
\multicolumn{5}{l}{April 13, 2019 - \openaifinals}\\
\hline
OG & win & 38:18 & 7.21d & Drafted\\
OG & win & 20:51 & 7.21d & Drafted\\
\end{tabular}
\caption{Major matches of \openaifive against high-skill human players.}
\label{table:milestonegames}
\end{table}

See \autoref{table:milestonegames} for a listing of the games \openaifive played against high-profile teams.

\section{\trueskillname: Evaluating a \dota Agent Automatically}\label{appendix:trueskill}
We use the \trueskillname\cite{herbrich2007trueskill} rating system to evaluate our agents. 

We first establish a pool of many reference agents of known skill. 
We evaluate the reference agents' \trueskillname by playing many games between the the various reference agents, and using the outcome of the games to compute a \trueskillname for each agent. 
Our \trueskillname environment use the parameters $\sigma=25/3$, $\beta=\sigma/2$, $\tau=0.0$, $\texttt{draw\_probability}$=0.02.
Reference agents' $\mu$ are aligned so that an agent playing randomly has $\mu=0$.
A hand-crafted scripted agent which we wrote, which can defeat beginners but not amateur players, has \trueskillname around 105.

During our experiments we continually added new reference agents as our agent ``outgrew'' the existing ones.
For all results in this work, however, use a single reference agent pool containing mostly agents from \openaifive's training history along with some other smaller experiments at the lower end. 
The 83 reference agents range in \trueskillname from 0 (random play) to 254 (the version that beat the world champions).

To evaluate a training run during training, a dedicated set of computers continually download the latest agent parameters and plays games between the latest trained agent and the reference agents.
We attempt to only play games against reference agents that are nearby in skill in order to gain maximally useful information; we avoid playing agents more than 10 \trueskillname points away (corresponding to a winrate less than 15\% or more than 85\%).
When a game finishes, we use the \trueskillname algorithm to update the test agent's \trueskillname, but treat the reference agent's \trueskillname as a constant.
After 750 games have been reported, we log that version's \trueskillname and move on to the new current version.
New agents are initialized with $\mu$ equal to the final $\mu$ of the previous agent. 
This system gives us updates approximately once every two hours during running experiments.

One difficulty in using \trueskillname across a long training experiment was maintaining consistent metrics with a changing environment. 
Two agents that were trained on different game versions must ultimately play on a single version of the game, which will result in an inherent advantage for the agent that trained on it. 
Older agents had their code upgraded in order to always be compatible with the newest version, but this still leads to metric inflation for newer agents who got to train on the same code they are evaluated on. 
This included any updates to the hero pool (adding new heroes that old agents didn't train with), game client updates or balancing changes, and adding any new actions (using a particular consumable or item differently).

\section{\dota Gym Environment}
\label{appendix:gym-ocean}
\subsection{Data flow between the training environment and \dota}

\dota includes a scripting API designed for building bots. The provided API is exposed through Lua and has methods for querying the visible state of the game as well as submitting actions for bots to take. Parts of the map that are out of sight are considered to be in the fog of war and cannot be queried through the scripting API, which prevents us from accidentally ``cheating'' by observing anything a human player would not be able to see (although see~\autoref{appendix:bloopers}).

We designed our \dota environment to behave like a standard OpenAI Gym environment\cite{brockman2016gym}. This standard respects an API contract where a {\tt step} method takes action parameters and returns an observation from the next state of the environment. To send actions to \dota, we implemented a helper process in Go that we load into \dota through an attached debugger that exposes a gRPC server. This gRPC server implements methods to configure a game and perform an environment step. By running the game with an embedded server, we are able to communicate with it over the network from any remote process.

When the step method is called in the gRPC server, it gets dispatched to the Lua code and then the method blocks until an observation arrives back from Lua to be returned to the caller. In parallel, the \dota engine runs our Lua code on every step, sending the current game state observation\footnote{Originally the Lua scripting API was used to iterate and gather the visible game state, however this was somewhat slow and our final system used an all-in-one game state collection method that was added through cooperation with Valve} to the gRPC server and waiting for it to return the current action. The game blocks until an action is available. These two parallel processes end up meeting in the middle, exchanging actions from gRPC in return for observations from Lua. Go was chosen to make this architecture easy to implement through its channels feature.

Putting the game environment behind a gRPC server allowed us to package the game into a Docker image and easily run many isolated game instances per machine. It also allowed us to easily setup, reset, and use the environment from anywhere where Docker is running. This design choice significantly improved researcher productivity when iterating on and debugging this system.
\nicetohave{(jie) I think it would be extremely interesting to get timings for different parts of this process. emphasizes complexity of the environment}

\section{Reaction time}
\label{appendix:reaction-time}
\begin{figure}
    \centering
    \includegraphics[width=0.7\textwidth]{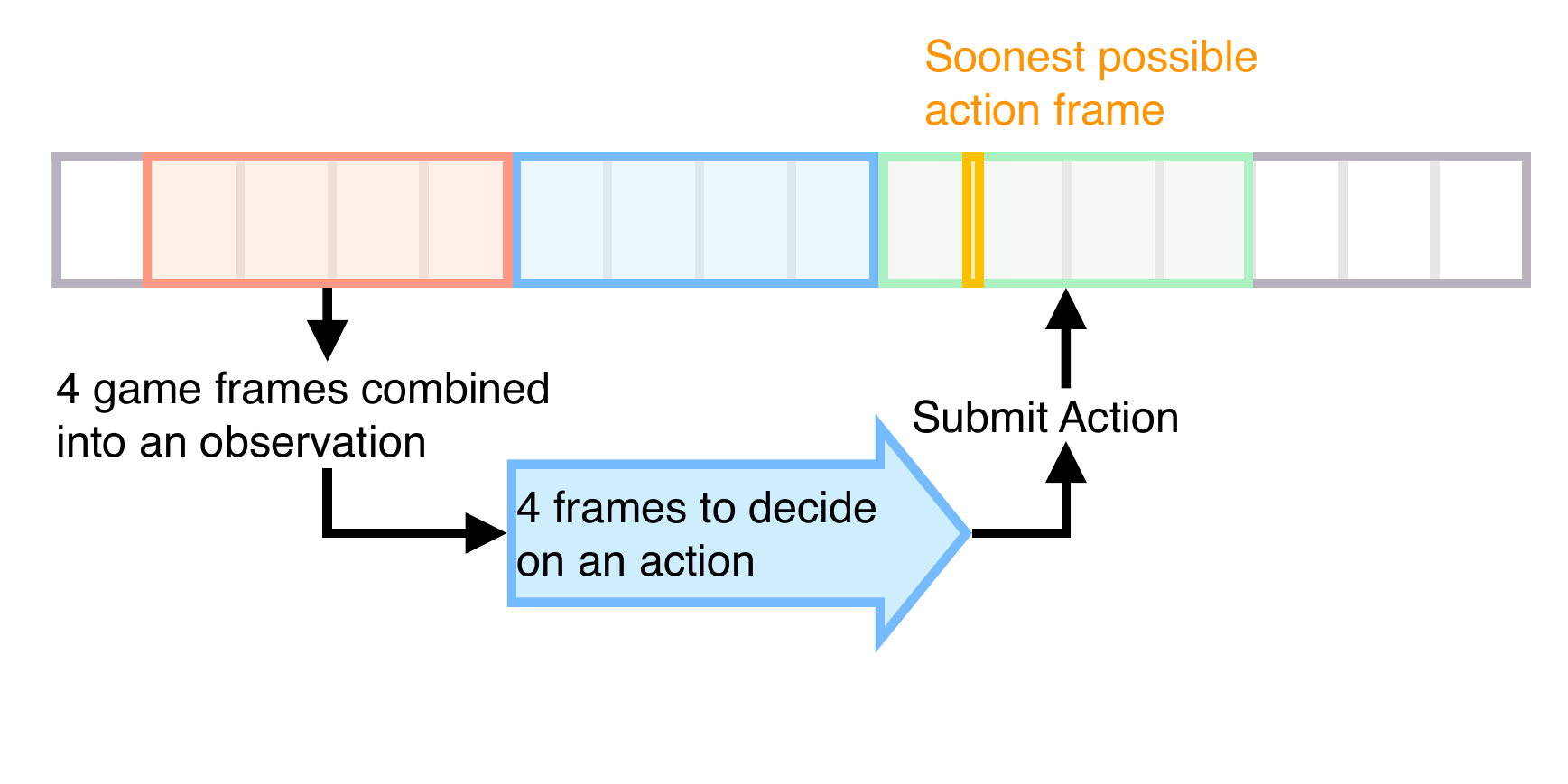}
    \caption{\textbf{Reaction Time: } OpenAI Five observes four frames bundled together, so any surprising new information will become available at a random frame in the red region. 
    The model then processes the observation in parallel while the game engine runs forward four more frames.
    The soonest it can submit an action based on the red observations is marked in yellow.
    This is between 5 and 8 frames (167-267ms) after the surprising event.
    }
    \label{fig:reaction-time}
\end{figure}

The \dota game engine runs at 30 steps per second so in theory a bot could submit an action every 33ms.
Both to speed up our game execution and in order to bring reactions of our model closer to the human scale we downsample to every 4th frame, which we call {\it frameskip}.
This yields an effective observation and action rate of 7.5 frames per second.
To allow the model to take precisely timed actions, the action space includes a ``delay'' which indicates which frame during the frameskip the model wants this action to evaluate on.
Thus the model can still take actions at a particular frame if so desired, although in practice we found that the model did not learn to do this and simply taking the action at the start of the frameskip was better.

Moreover, we reduce our computational requirements by allowing the game and the machine learning model to run concurrently by asynchronously issuing actions with an {\it action offset}. 
When the model receives an observation at time $T$, rather than making the game engine wait for the model to produce an action at time $T$, we let the game engine carry on running until it produces an observation at time $T+1$. 
The game engine then sends the observation at time $T+1$ to the model, and by this time the model has produced its action choice based on the observation at time $T$.
In this way the action which the model takes at time $T+1$ is based upon the observation at time $T$.
In exchange for this penalty in available ``reaction time,'' we are able to utilize our compute resources much more efficiently by preventing the two major computations from blocking one another
(see \autoref{fig:reaction-time}).

Taken together, these effects mean that the agent can react to new information with a reaction time randomly distributed between 5 and 8 frames (167ms to 267ms), depending on when during the frameskip the new information happens to occur.
For comparison, human reaction time has been measured at 250ms in controlled experimental settings\cite{jain2015reaction}. This is likely an underestimate of reaction time during a Dota game.

\section{Scale and Data Quality Ablation Details}
\label{sec:methods}\label{appendix:substudies}

As shown in \autoref{fig:lotsa-results} of the main text, we studied several key ingredients of RL at this scale, and learned important lessons which we conjecture should generalize beyond this environment. 
In this section we explain the details of these experiments. 

\begin{figure}
    \centering
    \begin{subfigure}{0.49\linewidth}
        \includegraphics[width=\textwidth]{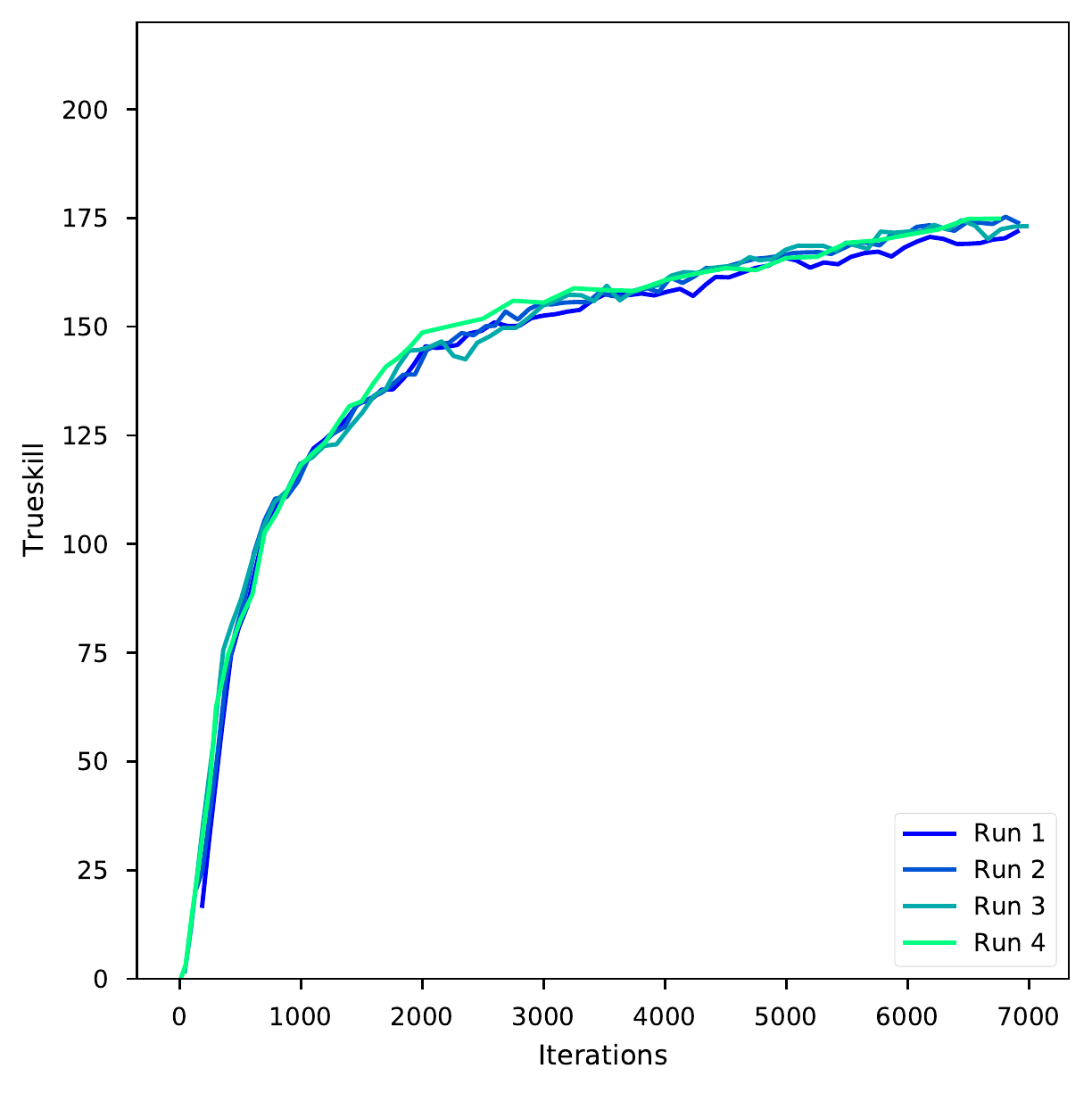}
    \end{subfigure}
    \begin{subfigure}{0.49\linewidth}
        \includegraphics[width=\textwidth]{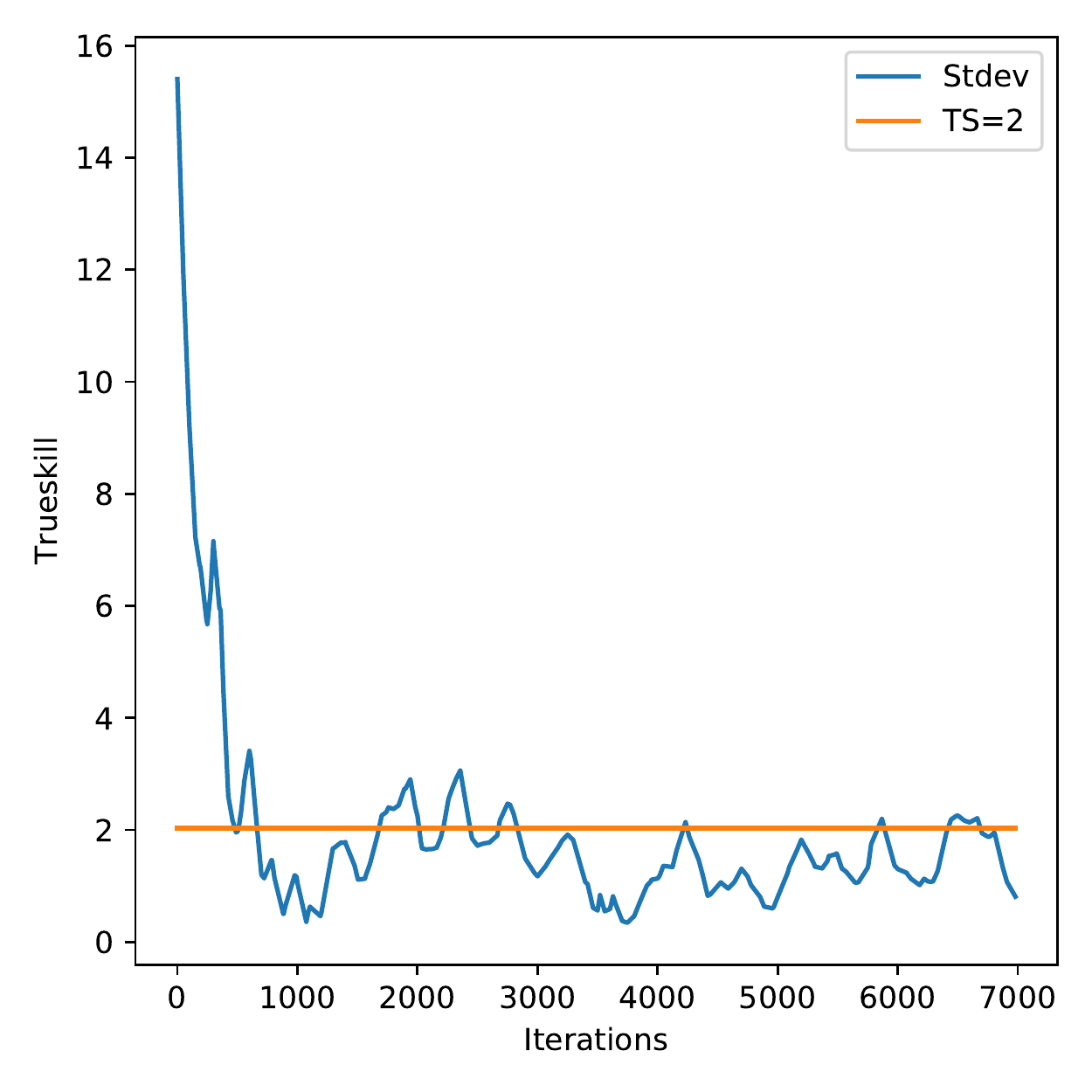}
    \end{subfigure}
    \caption{\textbf{Variation in 5v5 baseline training:} On the left, the \trueskillname over the course of training for different ``baseline'' experiments, using identical settings and hyperparameters. On the right, the standard deviation in \trueskillname across four runs. See \autoref{appendix:hyperparams} for the hyperparameters used. Although we only have 4 runs, we can estimate that different runs tend to vary by about 2 \trueskillname.}
    \label{fig:multiple-baselines}
\end{figure}

Training runs the size of \openaifive are expensive; running a scan of 4 different variants would be prohibitively expensive. For this reason we use the normal \dota environment, simply using a batch size 8x smaller than \cleanexpname (which itself was 2-3 times smaller than \openaifive). 
See \autoref{fig:multiple-baselines} for an estimate of the variation in these training runs.

Throughout the following sections we scan over various parameters of the experimental setup and monitor the results in terms of \trueskillname (see \autoref{appendix:trueskill}) and speedup (see \autoref{eqn:speedup}).

Our the uncertainty on speedup comes from uncertainty in both the numerator and the denominator. 
Although we have some understanding in the variance in the number of iterations for a baseline to reach each \trueskillname (see \autoref{fig:multiple-baselines}), we do not have the luxury of multiple runs of every experiment.
Instead, we use as proxy for the uncertainty on the number of iterations to reach \trueskillname $T$, the number of iterations to reach to reach $T\pm\Delta T$ where $\Delta T$ is the variance in \trueskillname across the variations in \autoref{fig:multiple-baselines}, approximately 2 \trueskillname points.
We combine the numerator and denominator uncertainty in quadrature to attain an overall uncertainty for the speedup.

In each experiment the baseline uses hyperparameters given in \autoref{appendix:hyperparams}, except as noted.

\subsection{Batch Size}\label{sec:methods:batchsize}



Training using small mini-batches is a generally accepted trade-off between convergence time and number of optimization steps. 
However, recent literature on large-scale supervised learning of image classifiers \cite{Goyal2017Imagenet, You2017Imagenet, You2018minutenet} explored much larger batch sizes and showed that strong scaling was possible by carefully tuning learning rate and initialization of the neural network. 
This renewed interest in reducing convergence-time and treating batch-size as a key design parameter also motivated the work of \cite{mccandlish2018empirical}, where an analytical tool is derived to estimate a training-time optimal batch size on per task basis by studying the ``noise scale'' of the gradients.

While existing literature on large-scale training of neural networks had focused on supervised learning, as far as we know using large batch sizes for reinforcement learning was novel when we began the \dota project.
These observations were later shown to be consistent with the analytical tools derived in \cite{mccandlish2018empirical}. 
In this section we demonstrate how large batch-sizes affect optimization time.

Because we average gradients across the pool of optimizer machines, the effective total batch size is given by the product of the number of GPU optimizers with the batch size on each optimizer. 
We always use the maximum batch size on each optimizer which will fit within the GPU's memory constraints (120 for our setup). 
Thus in order to change the overall batch size we increase the number of optimizer GPUs. 
We increase the size of the other machine pools in the experiment (rollout CPU workers, forward pass GPUs, etc), such that the larger batch size experiment is truly optimizing over more data, not simply reusing the same data more. 
This means that doubling the batch size causes the experiment to use twice as much computing power in almost all respects.
Because we do not have the resources to separately optimize these hyperparameters at each individual batch size, we keep all other hyperparameters fixed to those listed under ``baseline'' in \autoref{table:hyperparams}.



\begin{figure}
    \centering
    \begin{subfigure}[t]{0.49\linewidth}
        \vskip 0pt 
        \includegraphics[width=\textwidth]{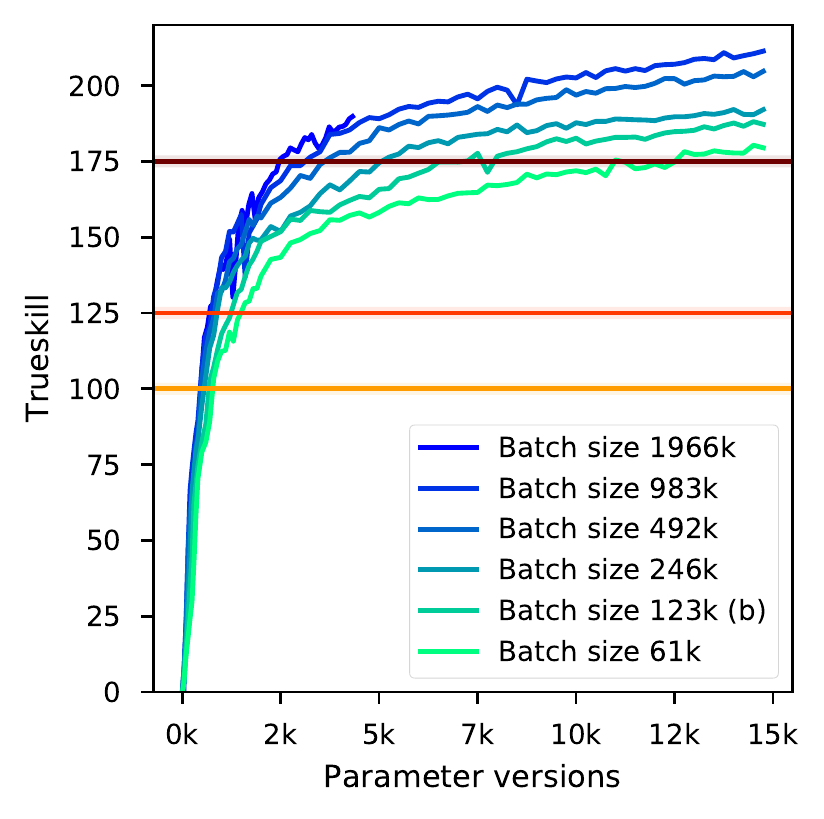}
    \end{subfigure}
    \begin{subfigure}[t]{0.49\linewidth}
        \vskip 0pt 
        \includegraphics[width=\textwidth]{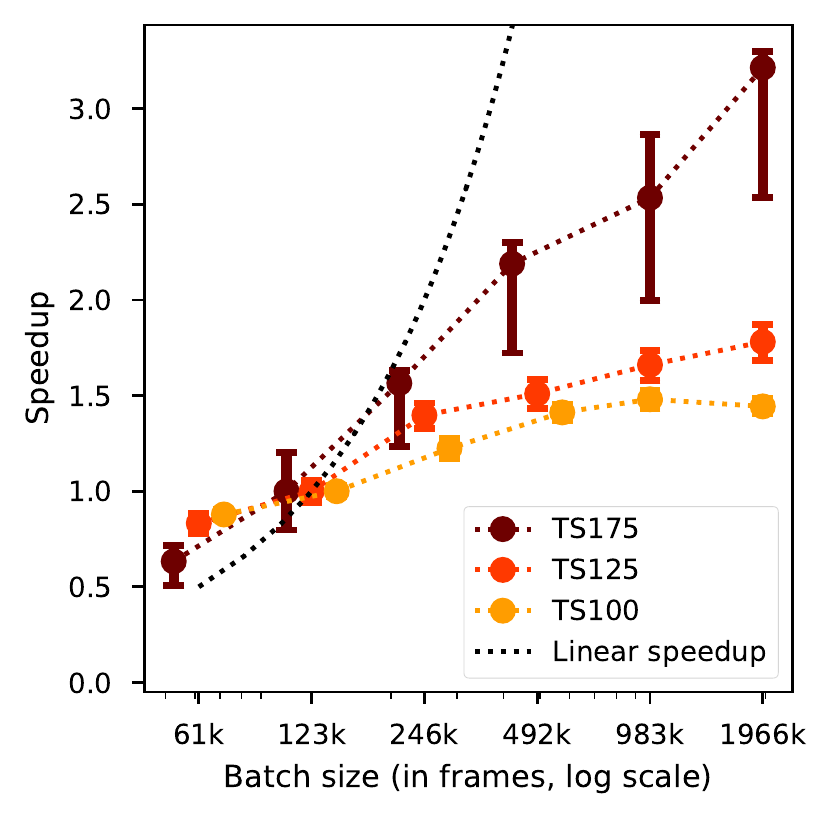}
    \end{subfigure}
    \caption{\textbf{Effect of batch size on training speed:} (Replicated from main text \autoref{subfig:batch-size}) \genericcaptionsentence{increasing the batch size.} The dotted line indicates perfect linear scaling (using 2x more data gives 2x speedup). 
    Larger batch size significantly speeds up training, but the speedup is sublinear in the resources consumed. 
    Later training (\trueskillname 175) benefits more from increased scale than earlier training (\trueskillname 100). 
    Note that \trueskillname 175 is still quite early in the overall training of \openaifive which ultimately reaches above 250 (see \autoref{fig:final-ts}), so these results are inconclusive about whether large batch size causes linear speedup for the bulk of the training time.}
    \label{fig:batchsize}
\end{figure}

Results can be seen in \autoref{subfig:batch-size}, with discussion in the main text. 
\subsection{Sample Quality --- Staleness}\label{sec:methods:staleness}

In an ideal world, each piece of data in the optimizer would be perfectly on-policy (to obtain unbiased gradients), would be used exactly once and then thrown out (to avoid overfitting), would be from a completely different episode than every other piece of data (to eliminate correlations), and more.
Because of our enormous batch size and small learning rate, we hypothesized that loosening the above constraints would not be a large price to pay in exchange for the benefits of asynchronous processing.
However, we actually learned that issues like this surrounding data quality can be quite significant.
In this and next section we will focus on two of these issues, which we call {\it staleness} and {\it sample reuse}.

Early on in the development of our agent we would play the whole game of \dota using single set of parameters, then send this huge package of sample data to optimizers for training.
One of the negative effects of this approach was that this would render data stale; the policy parameters which played the start of the game would be an hour old or more, making the gradients estimated from them incorrect.
Therefore we have switched to accumulating small amount of training data; sending it over to optimizers and updating agent parameters; then continuing with the same game.

In order to generate rollouts with a certain version of the parameters, a long round-trip has to happen (see \autoref{fig:training-architecture}).
This new set of parameters is published to the controller, then independently pulled by forward pass machines, which only then will start using this version of parameters to perform forward-passes of our agent.
Then some amount of gameplay must be rolled forward and after that the data is finally sent to the optimizers.
In the meanwhile, the optimizers have been running on previously-collected data and advanced by some number of new gradient descent steps.
In our setup where rollouts send about 30 seconds of gameplay in each chunk, this loop takes 1-2 minutes.
Because our learning rate is small, and this is only a few minutes on the scale of a multi-week learning endeavor, one might expect this to be a minor concern --- but to the contrary, we observe that this it can be a crucial detail.

\begin{figure}
    \centering
    \includegraphics[width=0.5\textwidth]{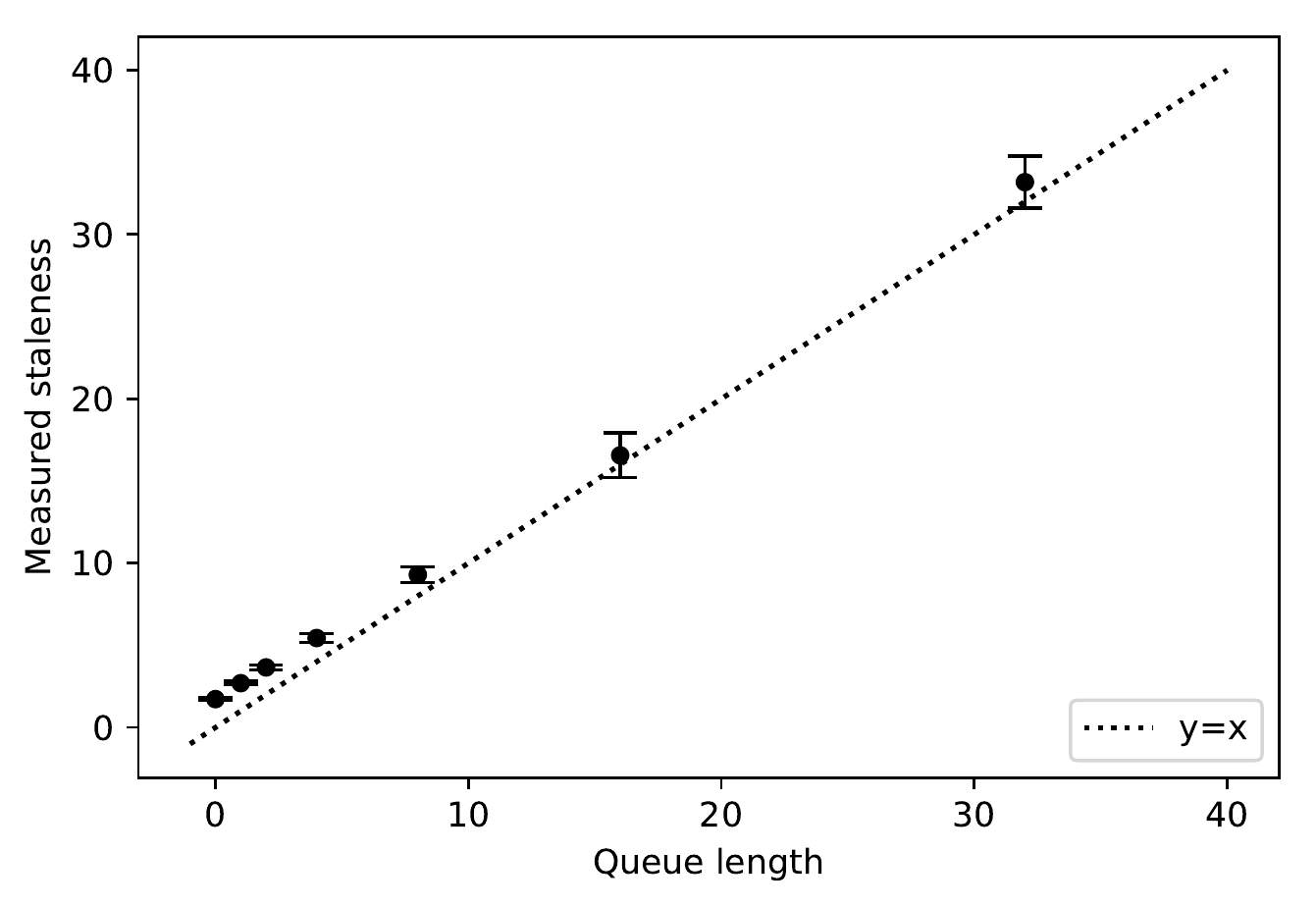}
    \caption{Adding a queue that buffers rollout data on the way to optimizers increases measured staleness in a predictable manner. Error bars indicate the standard deviation of measured staleness as it varied over the course of training due to distributed systems fluctuations.
    }
    \label{fig:full-queue-to-freshness}
\end{figure}

In this study we artificially introducing additional delay to see the effect.
This is implemented on the rollout workers; instead of sending their data immediately back to the optimizers, they now put it in a queue, and pop data off the end of it to send to the optimizers.
Thus the length of the queue determines the amount of artificial staleness introduced. See \autoref{fig:full-queue-to-freshness}; we observe the desired increase in measured staleness with the length of the queue.

\begin{figure}
    \centering
    \begin{subfigure}[t]{0.45\textwidth}
        \centering
        \includegraphics[width=\textwidth]{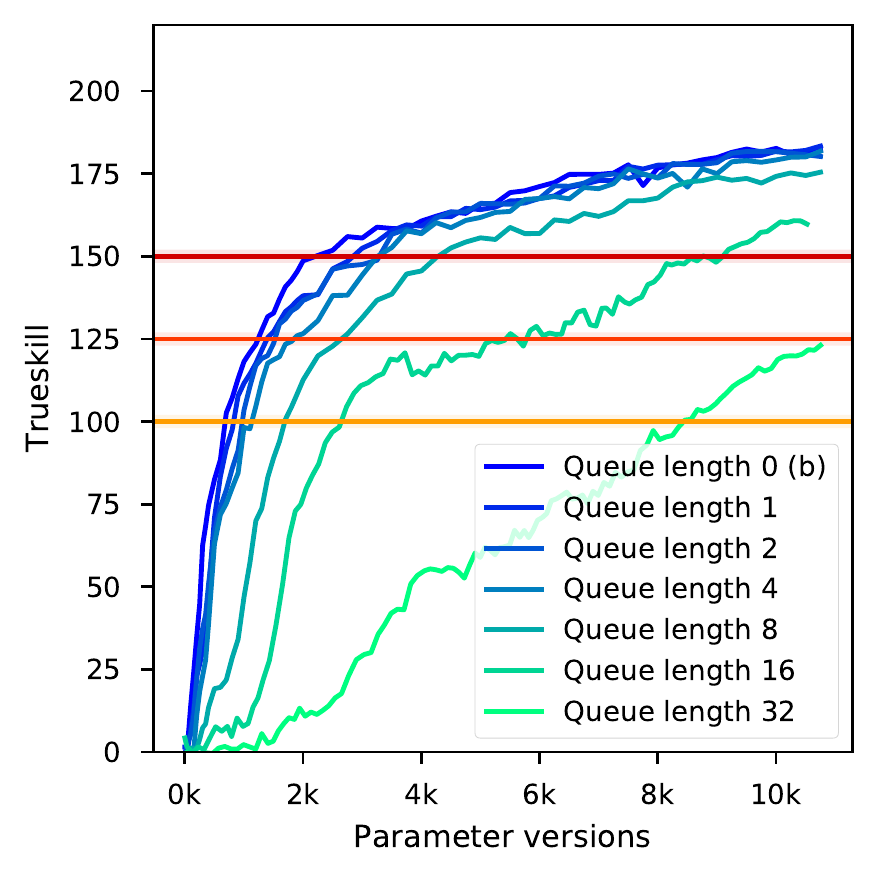}
        \label{fig:full-staleness}
    \end{subfigure}\hfill
    \begin{subfigure}[t]{0.45\textwidth}
        \centering
        \includegraphics[width=\textwidth]{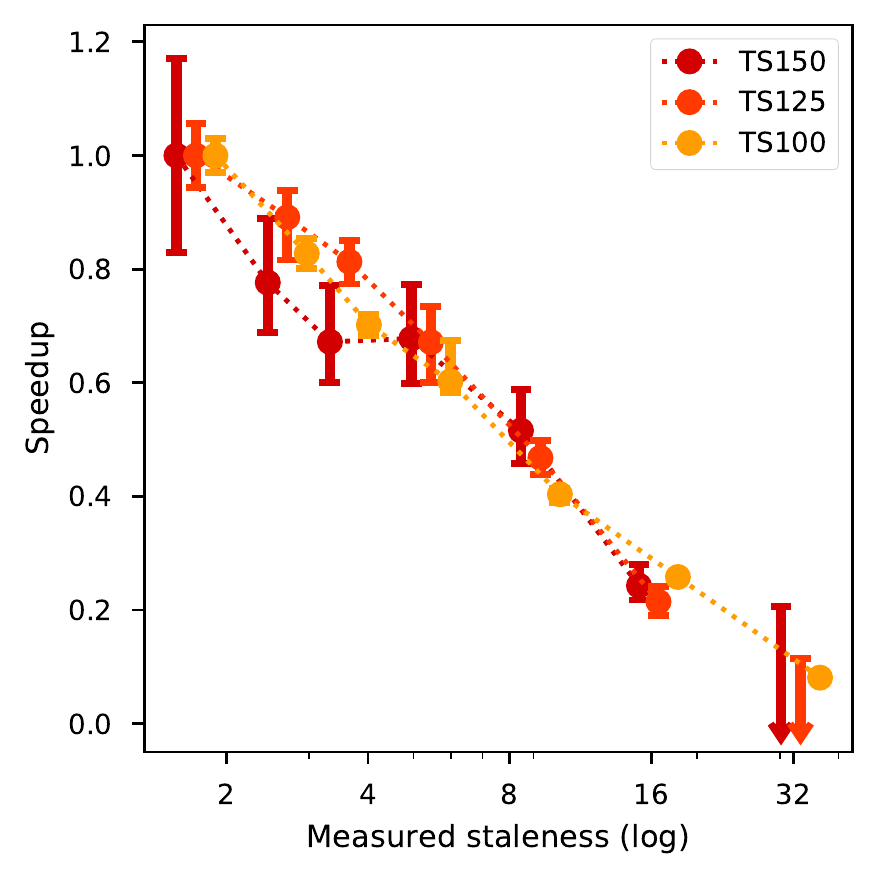}
        \label{fig:full-staleness-speedup}
    \end{subfigure}
    \caption{\textbf{Effect of Staleness on training speed.}
    (Replicated from main text \autoref{subfig:staleness}) \genericcaptionsentence{increasing Staleness.} Increasing staleness of data causes significant losses in training speed.}
    
    \label{fig:staleness}
\end{figure}

The results can be found in the main text in \autoref{subfig:staleness}, and are reproduced in \autoref{fig:staleness}. 
Staleness negatively affects speed of training, and the drop can be quite severe when the staleness is larger than a few versions. 
For this reason we attempt to keep staleness as low as possible in our experiments.

\subsection{Sample Quality --- Sampling and Sample Reuse}\label{sec:methods:samplereuse}

Our asynchronous training system reuses samples in multiple optimization steps. 
Each optimizer's experience buffer is constantly asynchronously collecting data from rollout machines. 
At each optimization step, a batch of data is sampled from this buffer. The buffer is configured to hold 4096 samples. 
Our optimizers compute the average sample reuse as the ratio between data arrival and consumption rates:
\begin{equation}\label{eqn:sample-reuse}
    \textrm{Sample Reuse} \equiv \frac
    {\left(\textrm{samples per batch}\right) \times \left(\textrm{batches per second}\right)}
    {\left(\textrm{experience buffer intake samples per second}\right)}
\end{equation}


Sample reuse is a function of the round trip time between rollout machines and optimizers, the ratio of rollout machines to optimizers, and other factors, and thus we only approximately hit target values but do not set them exactly. 
We measure the effect of sample reuse by varying the rate of incoming samples to the optimizers. 
In practice, the rate of data production from each rollout worker stays relatively stable, so we vary this rate by changing the number of rollout CPU workers and forward pass GPUs while keeping the number of optimizers and everything else fixed.

\begin{figure}
    \centering
    \includegraphics[width=0.5\textwidth]{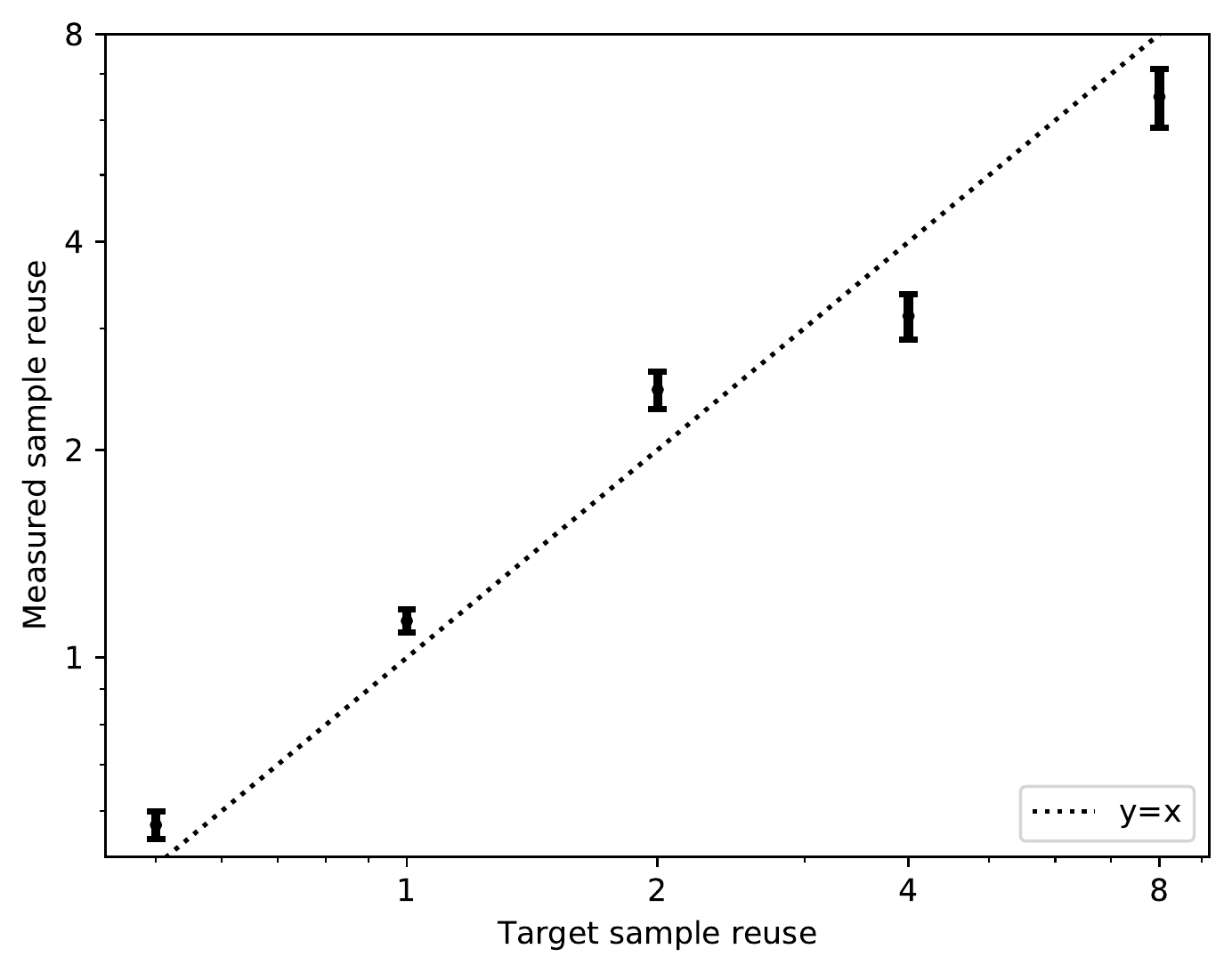}
    \caption{As our target sample reuse increases measured sample reuse increases predictably. Error bars indicate the standard deviation of measured sample reuse as it varied over the course of training.
    }
    \label{fig:samplereuse:scale-effect}
\end{figure}

Our baseline experiment is tuned to have a sample reuse of approximately 1. 
To measure the effect of sample reuse we reduced the number of rollouts by 2, 4, and 8x to induce higher sample reuse. 
Additionally we also doubled the number of rollouts for one experiment to investigate the regime where sample reuse is lower than 1. 
These adjustments yielded sample reuse measurements between 0.57 and 6.3 (see \autoref{fig:samplereuse:scale-effect}). 
It is important to highlight that adjusting the number of rollouts directly affects the number of simultaneous games being played, which affects the diversity of games that are used for training.

\begin{figure}
    \centering
    \begin{subfigure}[t]{0.49\linewidth}
        \vskip 0pt 
        \includegraphics[width=\textwidth]{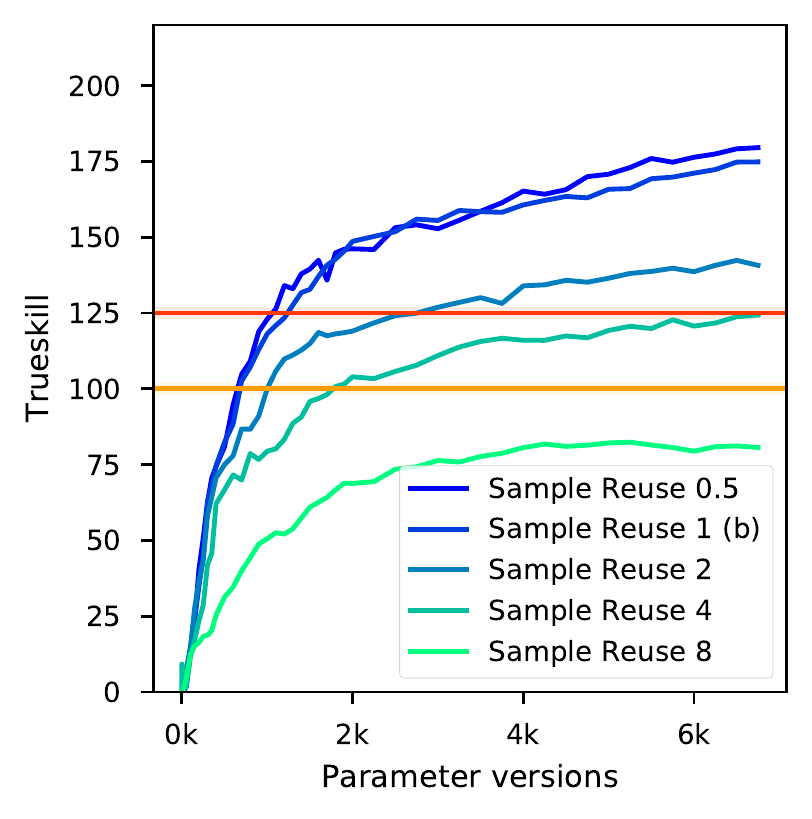}
    \end{subfigure}
    \begin{subfigure}[t]{0.49\linewidth}
        \vskip 0pt 
        \includegraphics[width=\textwidth]{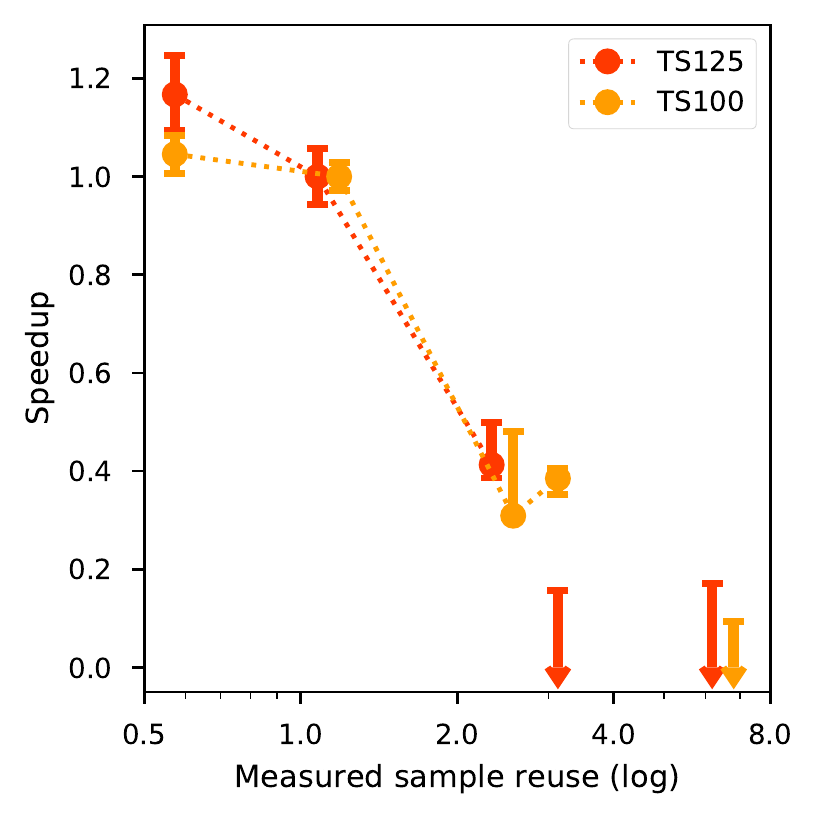}
    \end{subfigure}
    \caption{\textbf{Effect of Sample Reuse on training speed.}
    (Replicated from main text \autoref{subfig:sample-reuse})
    \genericcaptionsentence{increasing Sample Reuse.} Increasing sample reuse causes significant slowdowns. In fact, the run with 1/8th as many rollout workers (sample reuse around 6.3), seems to have converged to less than 75 \trueskillname.}
    \label{fig:samplereuse}
\end{figure}

The results can be found in the main text in \autoref{subfig:sample-reuse}, and are reproduced in \autoref{fig:samplereuse}. 
We found that increasing sample reuse causes a significant decrease in performance. 
As long as the optimizers are reusing data, adding additional rollout workers appears to be a relatively cheap way to accelerate training.
CPUs are often easier and cheaper to scale up than GPUs and this can be a significant performance boost in some setups.

The fact that our algorithms benefit from extremely low sample reuse underlines how sample inefficient they are. 
Ideally, our training methods could take a small amount of experience and use that to learn a great deal, but currently we cannot even usefully optimize over that experience for more than a couple of gradient steps. 
Learning to use rollout data more efficiently is one of the major areas for future work in RL research.

This investigation suggests that sample reuse below one can be beneficial. 
This experiment out performed all others after around iteration 5,000, including the experiment with sample reuse 1.
The improvement over sample reuse 1 is minor compared to the gaps between more severe sample reuses, but it is significant.
Intuitively one might expect that using each sample exactly once would be the most optimal, as no data would get wasted and no data would get used twice; collecting more data and then not optimizing over it would not help.

However, the sample reuse is measured as an average rate of data production to consumption (\autoref{eqn:sample-reuse}).
Because the optimizers sample each batch randomly from the buffer, sample reuse 1 just means that on \emph{average} each sample is used once, but in fact many samples are used twice, and some not used at all.
For this reason producing twice as much data as we can consume still reduces the number of samples which get selected multiple times. 
Of course the magnitude of improvement is relatively small and the cost (doubling the number of rollout workers and forward pass GPUs) is significant.
Doubling the number of rollout workers may also decrease correlation across samples;
using two adjacent samples from the same game (when very little has changed between them) may have similar drawbacks to using the same sample twice.

\begin{figure}
    \centering
    \begin{subfigure}{0.49\linewidth}
        \includegraphics[width=\textwidth]{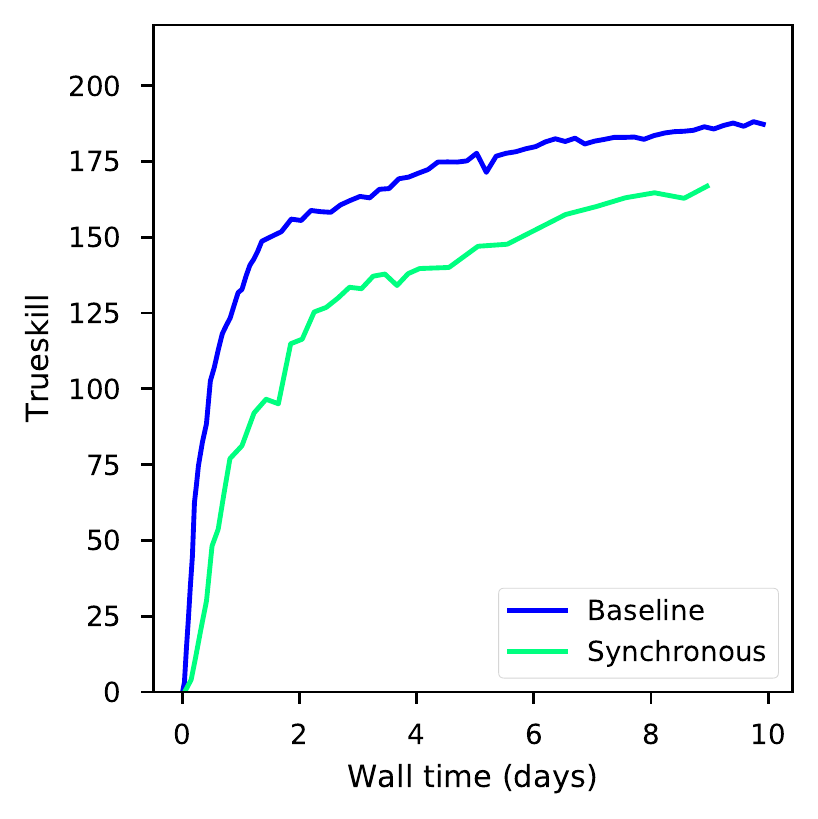}
    \end{subfigure}
    \begin{subfigure}{0.49\linewidth}
        \includegraphics[width=\textwidth]{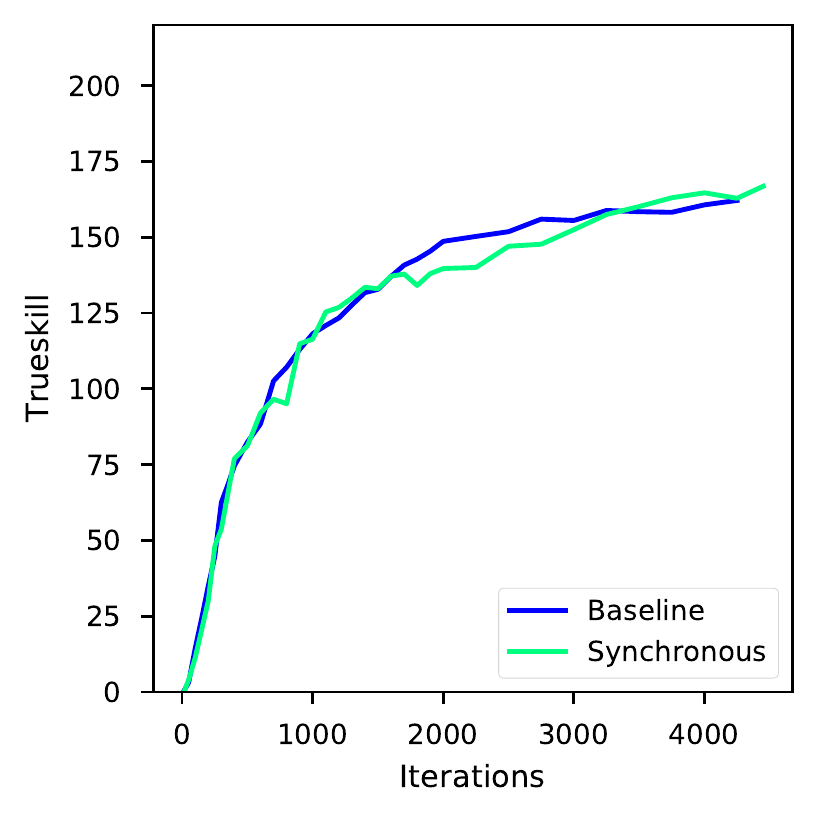}
    \end{subfigure}
    \caption{\textbf{Asynchronous training:} Plots of \trueskillname over the course of training for a ``baseline'' experiment together with a ``synchronous'' run using only on-policy data (staleness = 0) and restricting each sample to be used at most once (max sample reuse = 1). On the left, the x-axis is wall time. On the right, the x-axis is iterations. Asynchronous training is nearly 3x faster at achieving \trueskillname 150 when measuring by wall time, even though the two runs perform similarly as a function of the number of iterations.}
    \label{fig:sync-async}
\end{figure}

\section{Self-play}
\label{sec:selfplay}

  \begin{figure}
     \centering
     \includegraphics[width=\textwidth]{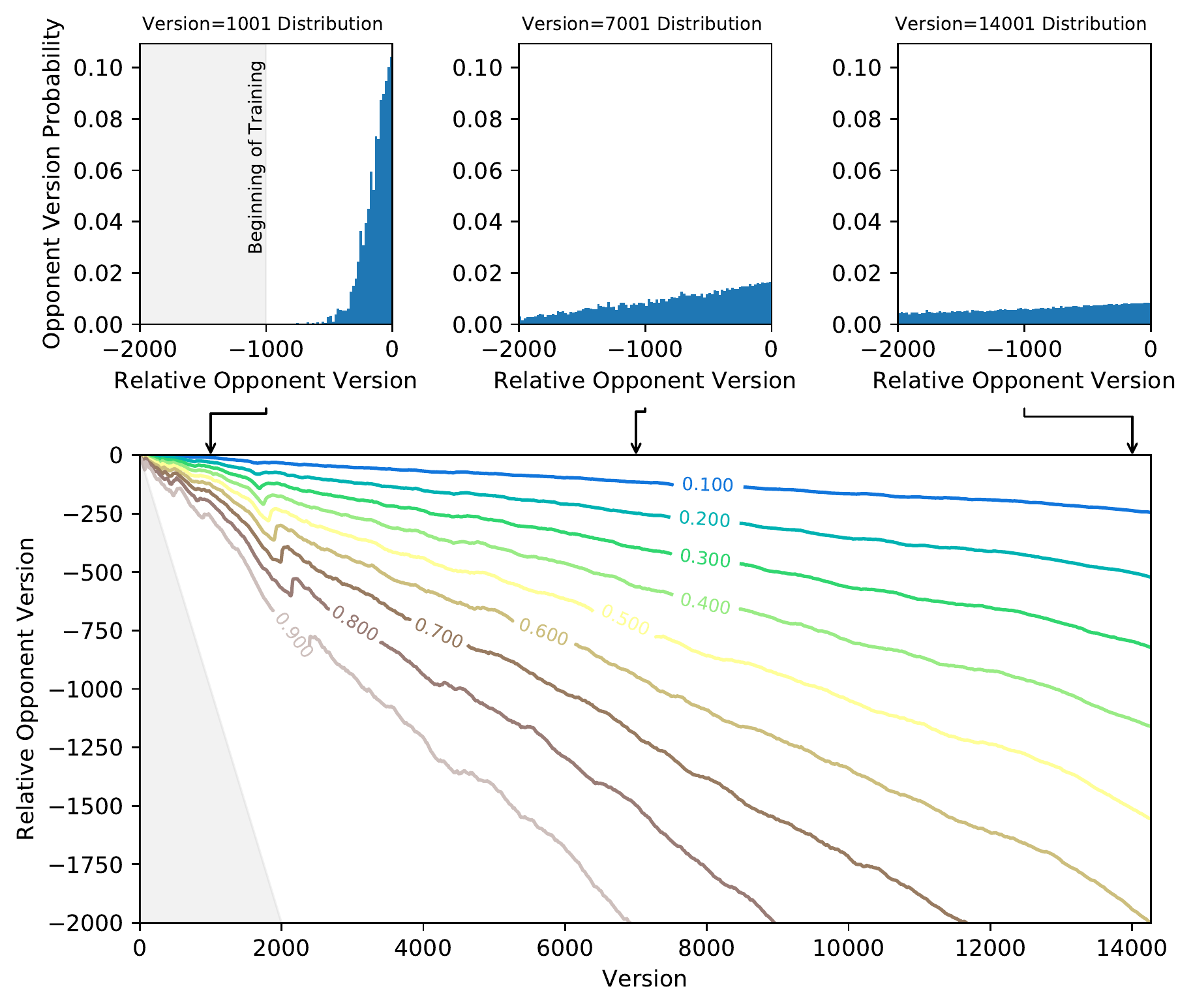}
     \caption{\textbf{Opponent Manager Distribution over past versions.} As the performance of the agent improves, the distribution over past versions changes to find stronger contenders. The slope in the distribution reflects how fast the current agent is outpacing previous versions: a slow falloff indicates that the agent is still challenged by far older versions, while a steep falloff is evidence that counter-strategies have been found that eliminate past agents. In later versions the opponent distribution includes many more past versions, suggesting that after a warmup period, skill progression slows.}
     \label{fig:selfplay}
 \end{figure}

\openaifive is trained without any human gameplay data through a self-improvement process named {\it self-play}. 
This technique was successfully used in prior work to obtain super human performance in a variety of multiplayer games including Backgammon, Go, Chess, Hex, StarCraft 2, Poker  \cite{tesauro1994td,silver2016mastering,anthony2017thinking,silver2017mastering,alphastarblog,brown2019superhuman}. 
In self-play training, we continually pit the current best version of an agent against itself or older versions, and optimize for new strategies that can defeat these past and present opponents.

In training \openaifive 80\% of the games are played against the latest set of parameters, and 20\% play against past versions.
We play occasionally against past parameter versions in order to obtain more robust strategies and avoid {\it strategy collapse} in which the agent forgets how to play against a wide variety of opponents because it only requires a narrow set of strategies to defeat its immediate past version (see \citet{DBLP:journals/corr/abs-1901-08106} for a discussion of cyclic strategies in games with simultaneous-turns and/or imperfect information).

\openaifive uses a dynamic sampling system in which each past opponent $i=1..N$ is given a {\it quality} score $q_i$. 
Opponent agents are sampled according to a softmax distribution; agent $i$ is chosen with probability $p_i$ proportional to $e^{q_i}$.
Every 10 iterations we add the current agent to past opponent pool and initialize its quality score to the maximum of the existing qualities.
After each rollout game is completed, if the past opponent defeats the current agent, no update is applied. If the current agent defeats a past opponent, an update is applied proportional to a learning rate constant $\eta$ (which we fix at 0.01):
\begin{equation}
    q_i \leftarrow q_i - \frac{\eta}{N p_i}
\end{equation}


In \autoref{fig:selfplay} we see the opponent distribution at several points in early training. 
The spread of the distribution gives a good picture of how quickly the agent is improving: when the agent is improving rapidly, then older opponents are worthless to play against and have very low scores; when progress is slower the agent plays against a wide variety of past opponents.

\section{Exploration}\label{sec:methods:exploration}

Exploration is a well-known and well-researched problem in the context of reinforcement learning.\nicetohave{citations-exploration}
We encourage exploration in two different ways: by shaping the loss (entropy and team spirit) and by randomizing the training environment.

\subsection{Loss function}
\begin{figure}
    \centering
    \begin{minipage}{0.45\textwidth}
        \centering
        \includegraphics[width=\textwidth]{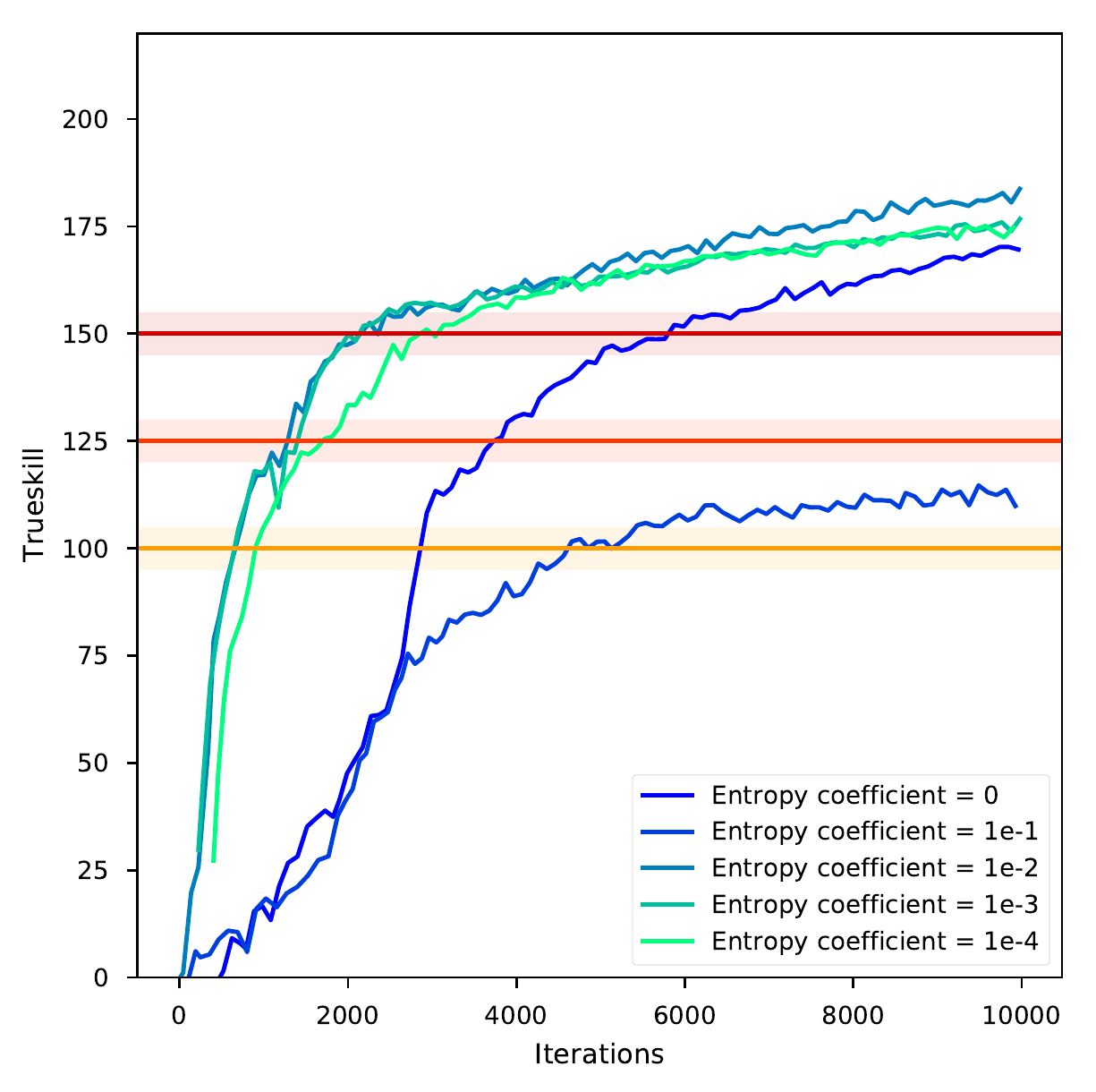}
    \end{minipage}\hfill
    \begin{minipage}{0.45\textwidth}
        \centering
        \includegraphics[width=\textwidth]{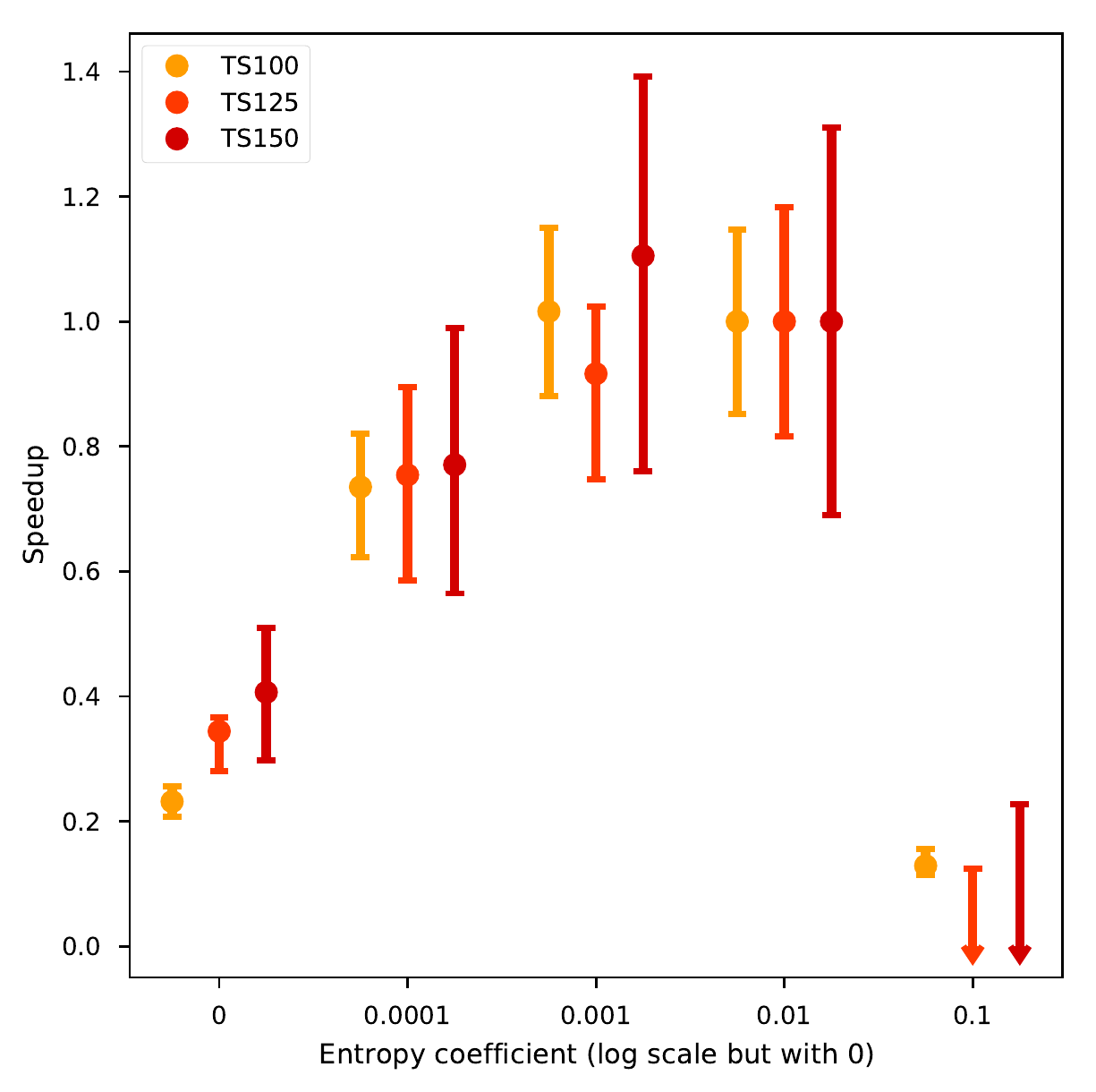}
    \end{minipage}
    \caption{\textbf{Entropy in early training: } \trueskillname and speedup with varied entropy coefficients.
    Lower entropy performs worse because the model has a harder time exploring; higher entropy performs much worse because the actions are too random.
    }
    \label{fig:full-entropy}
\end{figure}

Per \cite{schulman2017proximal}, we use entropy bonus to encourage exploration.
This bonus is added to the PPO loss function in the form of $c S[\pi_{\theta}](s_t)$, where $c$ is a hyperparameter referred to as entropy coefficient.
In initial stages of training a long-running experiment like \openaifive or \cleanexpname we set it to an initial value and lower it during training.
Similarly to \cite{schulman2017proximal}, \cite{mnih2016asynchronous}, or \cite{williams1991function}, we find that using entropy bonus prevents premature convergence to suboptimal policy.
In \autoref{fig:full-entropy}, we see that entropy bonus of 0.01 (our default) performs best.
We also find that setting it to $0$ in early training, while not optimal, does not completely prevent learning.

\begin{figure}
    \centering
        \begin{subfigure}[t]{0.49\textwidth}
        \centering
        \includegraphics[width=\textwidth]{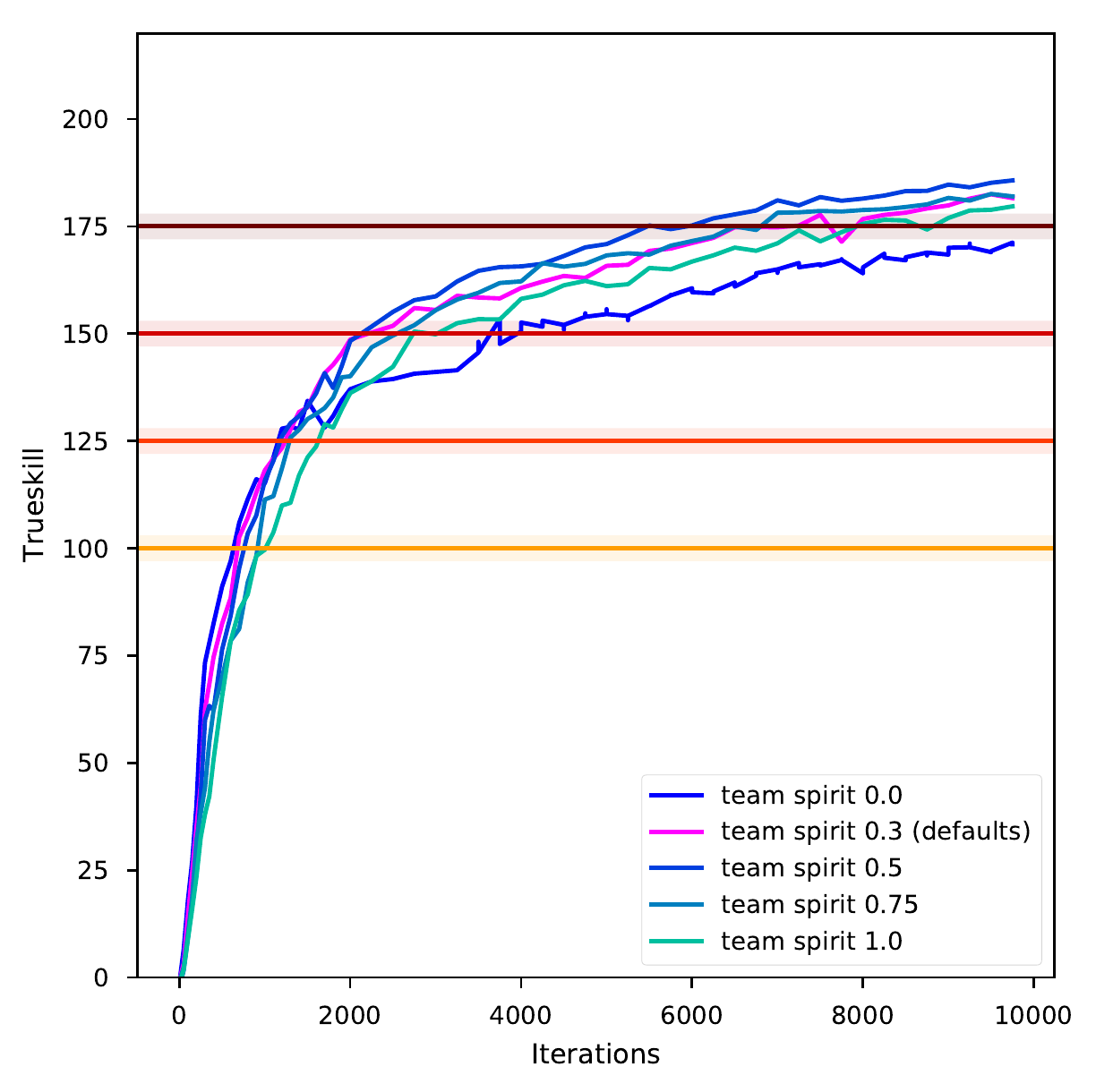}
    \end{subfigure}\hfill
    \begin{subfigure}[t]{0.49\textwidth}
        \centering
        \includegraphics[width=\textwidth]{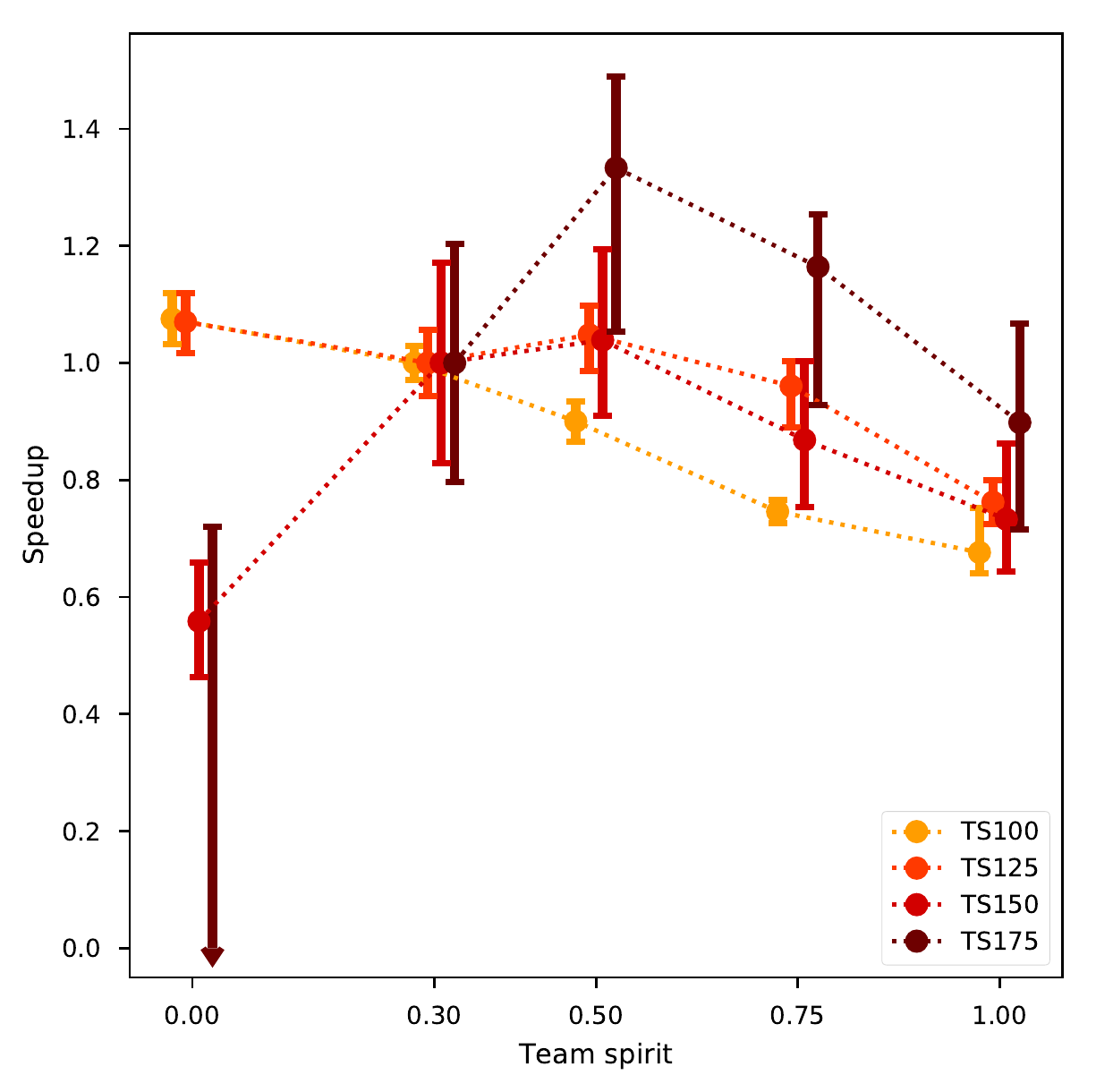}
    \end{subfigure} 
    \caption{\textbf{Team Spirit in early training: } 
    Very early in training (\trueskillname<125) the run with team spirit 0 does best; this can be seen by the speedup for lower \trueskillname being highest at team spirit 0.
    The maximum speedup quickly moves to 0.5 in the medium \trueskillname regime (150 and 175).
    }
    \label{fig:team-spirit}
\end{figure}

\begin{figure}
    \centering
        \includegraphics[width=0.5\textwidth]{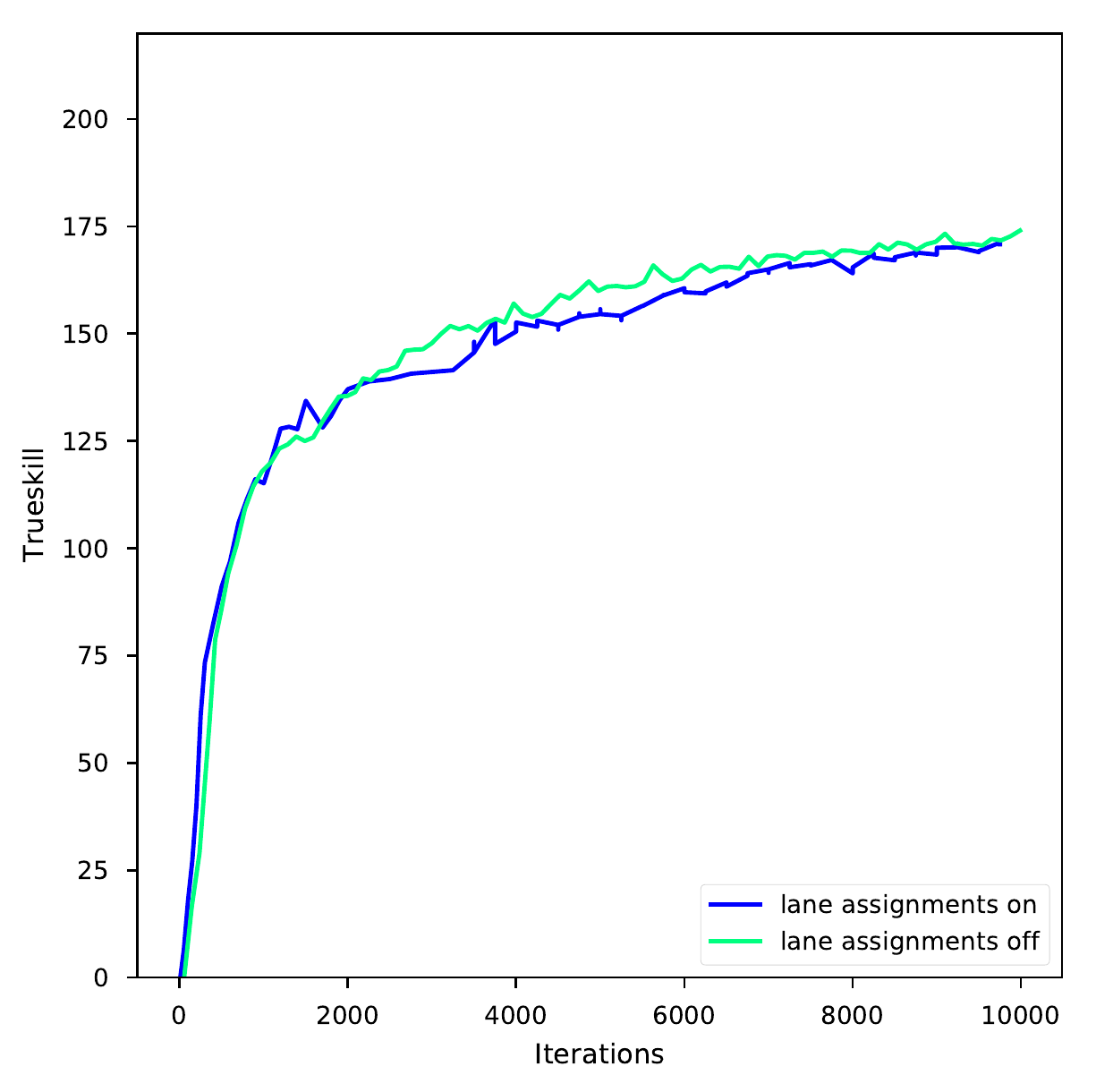}
    \caption{\textbf{Lane Assignments: } 
    ``Lane assignments'' randomization on vs. off. In this ablation we see that this randomization actually provided little benefit.
    }
    \label{fig:exploration}
\end{figure}


As discussed in \autoref{appendix:rewards}, we introduced a hyperparameter {\it team spirit} to control whether agents optimize for their individual reward or the shared reward of the team.
Early training and speedup curves for team spirit can be seen in \autoref{fig:team-spirit}.
We see evidence that early in training, lower team spirits do better.
At the very start team spirit 0 is the best, quickly ovdertaken by team spirit 0.3 and 0.5. 
We hypothesize that later in training team spirit 1.0 will be best, as it is optimizing the actual reward signal of interest.

\subsection{Environment Randomization}
\label{sec:random}
We further encouraged exploration through randomization of the environment, with three simultaneous goals:

\begin{enumerate}
\item If a long and very specific series of actions is necessary to be taken by the agent in order to randomly stumble on a reward, and any deviation from that sequence will result in negative advantage, then the longer this series, the less likely is agent to explore this skill thoroughly and learn to use it when necessary.
\item If an environment is highly repetitive, then the agent is more likely to find and stay in a local minimum.
\item In order to be robust to various strategies humans employ, our agents must have encountered a wide variety of situations in training.
This parallels the success of domain randomization in transferring policies from simulation to real-world robotics\cite{dactyl}.
\end{enumerate}

\label{sec:lane-assignments}
We randomize many parts of the environment:
\begin{itemize}
\item \textbf{Initial State: }
In our rollout games, heroes start with random perturbations around the default starting level, experience, and gold, armor, movement speed, health regeneration, mana regeneration, magic resistance, strength, intellect, and agility.
\item \textbf{Lane Assignments: }
From a strategic perspective it makes sense for heroes to act in certain area of the map more than the others.
Most inter-team skirmishes happen on {\it lanes}  ($3$ distinct paths that connect opposing bases).
At a certain stage of our work, we noticed that our agents developed a preference to stick together as a group of $5$ on a single lane, and fighting any opponent coming their way.
This represents a large local minimum, with higher short-term reward but lower long-term one as the resources from the other lanes are lost.
After that we introduced {\it lane assignments}, which randomly assigned each hero to a subset of lanes, and penalized them with negative reward for leaving those lanes.
However, the ablation study in \autoref{fig:exploration} indicates that this may not have been necessary in the end.
\item \textbf{Roshan Health: }Roshan is a powerful neutral creature, that sits in a specific location on the map and awaits challengers.
Early in training our agents were no match for it; later on, they would already have internalized the lesson never to approach this creature.
In order to make this task easier to learn, we randomize Roshan's health between zero and the full value, making it easier (sometimes much easier) to kill.
\item 
\textbf{Hero Lineup: }
In each training game, we randomly sample teams from the hero pool.
While the hero randomization is necessary for robustness in evaluations  against human players (which may use any hero teams), we hypothesize that it may serve as additional exploration encouragement, varying the game and preventing premature convergence.
In \autoref{sec:multihero}, we see that training with additional heroes causes only a modest slowdown to training despite the extra heroes having new abilities and strategies which interact in complex ways.
\item \textbf{Item Selection: }
Our item selection is scripted: in an evaluation game, our agents always buy the same set of items for each specific hero.
In training we randomize around that, swapping, adding, or removing some items from the build.
This way we expose our agents to enemies playing with and using those alternative items, which makes our agents more robust to games against human players.
There are shortcomings to this method, e.g. a team with randomly picked items is likely to perform worse, as our standard build is carefully crafted.
In the end our agent was able to perform well against humans who choose a wide variety of items.
\end{itemize}

\section{Hero Pool Size}\label{sec:multihero}
\begin{figure}
    \centering
    \begin{subfigure}[t]{0.49\linewidth}
        \vskip 0pt 
        \includegraphics[width=\textwidth]{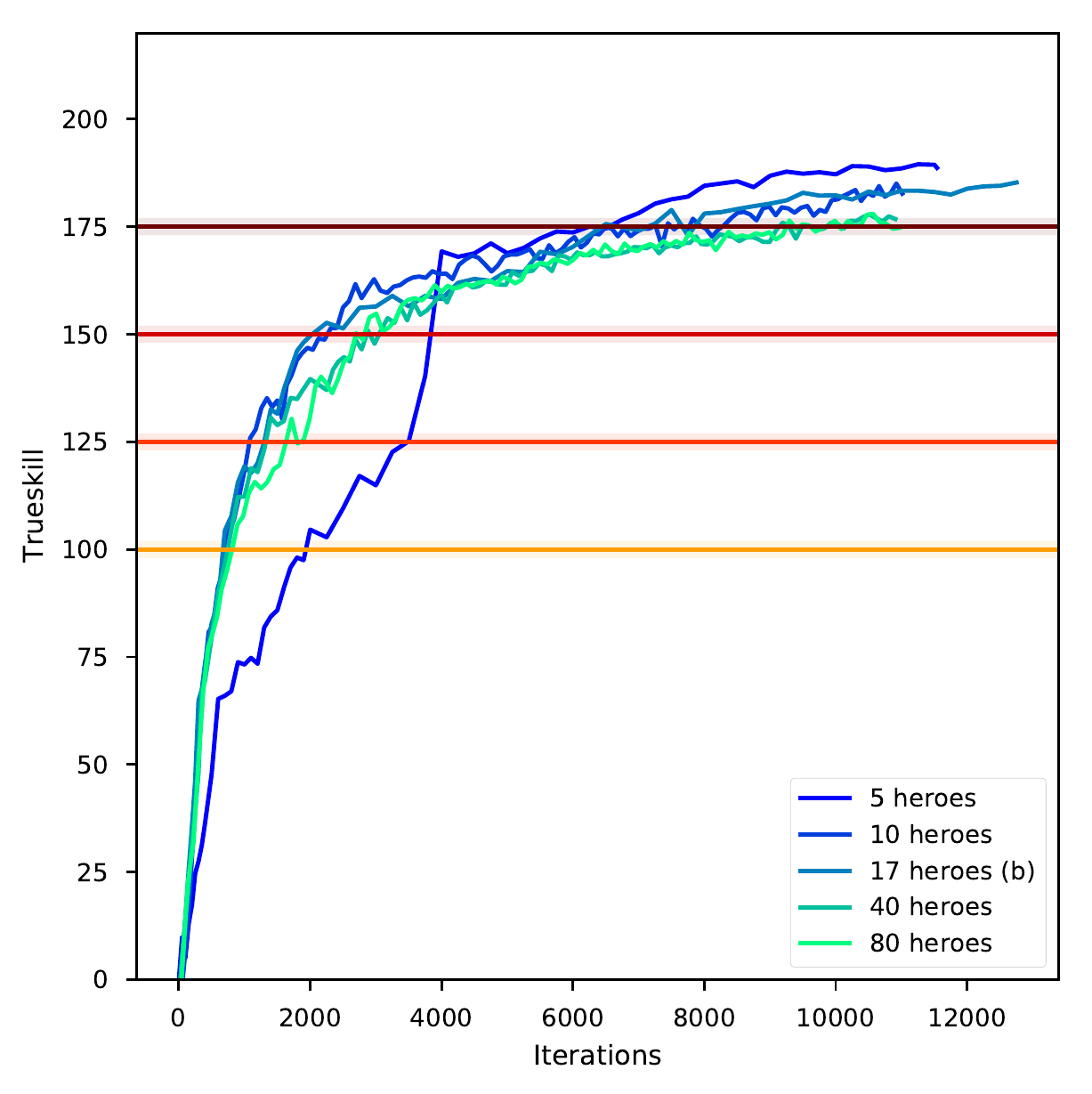}
    \end{subfigure}
    \begin{subfigure}[t]{0.49\linewidth}
        \vskip 0pt 
        \includegraphics[width=\textwidth]{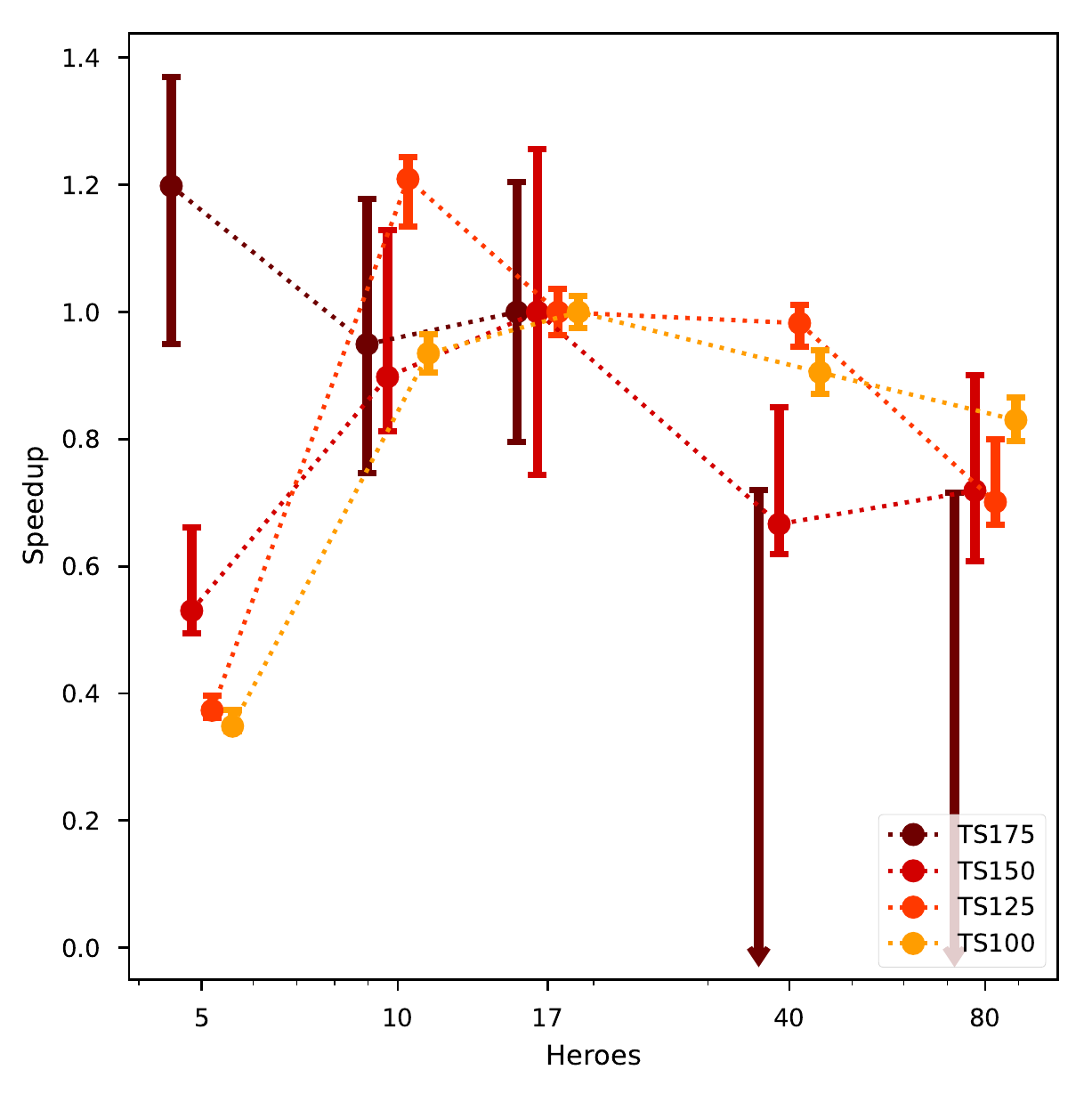}
    \end{subfigure}
    \caption{\textbf{Effect of hero pool size on training speed:} \genericcaptionsentence{varying the size of the hero pool.}  
    Additional heroes slows down early training only slightly.
    The severe underperformance of the 5-hero run for the first 4k versions was not investigated in detail.
    It is likely not due to the hero count but rather some instability in that particular training run.}
    \label{fig:multihero}
\end{figure}
One of the primary limitations of our agent is its inability to play all the heroes in the game.
We compared the progress in early training from training with various numbers of heroes.
In all cases, each training game is played using an independent random sampling of five heroes from the pool for each team.
To ensure a fair comparison across the runs, evaluation games are played using only the smallest set of heroes.
Because the test environment uses only five heroes, the runs which train with fewer heroes are training closer to the test distribution, and thus can be expected to perform better; the question is how much better?

In \autoref{fig:multihero}, we see that training with more heroes causes only a modest slowdown. 
Training with 80 heroes has a speedup factor of approximately 0.8, meaning early training runs 20\% slower than with the base 17 heroes.
From this we hypothesize that an agent trained on the larger set of heroes using the full resources of compute of \cleanexpname would attain a similar high level of skill with approximately 20\% more training time.
Of course this experiment only compares the very early stages of training; it could be that the speedup factor becomes worse later in training.


\section{Bloopers}\label{appendix:bloopers}
\subsection{Manually Tuned Hyperparameters}
\begin{figure}
    \centering
    \includegraphics[width=0.8\textwidth]{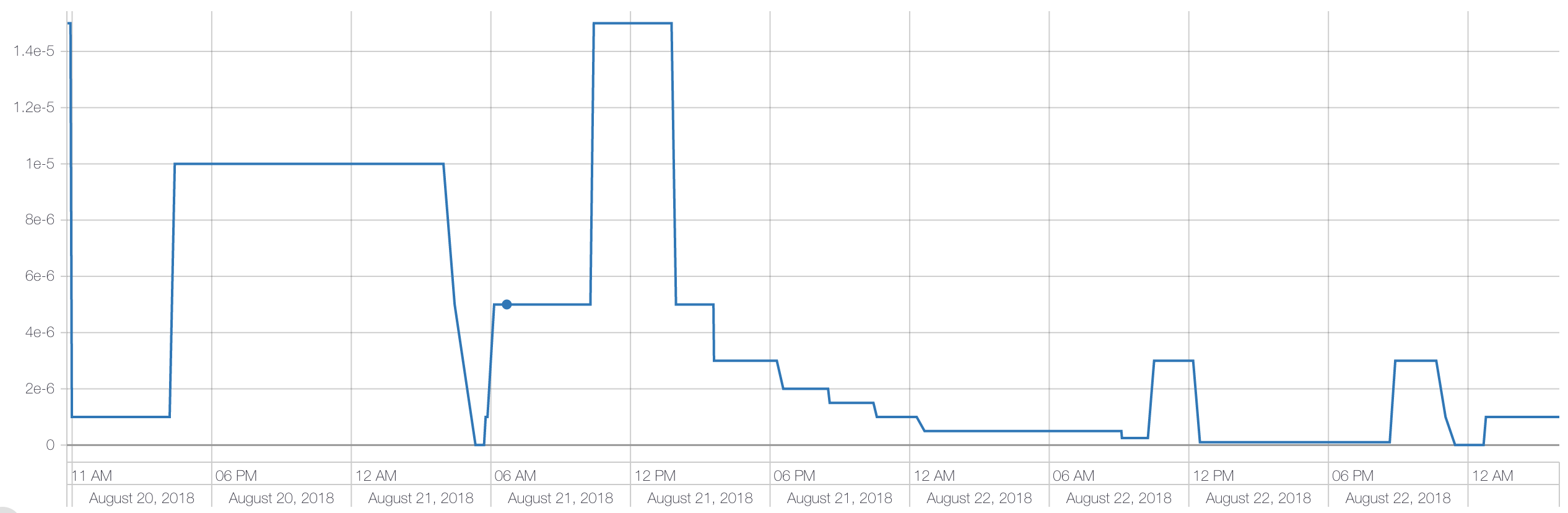}
    \caption{\textbf{Learning Rate during The International: }
    This is what happens when humans under time pressure choose hyperparameters.
    We believe that in the future automated systems should optimize these hyperparameters instead.
    After this event, our team began to internally refer to the act of frantically searching over hyperparameters as ``designing skyscrapers.''
    }
    \label{fig:ti-lr}
\end{figure}

Leading into The International competition in August 2018, we already a very good agent, but we felt it was likely not yet as good as the very best humans (as indeed turned out to be the case; we lost both games at that event).
In the final few days before the games, we sought to explore high-variance options which had a chance of offering a surprising improvement. 
In the end, however, we ultimately believe that human intuition, especially under time pressure, is not the best way to set hyperparameters. 
See \autoref{fig:ti-lr} for the history of our learning rate parameter during those few days.

\subsection{Zero Team Spirit Embedding}
One of our team members stumbled upon a very strange phenomenon while debugging a failed surgery. 
It turned out that replacing a certain set of 128 learned parameters in the model with zero increased the model's performance significantly (about 55\% winrate after versus before).
We believe that the optimizers were unable to find this direction for improvement because although the win rate was higher, the shaped reward (see \autoref{table:rewards}) was approximately the same.
A random perturbation to the parameters should have overwhelming probability of making things worse rather than better. 
We do not know why zero would be a special value for these parameters.

These parameters were certainly an unusual piece of the model.
In the early stages of applying team spirit (see \autoref{appendix:rewards}), we attempted to randomize the team spirit parameter in each game. 
We had a fixed list of four possible team spirit values; each rollout game one was chosen at random.
The agent was allowed to observe the current team spirit, via an embedding table with four entries.
We hoped this might encourage exploring games of different styles, some very selfless games and some very selfish games.

After training in this way for only a short period, we decided this randomization was not helping, and turned it off.
Because our surgery methods do not allow for removing parameters easily, we simply set the team spirit observation to always use a fixed entry in the embedding table.
In this way we arrived at a situation where the vector of ``global'' observations $g$ consisted of the real observations $g_r$, concatenated with 128 dimensions from this fixed embedding $E$; these extra dimensions were learned parameters which did not depend on the observations state:
\begin{equation}
    g = \left[g_r, E\right]
\end{equation}

Because this vector is consumed by a fully connected layer $Wg+B$, these extra parameters do not affect the space of functions representable by the neural network. They are exactly equivalent to not including $E$ in the global observations and instead using a modified bias vector:
\begin{equation}
    B'=W\left[0, E\right] + B
\end{equation}
For this reason we were comfortable leaving this vestigial part of the network in place.

Because it was an embedding, there should be nothing special about 0 in the 128-dimensional space of possible values of $E$. 
However we see clear evidence that zero \emph{is} special, because a generic perturbation to the parameters should have a negative effect.
Indeed, we tried this explicitly --- perturbing these parameters in other random directions --- and the effect was always negative except for the particular direction of moving towards zero.

\subsection{Learning path dependency}

The initial training of \openaifive was done as a single consecutive experiment over multiple months. During that time new items, observations, heroes, and neural network components were added. The order of introduction of these changes was a priori not critical to learning, however when reproducing this result in \cleanexpname with all the final observations, heroes, and items included, we found that one item --- Divine Rapier --- could cause the agents to enter a negative feedback loop that reduced their skill versus reference opponents. As Rapier began to be used, we noted a decrease in episodic reward and \trueskillname. When repeating this experiment with Rapiers banned from the items available for purchase, \trueskillname continues to improve.

We hypothesize that this effect was not observed during our initial long-lasting training because Rapier was only added after the team spirit hyperparamater was raised to 1 (see \autoref{appendix:rewards}). Rapier is a unique item which does not stay in your inventory when you die, but instead falls to the ground and can be picked up by enemies or allies. Because of this ability to transfer a high-value item, it is possible that the reward collected in a game increases in variance, thereby preventing \openaifive from learning a reliable value function.


\end{document}